\documentclass{article}

\usepackage[main, final]{neurips_2026}

\usepackage[utf8]{inputenc} 
\usepackage[T1]{fontenc}    
\usepackage{hyperref}       
\usepackage{url}            
\usepackage{booktabs}       
\usepackage{amsfonts}       
\usepackage{nicefrac}       
\usepackage{microtype}      
\usepackage{xcolor}         

\usepackage{amsmath}
\usepackage{amssymb}
\usepackage{mathtools}
\usepackage{amsthm}
\usepackage{bbm}
\usepackage{enumitem}
\usepackage{ulem}
\usepackage[blocks]{authblk}

\usepackage{algorithm} 
\usepackage{algorithmic}

\usepackage{graphicx}
\usepackage{subcaption}

\theoremstyle{plain}
\newtheorem{theorem}{Theorem}[section]
\newtheorem{proposition}[theorem]{Proposition}
\newtheorem{lemma}[theorem]{Lemma}

\theoremstyle{definition}
\newtheorem{definition}[theorem]{Definition}
\newtheorem{assumption}[theorem]{Assumption}
\theoremstyle{remark}
\newtheorem{remark}[theorem]{Remark}

\newcommand{\prob}[1]{\mathrm{Pr}\left(#1\right)}

\newcommand{\expecto}[2]{\mathbb{E}_{#1}\left[#2\right]}

\newcommand{\abs}[1]{\left|#1\right|}
\newcommand{\id}[1]{\mathbb{I}\left(#1\right)}

\title{Practical Estimation of the Bayes Optimal Fairness-Accuracy Tradeoff with Soft Labels}

\author{
    \begin{minipage}{\textwidth}
        \centering
        \textbf{Mohit Sharma}$^{1}$\thanks{Work done while the author was an intern at RIKEN-AIP. All correspondence to mohit.sharma.cs29@gmail.com.} \quad \textbf{Okan Ko\c{c}}$^{2}$ \quad \textbf{Amit Deshpande}$^{3}$ \quad \textbf{Takashi Ishida}$^{4,2}$ \par
        \vspace{0.3em}
        \textbf{Gang Niu}$^{2}$ \quad \textbf{Rajiv Ratn Shah}$^{5}$ \quad \textbf{Masashi Sugiyama}$^{4,2}$ \par
        \vspace{0.8em}
        \small $^{1}$IIIT Delhi \quad $^{2}$RIKEN-AIP \quad $^{3}$Microsoft Research \quad $^{4}$The University of Tokyo \quad $^{5}$IIT Kanpur
    \end{minipage}
}

\begin{document}

\maketitle

\begin{abstract}
  There is a fundamental limit to the best prediction error of any model for a given data distribution. In classification, the Bayes error quantifies this limit and serves as both a benchmark for trained models and a criterion to detect overfitting. The state-of-the-art Bayes error estimation techniques for binary classification use only soft labels (positive class probabilities) and require no input features or auxiliary training. We extend such notions to constrained classification problems, such as fairness. Fair classification is an important case of constrained or multi-objective classification. It requires a model not only to be accurate but also to yield fair outcomes (equal or near-equal true positive rates) across different demographic groups (e.g., race and gender). This leads to an inherent fairness-accuracy tradeoff or a Pareto frontier that is used to compare fair classifiers against each other. The focus of our work is to estimate the fundamental limit for this fairness-accuracy tradeoff. We derive an explicit expression for the fair Bayes error rate (i.e., best achievable error rate subject to given fairness constraints), and construct consistent, instance-free estimators for it using only soft labels and group information. We provably bound the bias and the sample complexity of our estimators. Empirically, our method produces a robust, low-variance estimate of the optimal fairness-accuracy tradeoff curve, even when the soft labels are noisy and derived from a classification model on hard labels.
\end{abstract}

\section{Introduction}
\label{sec:intro}

The notion of Bayes optimality in machine learning defines the best achievable performance, given a data distribution, with respect to the $0$-$1$ loss \cite{cover1967nearest, fukunaga2013introduction, devroye2013probabilistic, mohri2018foundations}. For binary classification, an observation consist of features \(X \in \mathcal X \subseteq \mathbb R^d\), a label \(Y \in \mathcal Y=\{0,1\}\), and a sensitive attribute \(A \in \mathcal A=\{0,1\}\), governed by an underlying distribution $(X, Y, A) \sim \mathcal{D}$. A group-aware classifier \(g:\mathcal X\times\mathcal A\to\{0,1\}\) predicts labels using both features and sensitive attributes. A well-known result states that the Bayes optimal classifier $g^*$ can be expressed as the following decision rule: $g^*(x, a) = \id{\eta(x,a) \geq \nicefrac{1}{2}}$ \cite{mohri2018foundations, zeng2024bayes}. Here, $\id{}$ denotes the indicator function and $\eta(x,a)\;:=\;\Pr(Y=1\mid X=x, A=a)$ defines the posterior, which assigns a soft label \(c_i:=\eta(x_i,a_i)\) to an instance.
This rule also allows us to estimate the Bayes error, the best achievable performance for a given distribution, without access to any instances \citep{mohri2018foundations, ishidaperformance}: $\beta = \expecto{X, A}{\min\{ C, 1 - C\}}$, 
where $C := \eta(X,A)$ is the random variable corresponding to the soft labels.

This notion can also be extended to multi-objective classification problems, such as those that enforce fairness constraints \citep{barocas2023fairness}. With the widespread deployment of machine learning models, it is essential that models are fair and do not produce disproportionate errors and harm to certain sensitive groups in the data \citep{barocas2023fairness, angwin2016machine, buolamwini2018gender}. However, applying fairness constraints generally introduces a tradeoff with accuracy, since common fairness constraints are incompatible with the standard notion of accuracy \citep{pleiss2017fairness, menon2018cost}. This tradeoff depends sharply on (i) the underlying data distribution, (ii) the fairness metric (equality of positives or true positives), and (iii) the degree of allowed unfairness tolerance.

\begin{figure*}
    \centering
    \begin{subfigure}{0.4\linewidth}
        \includegraphics[width=\linewidth]{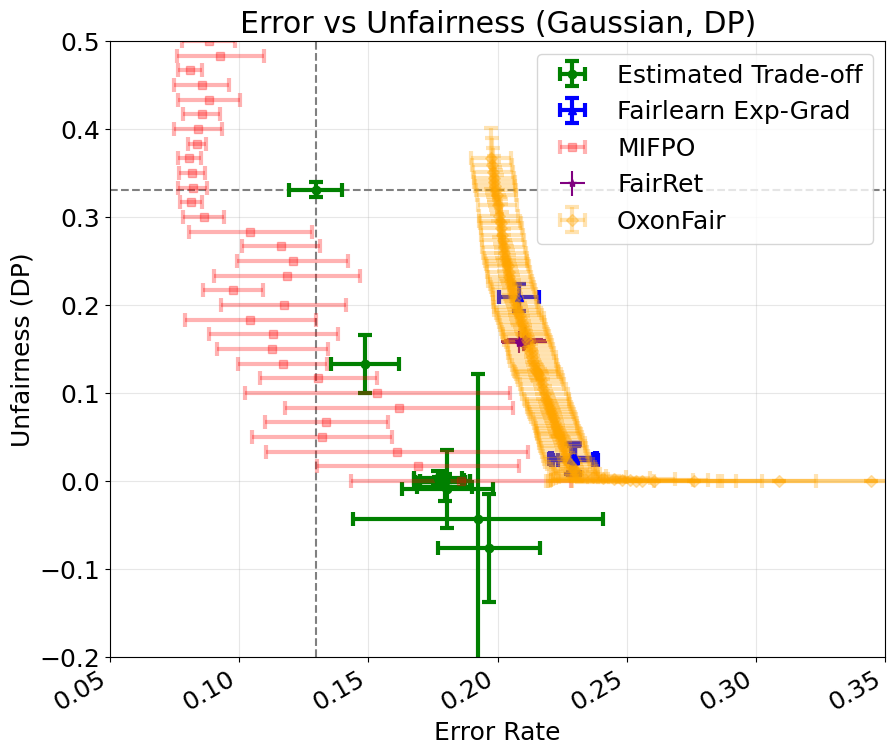}
    \end{subfigure}
    \begin{subfigure}{0.4\linewidth}
        \includegraphics[width=\linewidth]{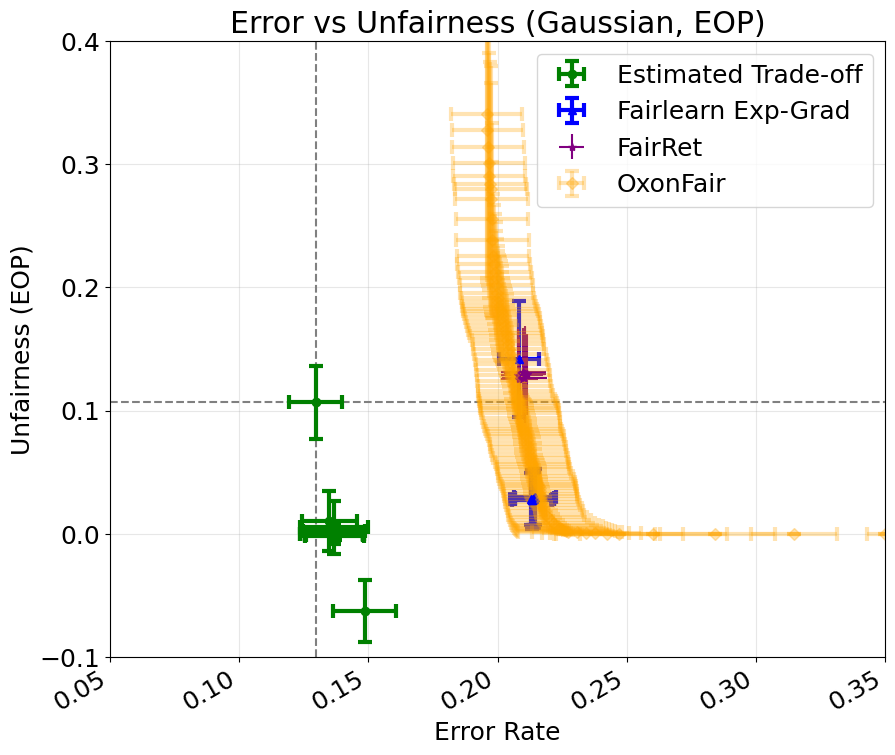}
    \end{subfigure}

    \caption{Unfairness-error tradeoffs on synthetic data \cite{zeng2022bayes} for Demographic Parity (DP) and Equal Opportunity (EOP) constraints. The proposed estimate (green) establishes a lower bound against competing algorithms. Dotted lines denote the unconstrained Bayes Optimal classifier at $\lambda=0$.}
    \label{fig:introimage}
\end{figure*}

Bayes optimal fair classifiers can often be characterized as adjustments of the classification thresholds: $g^*_{\mathrm{fair}}(x,a) = \id{\eta(x,a) \geq \nicefrac{1}{2} + t_a(\delta)}$ where $t_a(\delta) > 0$ is a group-dependent adjustment that depends on the fairness metric, the underlying data distribution and the tolerance for unfairness $\delta$. This is precisely where existing works on estimating the Bayes error do not generalize \citep{ishidaperformance, jeong2024data, ushio2025practical}. 
While Bayes error estimators solely rely on posterior probabilities (soft scores), resulting in a constant threshold of $1/2$, this is not the case for cost-sensitive thresholds, such as fairness. The error estimators for these cost-sensitive thresholds also require an estimator for the underlying thresholds.

Therefore, to estimate the optimal fairness-accuracy tradeoff, we first show how we can disentangle the unfairness level $\delta$ and the corresponding threshold adjustment $t_a(\delta)$ by studying when the classifiers become trivial (Lemma \ref{lem:lambda_max}). Using this, we can estimate the threshold and, eventually, the unfairness and error rate at several points along the tradeoff curve by leveraging its monotonicity properties (Lemma \ref{lem:monotone-lambda}). We show that fair classification can be reduced to cost-sensitive thresholding with appropriate $\lambda$-dependent thresholds (Lemma \ref{lem:scott_def}), where $\lambda \in \mathbb{R}$ is a parameter that directly controls the correction to the classification threshold and depends on $\delta$. We then derive consistent estimators for the fair classification thresholds, only using soft labels, at any fixed $\lambda$ (Theorem \ref{thm:alpha_cons}).

Once we have consistent estimates of thresholds, we then, as an independent result, generalize the Bayes error estimators in prior works \cite{ishidaperformance, jeong2024data, ushio2025practical} to work with any cost-sensitive risk (Theorem \ref{thm:beta_cs_cons}). Since our thresholds and estimators are both constructed using finite soft labels, we move beyond the scope of prior art to study consistency, bias, and sample complexity of the estimation of the cost-sensitive Bayes risk (Theorem \ref{thm:bayes_cs_hat_cons}). These results and rates apply to many cost-sensitive learning scenarios \citep{koyejo2014consistent, singh2022optimal} that require distribution-dependent classification thresholds.

We then focus on estimating the optimal fairness-accuracy tradeoff using only soft labels. We first derive the estimated risk at any threshold and show its consistency properties (Theorem \ref{thm:err_01_cons}). We then also provide estimators for the expected unfairness under demographic parity and equal opportunity, and derive their consistency and sample complexity properties (Theorem \ref{thm:beta_unf_cons}). All of them culminate in a bisection-search driven algorithm with our estimators (Algorithm \ref{algo:pareto}), inspired by a similar algorithm for the Bayes optimal fair classifiers \citep{zeng2024bayes}, but only with soft labels and without instances. An example of this bisection algorithm in action is presented in Figure \ref{fig:introimage} (green curve).


Finally, we estimate the fairness-accuracy tradeoff on real-world datasets. We compare our estimated tradeoff against several recent works on estimating the fairness-accuracy tradeoff \citep{delaney2024oxonfair, kozdoba2024efficient} and differentiable objectives for fair classification \citep{buyl2023fairret}, in Figures \ref{fig:introimage} and \ref{fig:siximages}. We show that our estimated tradeoff lower bounds the tradeoff estimated by these methods. Whenever it does not (for example, in the case of the algorithm proposed in \cite{kozdoba2024efficient}), it is explained by the algorithm's optimization of a completely different lower bound on the Pareto frontier. We also perform experiments with noisy soft scores generated by a Logistic Regression model on the Adult dataset \citep{lichman2013uci} and find that our method still lower-bounds all other fairness tradeoff algorithms.

Overall, in a similar spirit to prior work \cite{ishidaperformance, jeong2024data, ushio2025practical}, where the authors use the Bayes error as a tool to validate performance and check for overfitting, our estimated tradeoff with soft labels establishes a similar precedent: Any current or future methods aiming to compare fair algorithms should always close the gap between their estimated tradeoff and the theoretically best, that we estimate. 

\textbf{Related Work}: Our work is inspired by the prior results \citep{ishidaperformance, jeong2024data, ushio2025practical} that characterize the performance limitations in terms of accuracy using the threshold characterization of the Bayes optimal classifier. 
There are several other works on Bayes error estimation that aim to estimate the Bayes error indirectly \citep{berisha2015empirically, noshad2019learning, theisen2021evaluating, renggli2021evaluating}. But since most of these methods estimate the Bayes error indirectly, without leveraging soft labels or thresholding rules, generalizing them to constrained optimization settings, such as fairness constraints and cost-sensitive risk, is future work beyond the scope of this work. A major theme of this paper centers around fair machine learning, which has received rightful attention in recent years \cite{barocas2023fairness}. Several fairness metrics, such as demographic parity \citep{dwork2012fairness} and equal opportunity \citep{hardt2016equality}, have been proposed, and corresponding fair Bayes optimal classifiers have been extensively studied in the literature \cite{woodworth2017learning, menon2018cost}. We make extensive use of threshold characterizations of fair Bayes optimal classifiers \citep{menon2018cost, chzhen2019leveraging, zeng2024bayes}. Finally, a major part of our work also concerns deriving consistency and sample-complexity properties of the thresholds and risk and unfairness estimators. Some related estimators have also been studied in prior work \cite{khurana2021statistical, jeong2024data}.

\section{Notations and Assumptions}

Let $\pi_a = \prob{A=a}$, $\pi_{ya} = \prob{Y=y, A=a}$, $\pi_{y|a} = \prob{Y=y|A=a}$, $\pi = \prob{Y=1}$, and 
$\prob{Y=y} = \underset{a \in \mathcal{A}}{\sum} \pi_{ya}$. We assume the existence of all groups and classes, i.e., $\pi_a > 0, \pi_{y|a} >0$ for all $y \in \mathcal{Y}$ and for all $a \in \mathcal{A}$. Finally, we let $\overset{p}{\rightarrow}$ denote convergence in probability. All probabilities and expectations are taken with respect to the underlying distribution \(\mathcal{D}\).

Whenever we work with a finite sample, we assume that we are given access to $N$ i.i.d. soft labels of the form $(c_i, a_i)$, and we will denote our dataset with $D = \{(c_i, a_i)\}_{i=1}^N$. Since our approach is model-free, we do not require access to features $x_i$. We will also partition our dataset $D$ into two parts: $D = D_\alpha \cup D_\beta$, with $D_\alpha \cap D_\beta = \varnothing$, where $D_\alpha$ (with $N_\alpha$ samples) will be the data used to estimate the thresholds that we will put on $\eta(x, a)$, and $D_\beta$ (with $N_\beta$ samples) will be used to construct the estimators of classification risk and unfairness, with $N = N_\alpha + N_\beta$. The splits are chosen independently of the sample values. Furthermore, we will denote with $N_{\alpha, a}$ (or $N_{\beta, a}$) the number of samples from group $A=a$. Lastly, $i \in [D]$ denotes drawing an index from the dataset $D$. We need two independent sample sets because we estimate the required quantities in two steps. First, using $D_{\alpha}$, we estimate the distribution priors and group-dependent thresholds (Theorem \ref{thm:alpha_cons}), and next using $D_{\beta}$, we estimate the cost-sensitive Bayes risk, the expected error, and unfairness (Theorems \ref{thm:bayes_cs_hat_cons}, \ref{thm:err_01_cons}, and \ref{thm:beta_unf_cons} respectively). We now assume the following about our soft labels:

\begin{assumption}(No interior atoms / Continuity of the Cumulative Distribution function of $\eta$) \label{assump: eta-continous}
    The mapping $t \rightarrow F_a(t) = \prob{\eta(X, a) \leq t|A=a}, \forall a \in \mathcal{A}$ is continuous in $(0,1)$. Moreover, $\prob{\eta(X,a) \geq \nicefrac{1}{2}| A=a} > 0$ for all $a \in \mathcal{A}$.
\end{assumption}


\begin{assumption}(Lipschitz continuous group CDF) \label{assump: eta-density}
    For each $a \in \mathcal{A}$, the group-conditional distribution of $C=\eta(X,a) | A=a$ admits a density $f_a$ on $(0,1)$ such that $\underset{t \in (0,1)}{\sup} f_a(t) \leq L_a < \infty$. In particular, $F_a$ is $L_a$-Lipschitz: $|F_a(u) - F_a(v)| \leq L_a\cdot |u - v|$ for all $u,v \in (0,1)$.
\end{assumption}

Let $\hat{\pi}_a = \frac{1}{N_\alpha} \sum_{i \in [D_\alpha]} \id{a_i = a}$ denote our estimator for $\pi_a$. By the Weak Law of Large Numbers, $\hat{\pi}_a \overset{p}{\rightarrow} \pi_a$. Furthermore, the bias of the estimator $\hat{\pi}_a$ is zero, which is easy to see via the linearity of expectation. We can derive a similar unbiased estimator for $\pi = \prob{Y=1}$ as $\hat{\pi} = \frac{1}{N_\alpha} \sum_{i \in [D_\alpha]} c_i$.

Similarly, let $\hat{\pi}_{1a} = \frac{1}{N_\alpha} \sum_{i \in [D_\alpha]} \id{a_i = a}c_i$ denote our estimator for $\pi_{1a}$. By the Weak Law of Large Numbers, $\hat{\pi}_{1a} \overset{p}{\rightarrow} \pi_{1a}$. Furthermore, the bias of the estimator $\hat{\pi}_{1a}$ is zero, which is easy to see via the linearity of expectation. The same analysis holds for the unbiased estimator $\hat{\pi}_{1|a} = \frac{1}{N_{\alpha,a}} \sum_{i \in [D_\alpha]} \id{a_i = a}c_i$ and its expected value $\pi_{1|a}$. Since we will work with inverse functions of $\hat{\pi}_a$ (and other estimators), we also define $\hat{\pi}^+_a = \max\{\hat{\pi}_a, \frac{1}{N_\alpha}\}$, which is still statistically consistent, but a biased estimator (same for $\hat{\pi}^+_{1a}$ and $\hat\pi_{1|a}^+$). Finally, let $\Pi_{[0,1]}(t):= \min\{1, \max\{0,t\}\}$ denote the truncation operation, which we will frequently use to ensure our estimators are bounded between $0$ and $1$. All proofs are provided in the Appendix. 


\section{Bayes Optimal Fair Classifiers}
\label{sec:prelim}

We will now describe a well-known result that describes the Bayes optimal classifier for binary classification, when we know the underlying distribution.


\begin{lemma} \label{lem:uncons_bo}
    \citep{devroye2013probabilistic, menon2018cost, zeng2024bayes} The Bayes optimal classifier $g^*$ for a given distribution can be precisely characterized as: $g^*(x,a) = \id{\eta(x,a) \geq \nicefrac{1}{2}}$.
\end{lemma}


However, as noted in the literature \citep{menon2018cost, chzhen2019leveraging, zeng2024bayes}, the unconstrained Bayes optimal classifier for a distribution may not satisfy fairness constraints. To study this phenomenon in more detail, we now introduce definitions of the fairness constraints we consider in this work.

\begin{definition} \label{def:metrics}
A classifier $g$ is said to be $\delta$-fair if: (1) Demographic Parity (DP) \citep{dwork2012fairness}: $|\prob{g(X,A) = 1| A=1} - \prob{g(X,A) = 1| A=0}| \leq \delta$, and (2) Equal Opportunity (EOP) \citep{hardt2016equality}: $|\prob{g(X,A) = 1| Y=1, A=1}$ $- \prob{g(X,A) = 1| Y=1, A=0}| \leq \delta$.
\end{definition}

Intuitively, a classifier satisfies demographic parity whenever it classifies approximately the same number of individuals from different groups as positive. However, this does not account for the error incurred within each group. The equal opportunity constraint overcomes this by enforcing approximately equal true positive rates between the two groups. Zeng et al. \cite{zeng2024bayes} recently characterized the Bayes optimal $\delta$-fair classifiers whenever the fairness metric can be expressed as a linear function of a classifier $g$. We introduce their definition below.


\begin{definition} \label{def:linear-disparity}
A fairness disparity measure $\Delta(g)$ is \emph{linear} if there exists a $w: \mathcal{X}\times\mathcal{A} \rightarrow \mathbb{R}$ such that for all classifiers $g$, $\Delta(g) = \int_\mathcal{A}\int_\mathcal{X} g(x,a)~w(x,a)d\mathbb{P}(x,a)$. Furthermore, it is \emph{bilinear} if there exist distribution-dependent quantities $s, b$  such that $ w(x, a) = s\cdot\eta(x, a) + b$.
\end{definition}

\begin{remark} \label{rem:dp_eo_dis}
    Proposition $3.4$ in Zeng et al. \cite{zeng2024bayes} shows the suitable choices of $s$ and $b$ for the fairness constraints mentioned in Definition \ref{def:metrics}. For demographic parity, $w_{DP}(x,a) = \frac{(2a - 1)}{\pi_a}$ and for equal opportunity, $w_{EOP}(x,a) = \frac{(2a - 1)\eta(x,a)}{\pi_{1a}}$.
\end{remark}

We will now define the risk of a classifier $g$ and its disparity for linear disparity metrics \cite{zeng2024bayes}.



\begin{definition} (Equation A.1 and 4.2 from \citep{zeng2024bayes}) \label{def:risk_disparity}
For a classifier $g$, we can define the risk $R(g)$ and disparity $\Delta(g)$ as: $R(g) = \underset{a \in \mathcal{A}}{\sum} \pi_a \int_\mathcal{X}(1 - 2\eta(x,a))g(x,a)d\mathbb{P}_{X|A=a}(x) + \underset{a \in \mathcal{A}}{\sum} \pi_a \int_\mathcal{X} \eta(x,a)d\mathbb{P}_{X|A=a}(x)$, and $\Delta(g) = \underset{a \in \mathcal{A}}{\sum} \pi_a \int_\mathcal{X} \left[ w(x,a)\cdot g(x,a)\right] d\mathbb{P}_{X|A=a}(x)$.
\end{definition}

Because we are working with binary classification, introducing the class posterior probability allows us to decompose the risk into purely classifier-dependent and distribution-dependent terms. We now introduce the optimal fair classification problem. For at most $\delta$-unfairness, we want to solve the following optimization problem: $g_\delta ~\in~ \underset{g}{\arg\min} ~R(g) ~\text{ subject to }~ |\Delta(g)| \leq \delta$. For demographic parity and equal opportunity, we can precisely write down the fair Bayes optimal classifier that solves this optimization problem, where the threshold depends on the distribution. 


\begin{lemma} \label{lem:fair-threshold}
(Proposition 2.3 in \citep{chzhen2019leveraging}, Theorem 4.2 in \citep{zeng2024bayes}) Under Assumption \ref{assump: eta-continous}, the Bayes optimal $\delta$-fair classifier takes the following form: $g_\delta(x,a) = \id{\eta(x,a) \geq \frac{1}{2} + \frac{\lambda^*(\delta)}{2}w(x,a)}$,
where $\lambda^*(\delta)$ is the optimal dual parameter for the optimization program.
\end{lemma}

\begin{remark} \label{rem:fair-threshold}
    Using Remark \ref{rem:dp_eo_dis}, we can immediately write the threshold characterizations for demographic parity: $g_\delta(x,a) = \id{\eta(x,a) \geq \frac{1}{2} + \frac{(2a - 1)\lambda^*(\delta)}{2\pi_a}}$ and equal opportunity: $g_\delta(x,a) = \id{\eta(x,a) \geq \frac{1}{2 - \frac{(2a - 1)\lambda^*(\delta)}{\pi_{1a}}}}$.
\end{remark}


Lemma \ref{lem:fair-threshold} implies that for a fixed $\delta$, there may exist a $\lambda^*$ (that depends on whether for a given $\delta$, finding a $\lambda^*(\delta)$ is feasible), with which we can adjust the unconstrained Bayes optimal threshold of $\nicefrac{1}{2}$. This correction depends on which group $A=a$ is underprivileged, as well as on the distribution-dependent quantities like $\pi_a, \pi_{ya}$.

Characterizing the precise value of $\lambda^*$ as a function of $\delta$ and distribution-dependent quantities is a challenging problem, and several prior works (e.g., \citep{menon2018cost, celis2019classification, chzhen2019leveraging, zeng2024bayes}) treat it as another optimization problem. However, in our approach, we will show how we can obtain a feasible range for $\lambda$. By doing so, we will first be able to parametrize the risk and the unfairness in terms of $\lambda$, and then use a bisection search to get arbitrarily close to the required level of unfairness $\delta$. In a similar spirit to the Bayes error estimation work in the literature \citep{ishidaperformance, jeong2024data, ushio2025practical}, this will allow us to achieve the theoretically optimal fairness-accuracy tradeoff using only soft labels.

Let $g_{\lambda} = \underset{g}{\text{argmin}}~ R(g) + \lambda \cdot \Delta(g)$, for $\lambda \in \mathbb{R}$ denote the primal solution for a fixed $\lambda$. We dub this as the $\lambda$-fair classifier, since this can potentially be a $\delta(\lambda)$-fair classifier. We can now write down the Bayes optimal $\lambda$-fair classifier using the threshold characterization in Lemma \ref{lem:fair-threshold}, which will be useful later in this paper. We now prove how distribution-dependent quantities bound $\lambda$.

\begin{lemma} \label{lem:lambda_max}
    Assuming $w(., 0) < 0$ and $w(.,1) > 0$ (Definition \ref{def:linear-disparity}), when $\Delta(g_0) > 0, \lambda_{\max} = \frac{1}{\underset{(x,a)}{\inf} |w(x,a)|}$, so that $\lambda \in [0, \lambda_{\max}]$ is sufficient to interpolate the entire Pareto frontier, including the $\lambda$ for exact fairness $\lambda^*(0)$. And when $\Delta(g_0) < 0, \lambda_{\min} =  \frac{-1}{\underset{(x,a)}{\inf} |w(x,a)|}$, so that $\lambda  \in [\lambda_{\min}, 0]$, is sufficient to interpolate the entire Pareto frontier. 
\end{lemma}

This condition becomes much simpler for the equal opportunity and the demographic parity fairness metrics: $\lambda_{\max} = \underset{a \in \mathcal{A}}{\max}~ \pi_{1a}$ and $\lambda_{\max} = \underset{a \in \mathcal{A}}{\max}~ \pi_a$, respectively. Restricting the possible range of $\lambda$ allows us to set up a bisection search to estimate the fairness-accuracy tradeoff (Algorithm \ref{algo:pareto}). We now argue that approximating this tradeoff curve pointwise is sufficient for our scenario to obtain a high-probability estimate of the true fairness-accuracy tradeoff. 



\begin{proposition} \label{prop:pareto_proof}
If a classifier $g$ belongs to the Pareto-frontier, then $g = g_{\lambda}$, for some $\lambda \in \mathbb{R}$.   
\end{proposition}

In the Appendix, we show that  $g_\lambda$ can be constructed as a threshold classifier of the form $g_\lambda(x,a) = \id{\eta(x,a) \geq \alpha_{a,\lambda}}$, where the threshold $\alpha_{a,\lambda}$ is $\lambda$ and distribution dependent (Remark \ref{rem:fair-threshold}). 

\section{Estimating the Unfairness and Error Rates Using Soft Labels} \label{sec:estimation}

With all the tools in place, we will now derive expressions for the expected and finite-sample estimators. We will first derive consistent estimators of the cost-sensitive thresholds $\alpha_a$ that depend on $\lambda$, the unfairness metric of choice, and some distribution-dependent quantities in Theorem \ref{thm:alpha_cons}. Once we have that, we will first show a consistent estimator for the cost-sensitive Bayes risk, in a similar spirit to the results in prior work \cite{ishidaperformance, ushio2025practical} in Theorem \ref{thm:beta_cs_cons} and then its finite sample version in Theorem \ref{thm:bayes_cs_hat_cons}. Since we are interested in estimating the fairness-accuracy tradeoff, we will then derive consistent estimators for the risk of a classifier (Theorem \ref{thm:err_01_cons}) and then the unfairness (Theorem \ref{thm:beta_unf_cons}). Finally, we will present an algorithm to construct the tradeoff curve (Algorithm \ref{algo:pareto} in the Appendix).

\subsection{Estimating Fair Thresholds using Finite Soft Labels}


Unlike prior work \citep{ishidaperformance, jeong2023demystifying, jeong2024data, ushio2025practical}, where soft labels were enough to estimate the Bayes risk $\beta$, in the case of any cost-sensitive risk $\beta_{CS}$, unless the cost $\alpha_a$ is a constant, we need to estimate $\alpha_a$ from given soft labels, since $\alpha_a$ would often depend on distribution dependent quantities. This is certainly true for the fairness constraints defined in Lemma \ref{lem:fair-threshold}. We now define the estimated thresholds for both DP and EOP fairness constraints (Lemma \ref{lem:fair-threshold} and Remark \ref{rem:fair-threshold}). For demographic parity: $\hat{\alpha}_a^+ = \Pi(\Tilde{\alpha}_a^+), \text{ where } \Tilde{\alpha}_a^+ = \frac{1}{2} + \frac{(2a - 1)\lambda}{2\hat{\pi}^+_a}$, and for equal opportunity: $\hat{\alpha}_a^+ = \Pi(\Tilde{\alpha}_a^+), \text{ where } \Tilde{\alpha}_a^+ = \frac{1}{2 - \frac{(2a - 1)\lambda}{\hat{\pi}_{1a}^+}}$, where $\Pi$ is the truncation operation defined earlier, that ensures that our estimated thresholds stay in the interval $[0,1]$. 

We now show that the estimators of $\alpha_a$ for DP and EOP, with finitely many soft labels, are statistically consistent. Since the specific construction of the estimator requires knowledge of $w(X, A)$, we present estimators for demographic parity and equal opportunity, and the proof techniques can be generalized to tackle other cost-sensitive applications. 

\begin{theorem} \label{thm:alpha_cons}
    (Consistency and Sample Complexity for threshold estimation) Given a dataset $D_\alpha$ of clean soft labels ($c_i = \eta(x_i, a_i)$) and a fixed $\lambda \in \mathbb{R}$, consider the empirical estimator $\hat{\alpha}_a^+$. Assuming $2 - \frac{(2a -1)\lambda}{\pi_{1a}} \geq \tau > 0$ (denominator bounded away from zero), the following holds:
    \begin{enumerate}
    \setlength{\itemsep}{0pt}
    \setlength{\parsep}{0pt}
    \setlength{\parskip}{0pt}  
        \item Consistency: $\hat{\alpha}^+_a \overset{p}{\rightarrow} \alpha_a$.
        \item Bias: 
        \begin{itemize}[leftmargin=*]
        \setlength{\itemsep}{0pt}
        \setlength{\parsep}{0pt}
        \setlength{\parskip}{0pt}
            \item DP: $|\alpha_a -\expecto{D_\alpha}{\hat{\alpha}_a^+}| \leq \frac{|
        \lambda|}{\pi_a }\sqrt{\frac{1 - \pi_a}{\pi_a N_\alpha}} + 2\exp(-\frac{\pi_a^2}{2}N_\alpha)$,
            \item EOP:  $|\alpha_a -\expecto{D_\alpha}{\hat{\alpha}_a^+}| \leq \frac{4|\lambda|}{\tau^2\pi_{1a}^2}\sqrt{\frac{\pi_{1a}(1 - \pi_{1a})}{N_\alpha}} + 2\exp(-2\epsilon^2N_\alpha), \text{ where } \epsilon \leq \min\{\frac{\pi_{1a}}{2}, \frac{\tau\pi_{1a}^2}{4|\lambda|}\}$.
        \end{itemize}
        \item Sample Complexity: With $1 - \delta$ probability,  $\prob{|\hat{\alpha}_a ^+- \alpha_a| > \epsilon}\leq \delta$, as long as:
        \begin{itemize}[leftmargin=*]
        \setlength{\itemsep}{0pt}
        \setlength{\parsep}{0pt}
        \setlength{\parskip}{0pt}
            \item DP: $N_\alpha > \max\{ \frac{2}{\pi_a^2} \log{\frac{4}{\delta}}, \frac{\lambda^2}{2\epsilon^2\pi_a^4}\log\frac{4}{\delta}, \frac{2}{\pi_a} \}$,
            \item EOP: $N_\alpha \geq \frac{1}{2t^2} \log(\frac{2}{\delta})$ for $t = \min \Big\{\frac{\pi_{1a}}{2}, \frac{\tau\pi_{1a}^2}{4|\lambda|},  \frac{\epsilon\tau^2\pi_{1a}^2}{4|\lambda|} \Big\}$.
        \end{itemize}
    \end{enumerate}
\end{theorem}

\begin{remark}
    It is possible to remove the dependency of $\lambda$ in the sample complexity results by using Lemma \ref{lem:lambda_max} that tells us about the maximum $\lambda$, and study the worst-case sample complexity bounds. However, keeping $\lambda$-dependent results here highlights the (indirect) polynomial dependency on the unfairness through the correction factor $\lambda$. This also indicates that threshold estimation is non-uniform with respect to the underlying unfairness: the number of samples required may change drastically depending on the degree of correction and the relative group proportions in the distribution.
\end{remark}

\subsection{Bayes Error Estimation for Cost-Sensitive Classification}

Prior works \citep{ishidaperformance, ushio2025practical, jeong2024data} study an estimator for the Bayes error of a distribution for binary classification, by leveraging a simple observation: the Bayes error (which we can denote with $\beta$) can be written down as $\beta = \expecto{X}{\min(\eta(x), 1 - \eta(x))}$. They also show that $\hat{\beta} = \frac{1}{N}\underset{i \in [N]}{\sum}\min(\eta_i, 1 - \eta_i)$ is a consistent estimator of the Bayes risk. In the previous section, we derived the expression for cost-sensitive Bayes risk and the corresponding Bayes optimal classifier (Lemma \ref{lem:scott_def}). Let $\beta_{CS}$ denote the Bayes error for a distribution with respect to a fixed cost-sensitive risk, with group-dependent costs $\alpha_a$. As an independent result, and in a similar spirit to Proposition $3.1$ of Ishida et al. \cite{ishidaperformance}, we now derive the consistency and sample-complexity properties of an estimator for the cost-sensitive Bayes risk.

\begin{theorem} \label{thm:beta_cs_cons}
    Consider the Bayes optimal classifier for the cost-sensitive risk with group-dependent costs $\alpha_a$:
    \begin{align*}
        g^*_\alpha(x,a) = \begin{cases}
        1, & \text{if } \eta(x,a) \geq \alpha_a \\
        0, & \text{otherwise.}
      \end{cases}
    \end{align*}
    \begin{enumerate}
    \setlength{\itemsep}{0pt}
    \setlength{\parsep}{0pt}
    \setlength{\parskip}{0pt}  
        \item Bayes Risk: The Bayes error with respect to the cost-sensitive risk can be expressed as $\beta_{CS} = \expecto{X,A}{\min((1 - \alpha_A)\eta(X,A), \alpha_A(1 - \eta(X,A)))}$. 
        \item Consistency: $\hat{\beta}_{CS} = \frac{1}{N_\beta}\underset{i \in [N_\beta]}{\sum}\min((1 - \alpha_i)c_i, \alpha_i(1 - c_i)) \overset{p}{\rightarrow} \beta_{CS}$.
        \item Sample Complexity: For any $\delta > 0$, with probability at least $1 - \delta$, $|\hat{\beta}_{CS} - \beta_{CS}| \leq \frac{1}{4}\sqrt{\frac{1}{N_\beta} \log(\frac{2}{\delta})}$.
    \end{enumerate}
\end{theorem}

While this result is geared toward group-aware cost-sensitive risks, it provides a roadmap for deriving results similar to prior work \citep{ishidaperformance, ushio2025practical, jeong2024data} and for generalizing their estimators for Bayes error estimation in cost-sensitive and complex metric applications \citep{koyejo2014consistent, singh2022optimal}. We will now proceed to specialize Theorem \ref{thm:beta_cs_cons} towards fairness applications and construct finite sample estimators using soft labels.

With the fact $\hat{\alpha}_a^+ \overset{p}{\rightarrow} \alpha_a$ established in Theorem \ref{thm:alpha_cons}, let $\hat{\beta}_{CS, \hat{\alpha}_a^+} = \frac{1}{N_\beta}\underset{i \in [D_\beta]}{\sum}\min((1 - \hat{\alpha}_a^+)c_i, \hat{\alpha}_a^+(1 - c_i))$ denote the empirical estimator of the cost-sensitive Bayes error with the empirical estimation of cost-sensitive thresholds $\alpha_a$. We now show that $\hat{\beta}_{CS, \hat{\alpha}} \overset{p}{\rightarrow} \beta_{CS}$.

\begin{theorem} \label{thm:bayes_cs_hat_cons}
    Suppose the dataset $D_\alpha$ with $N_\alpha$ samples is used to construct a consistent estimator of the cost-sensitive threshold $\alpha_a$ (Theorem \ref{thm:alpha_cons}). Then, given a dataset $D_\beta$ with $N_\beta$ samples, and assuming $2 - \frac{(2a -1)\lambda}{\pi_{1a}} \geq \tau > 0$ (denominator bounded away from zero), the following statements are true:
    \begin{enumerate}
    \setlength{\itemsep}{0pt}
    \setlength{\parsep}{0pt}
    \setlength{\parskip}{0pt}  
        \item Consistency: $\hat{\beta}_{CS, \hat{\alpha}} = \frac{1}{N_\beta}\underset{i \in [D_\beta]}{\sum}\min((1 - \hat{\alpha}_{a_i}^+)c_i, \hat{\alpha}_{a_i}^+(1 - c_i)) \overset{p}{\rightarrow} \beta_{CS}$. 
        \item Bias: For a fixed $\delta \in (0,1)$:
        \begin{itemize}[leftmargin=*]
        \setlength{\itemsep}{0pt}
        \setlength{\parsep}{0pt}
        \setlength{\parskip}{0pt}
            \item Demographic Parity: $|\beta_{CS} - \expecto{D_{\beta}, D_\alpha}{\hat{\beta}_{CS, \hat{\alpha}}}| = \mathcal{O}\left( |\lambda|\cdot \sqrt{\frac{\log(1/\delta)}{N_\alpha}} \right)$,
            \item  Equal Opportunity: $|\beta_{CS} - \expecto{D_{\beta}, D_\alpha}{\hat{\beta}_{CS, \hat{\alpha}}}| = \mathcal{O}\left(\frac{|\lambda|}{\tau^2} \sqrt{\frac{\log(1/\delta)}{N_\alpha}}\right)$.
        \end{itemize}
        \item Sample Complexity: With $1 - \delta$ probability, $|\beta_{CS} - \hat{\beta}_{CS, \hat{\alpha}}| < \epsilon$ as long as $N_\beta \geq \frac{1}{8\epsilon^2} \log(\frac{4}{\delta})$ and:
        \begin{itemize}[leftmargin=*]
        \setlength{\itemsep}{0pt}
        \setlength{\parsep}{0pt}
        \setlength{\parskip}{0pt}
            \item Demographic Parity: $N_\alpha = \mathcal{O}\left(\frac{\lambda^2}{\epsilon^2} \log\left(\frac{1}{\delta}\right)\right)$,
            \item Equal Opportunity: $N_\alpha = \mathcal{O}\left( \frac{\lambda^2}{\tau^4\epsilon^2}\log\left(\frac{1}{\delta}\right) \right)$.
        \end{itemize}
    \end{enumerate}
\end{theorem}

Unlike Bayes error estimation \citep{ishidaperformance, jeong2024data, ushio2025practical} where $\alpha_a=1/2$, due to distribution and unfairness dependent $\alpha_a$, we now have biased estimators and sufficient sample requirement that depends on the unfairness correction $\lambda$, and for $\tau$ that determines how well our denominator is defined.

\subsection{Expected Error Estimation}

However, when considering the fairness-accuracy tradeoff, we are interested in computing the error rate with respect to the expected $0$-$1$ loss. Recall from Definition \ref{def:risk_disparity} the definition of expected risk with respect to the $0$-$1$ loss: $R(g) = \underset{a \in \{0,1\}}{\sum} \pi_a\Big(\expecto{X|A=a}{(1 - 2\eta(X,a))g(X,a)}\Big) + \pi 
= \expecto{X, A}{(1 - 2C)\id{C \geq \alpha_A}} + \pi$. A finite sample estimator of the above risk can be written down as follows: $\hat{R}(\hat{g}) = \frac{1}{N_{\beta}} \underset{i \in [D_{\beta}]}{\sum} (1 - 2c_i)\id{c_i \geq \hat{\alpha}_{a_i}^+} + \hat{\pi}$.




\begin{theorem} \label{thm:err_01_cons}
    Consider $\hat{R}(\hat{g})$, the empirical estimator for the expected risk from Definition \ref{def:risk_disparity}. Then, the following properties hold:
    \begin{enumerate}
    \setlength{\itemsep}{0pt}
    \setlength{\parsep}{0pt}
    \setlength{\parskip}{0pt}
        \item Consistency: $\hat{R}(\hat{g}) \overset{p}{\rightarrow} R(g)$.
        \item Bias: For a fixed $\delta \in (0,1)$:
        \begin{itemize}[leftmargin=*]
        \setlength{\itemsep}{0pt}
        \setlength{\parsep}{0pt}
        \setlength{\parskip}{0pt}
            \item Demographic Parity: $|R(g) - \expecto{D_\alpha, D_\beta}{\hat{R}(\hat{g})}| =  \mathcal{O}\left(|\lambda|\cdot\max\{L_0, L_1\}\cdot \sqrt{\frac{\log(1/\delta)}{N_\alpha}} \right)$,
            \item Equal Opportunity: $|R(g) - \expecto{D_\alpha, D_\beta}{\hat{R}(\hat{g})}| =  \mathcal{O}\left(\frac{|\lambda| \max\{L_0, L_1\}}{\tau^2} \sqrt{\frac{\log(1/\delta)}{N_\alpha}}\right)$.
        \end{itemize}
        \item Sample Complexity: With $1 - \delta$ probability, $|R(g) - \hat{R}(\hat{g})| < \epsilon$ as long as $N_\beta \geq \frac{18}{\epsilon^2} \log\Big(\frac{6}{\delta}\Big)$ and:
        \begin{itemize}[leftmargin=*]
        \setlength{\itemsep}{0pt}
        \setlength{\parsep}{0pt}
        \setlength{\parskip}{0pt}
            \item Demographic Parity: $N_\alpha = \mathcal{O}\left(\frac{\lambda^2\max\{L_0^2, L_1^2\}}{\epsilon^2} \log\left(\frac{1}{\delta}\right)\right)$,
            \item Equal Opportunity: $N_\alpha = \mathcal{O}\left( \frac{\lambda^2\max\{L_0^2, L_1^2\}}{\epsilon^2\tau^4} \log\left(\frac{1}{\delta}\right) \right).$
        \end{itemize}
    \end{enumerate}
\end{theorem}

Note that estimating the expectation using the sample set $D_\beta$ is standard. The difficult part is that our underlying group-dependent thresholds are estimates with respect to $D_\alpha$. This makes estimating risk under constraints more difficult and introduces nontrivial differences from prior results on risk estimation with finite samples \citep{ishidaperformance, jeong2024data, ushio2025practical}.

\subsection{Unfairness Estimation}

We will now estimate unfairness using a finite number of soft labels. Since every fairness metric would require different distribution-dependent quantities, we will consider demographic parity and equal opportunity metrics here. 
To proceed with estimating the unfairness rates, we will need estimators for the True positive rates (TPR) and Positive rates (PR) later in the paper. We will first tackle TPR, and PR can be tackled similarly. A natural estimator for the TPR is to replace all quantities involved with their sample estimates, as done in Jeong et al. \cite{jeong2024data}. Let $\rho_{\mathrm{TPR}, a}(g_\alpha)$ denote the true positive rate for a threshold classifier $g_\alpha(x,a) = \id{\eta(x,a) \geq \alpha_a}$ on group $A=a$, which can be expressed as: $\rho_{\mathrm{TPR}, a}(g_\alpha) = \prob{g_{\alpha}(X, A)=1|Y=1, A=a}= \frac{\expecto{X|A=a}{\id{\eta(X, a) \geq \alpha_{a}} \eta(X, a)}}{\pi_{1|a}}$, and its sample estimate: $\hat{\rho}_{\mathrm{TPR},a}(g_{\hat{\alpha}_a^+}) = \frac{\frac{1}{N_{\beta,a}} \sum_{i \in [D_{\beta, a}]} (c_i(\id{c_i \geq \hat{\alpha}_a^+})}{\hat{\pi}_{1|a}^+}$, 
where $N_{\beta, a}$ are the number of samples in $D_\beta$ from group $A=a$, and $\hat{\alpha}_a^+$ is the threshold constructed using finite samples.



Once we have our estimators for the TPR, we can define the equal opportunity difference of a classifier $g$ as $\Delta_{\mathrm{EOP}}(g) = |\rho_{\mathrm{TPR},1}(g) - \rho_{\mathrm{TPR},0}(g)|$ and its empirical counterpart $\hat{\Delta}_{\mathrm{EOP}}(g) = |\hat{\rho}_{\mathrm{TPR},1}(g) - \hat{\rho}_{\mathrm{TPR},0}(g)|$. This defines $\Delta_{\mathrm{EOP}}(g_\alpha)$ as the true equal opportunity difference with respect to the true threshold $\alpha_a$ and $\hat{\Delta}_{\mathrm{EOP}}(g_{\hat{\alpha}})$ as the empirical estimate of equal opportunity with respect to the empirical estimate of the threshold. 

If we only care about the positive rate (for demographic parity), we can write: $\rho_{\mathrm{PR},a}(g_\alpha) = \prob{g_\alpha(X, a) = 1|A=a} = \expecto{X|A=a}{\id{\eta(X, a) \geq \alpha_a}}$, and its finite sample estimator: $\hat{\rho}_{\mathrm{PR},a}(g_{\hat{\alpha}_a^+}) = \frac{1}{N_{\beta, a}}  \sum_{i \in [D_{\beta, a}]} \id{c_i \geq \hat{\alpha}_a^+}$. Similar to EOP, we can write the DP difference as $\Delta_{DP}(g) = |\rho_{\mathrm{PR}, 1}(g) - \rho_{\mathrm{PR},0}(g)|$. Then, the true DP difference on the true threshold is $\Delta_{DP}(g_\alpha)$ and the empirical DP difference on the estimated threshold is $\hat{\Delta}_{DP}(g_{\hat{\alpha}})$.



\begin{theorem} \label{thm:beta_unf_cons}
Suppose a dataset $D_\beta$ is used to construct $\hat{\Delta}(g_{\hat{\alpha}_a^+})$, where $\hat{\alpha}_a^+$ is the estimator for the DP or the EOP threshold (Theorem \ref{thm:alpha_cons}). Then the following holds:
\begin{enumerate}
    \setlength{\itemsep}{0pt}
    \setlength{\parsep}{0pt}
    \setlength{\parskip}{0pt}
    \item Consistency: $\hat{\Delta}(g_{\hat{\alpha}_a^+}) \overset{p}{\rightarrow} \Delta(g_\alpha)$.
    \item Bias: For $\delta \in (0,1)$, 
    \begin{itemize}[leftmargin=*]
    \setlength{\itemsep}{0pt}
    \setlength{\parsep}{0pt}
    \setlength{\parskip}{0pt}
        \item DP: $|\Delta_{DP}(g) - \expecto{D_{\beta}, D_\alpha}{\hat{\Delta}_{DP}(\hat{g})}| = \mathcal{O} \left( \frac{1}{\sqrt{N_{\beta}}} +  |\lambda|\cdot\max\{L_0, L_1\} \sqrt{\frac{\log(1/\delta_a)}{N_\alpha}}\right)$.
        \item EOP: $|\Delta_{EOP}(g) - \expecto{D_{\beta}, D_\alpha}{\hat{\Delta}_{EOP}(\hat{g})}| = \mathcal{O}\Bigg( \frac{|\lambda|\cdot\max\{L_0, L_1\}}{\tau^2}   \sqrt{\frac{\log(1/\delta)}{N_\alpha}} + \frac{1}{\sqrt{N_{\beta}}} \Bigg)$.
    \end{itemize}
    \item Sample Complexity: With $1 - \delta$ probability, $|\hat{\Delta}(\hat{g}) - \Delta(g)| \leq \epsilon$, as long as $N_{\beta} = \mathcal{O}\left( \frac{1}{\epsilon^2} \log\left( \frac{1}{\delta} \right)\right)$ and:
    \begin{itemize}[leftmargin=*]
    \setlength{\itemsep}{0pt}
    \setlength{\parsep}{0pt}
    \setlength{\parskip}{0pt}
        \item DP: $N_\alpha = \mathcal{O}\left( \frac{\lambda^2\cdot\max\{L_0^2, L_1^2\}}{\epsilon^2} \log\left( \frac{1}{\delta} \right)\right)$,
        \item EOP:  $N_{\alpha} = \mathcal{O}\left( \frac{\lambda^2\max\{L_0^2, L_1^2\}}{\epsilon^2\tau^4}\log\left(\frac{1}{\delta}\right) \right)$.
    \end{itemize}
\end{enumerate}
\end{theorem}

Just like the bias and sample complexity results in Theorem \ref{thm:err_01_cons}, the rates of unfairness estimation depend sharply on the unfairness correction and our smoothness assumptions, apart from standard rates \citep{mohri2018foundations}. For readability, we subsume the dependence on prior group and subgroup probabilities ($\pi_a$ and $\pi_{ya}$) within the order notation. However, the proofs in the Appendix highlight this dependence explicitly, which is useful to study rare and imbalanced group behaviors. We provide a proof sketch, along with the proofs in the Appendix.

\textbf{Tradeoff Estimation}: We use Algorithm \ref{algo:pareto} (details in the Appendix) to estimate the Pareto frontier using the estimators derived using soft labels in the previous sections. The tradeoff algorithm appeals to the monotonicity and convexity of the Pareto frontier (Lemma \ref{lem:monotone-lambda} and Proposition $4.6$ in \cite{zeng2024bayes}) to set up a bisection search. However, since the guarantees for our estimators in previous sections are only derived for a fixed but arbitrary $\lambda$, we below note a couple of approaches to derive high-probability guarantees on the estimated Pareto frontier. 

\begin{remark}[On the consistency of the estimated frontier]
\label{rem:frontier-consistency}
Our aim is to estimate the fairness-accuracy tradeoff, estimated pointwise in $\lambda$, rather than the risk at a prescribed unfairness level $\delta$: the guarantees of Theorem~\ref{thm:alpha_cons} (and Theorems~\ref{thm:err_01_cons},~\ref{thm:beta_unf_cons}) are stated for a fixed $\lambda$. However, we can provide a high-probability guarantee on the `correctness' of the tradeoff: (1) Fixed grid: If the $\lambda$ values are fixed on a grid $\{\lambda_1,\dots,\lambda_K\}\subset[\lambda_{\min},\lambda_{\max}]$, the search points are data independent, and a union bound over the $K$ points controls the error; (2) Bisection: In its current state, Algorithm~\ref{algo:pareto} selects each $\lambda$ from the estimated disparity $\hat{\Delta}$, which makes the points data-dependent, and a union bound over them is not directly valid. This can be rectified by studying the \emph{uniform-in-$\lambda$} guarantee, $\sup_{\lambda}|\hat{\Delta}(h_\lambda)-\Delta(h_\lambda)|\to 0$. 
Since $\{h_\lambda\}$ is a low-complexity, one-parameter monotone threshold family, such a bound may be available via a uniform law of large numbers. Bisection is more query-efficient, but its analysis requires controlling the error uniformly in $\lambda$.
\end{remark}

Note that Algorithm \ref{algo:pareto} can not only be used to generate an estimate of the Pareto frontier, but can also be used to approximate the risk of the classifier when we desire disparity at most $\delta$. This is because we can always interpolate the curve between the obtained points to get a sense of the sharpness of the tradeoff, along with the bias and sample-complexity bounds in Theorems \ref{thm:err_01_cons} and \ref{thm:beta_unf_cons}. This can be readily used to enforce fairness and choose between fairness metrics, among many applications.

\section{Experiments}

We now present empirical verification of the estimated tradeoff. We work with different sources of soft labels: a synthetic Gaussian setting where we can analytically compute the soft labels \citep{zeng2022bayes}, an average of hard labels on the CIFAR-10H dataset \citep{peterson2019human}, directly borrowed from prior work, \citep{ishidaperformance, ushio2025practical} and noisy soft labels derived from a logistic regression model trained on the Adult dataset \citep{lichman2013uci}, a popular dataset for unfairness evaluation \citep{barocas2023fairness}. Further details of the experiments are provided in the Appendix.

We compare our computed tradeoff against several algorithms that return a tradeoff. We choose the `Exp-Grad' algorithm \cite{agarwal2018reductions} due to its theoretical and practical strengths \citep{sharma2023comparing, ghosh2023fair}, Oxonfair \citep{delaney2024oxonfair}, Fairret \citep{buyl2023fairret}, and the MIFPO algorithm \citep{kozdoba2024efficient}. 
Oxonfair \citep{delaney2024oxonfair} uses validation data to optimize classification thresholds of a pretrained predictor and supports a variety of fairness metrics. MIFPO (Model-Independent Fairness-Performance Optimization) \citep{kozdoba2024efficient} proposes a discrete optimization method to compute the fairness-accuracy tradeoff using soft scores from calibrated classifiers. 
Finally, Fairret \citep{buyl2023fairret} proposes differentiable regularizers as proxies for any linear-fractional form of fairness metrics \citep{celis2019classification}. All methods return a tradeoff for both fairness metrics, except MIFPO \citep{kozdoba2024efficient}, which returns a tradeoff only for DP. For a fair comparison, we ensure that MIFPO uses the same soft scores we do, rather than recomputing them. Note that our estimated tradeoff curve relies on bisection search and hence concentrates more around the unfairness value of $0$, which is our desired target unfairness. 

\begin{figure*}
    \centering
    
    
    
    \begin{subfigure}{0.45\linewidth}
        \includegraphics[width=\linewidth]{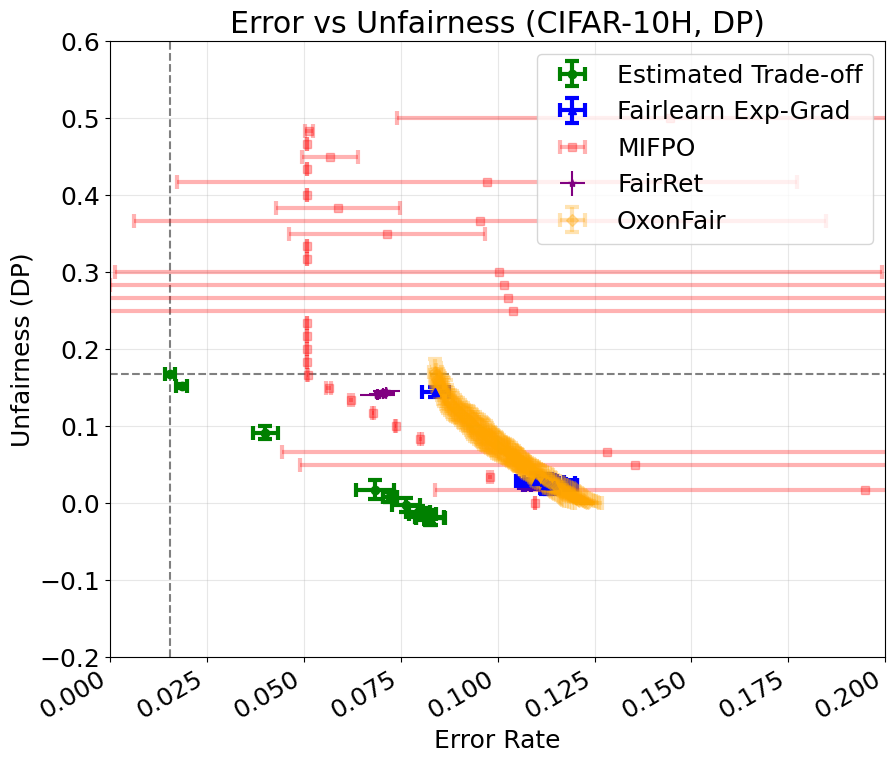}
    \end{subfigure}
    \begin{subfigure}{0.45\linewidth}
        \includegraphics[width=\linewidth]{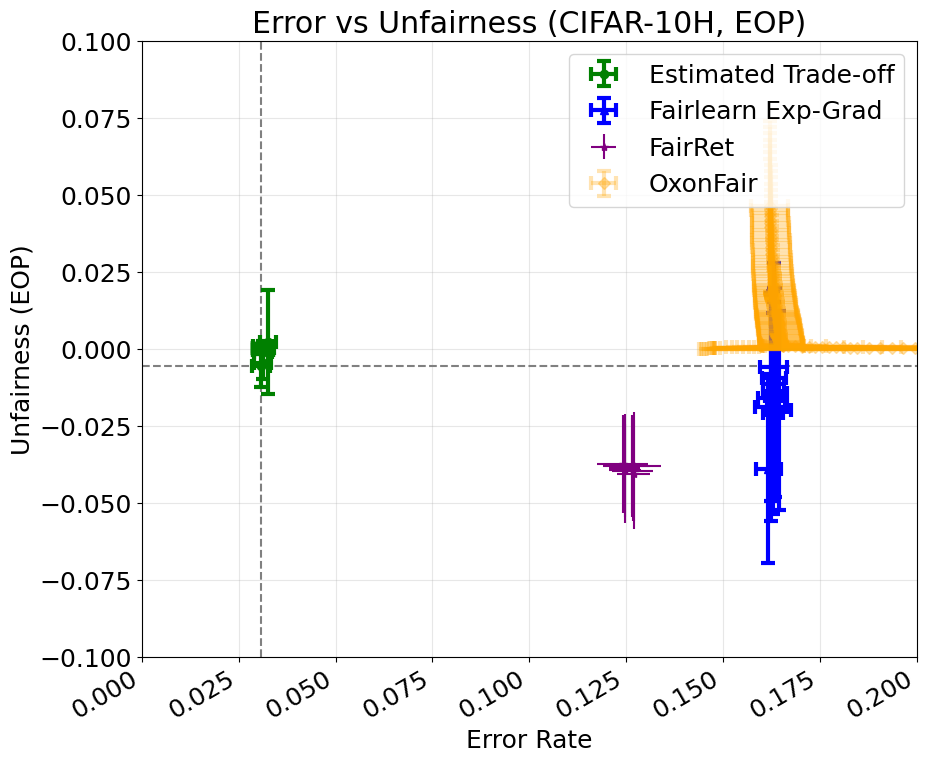}
    \end{subfigure}

    \vspace{0.5em} 
    
    \begin{subfigure}{0.45\linewidth}
        \includegraphics[width=\linewidth]{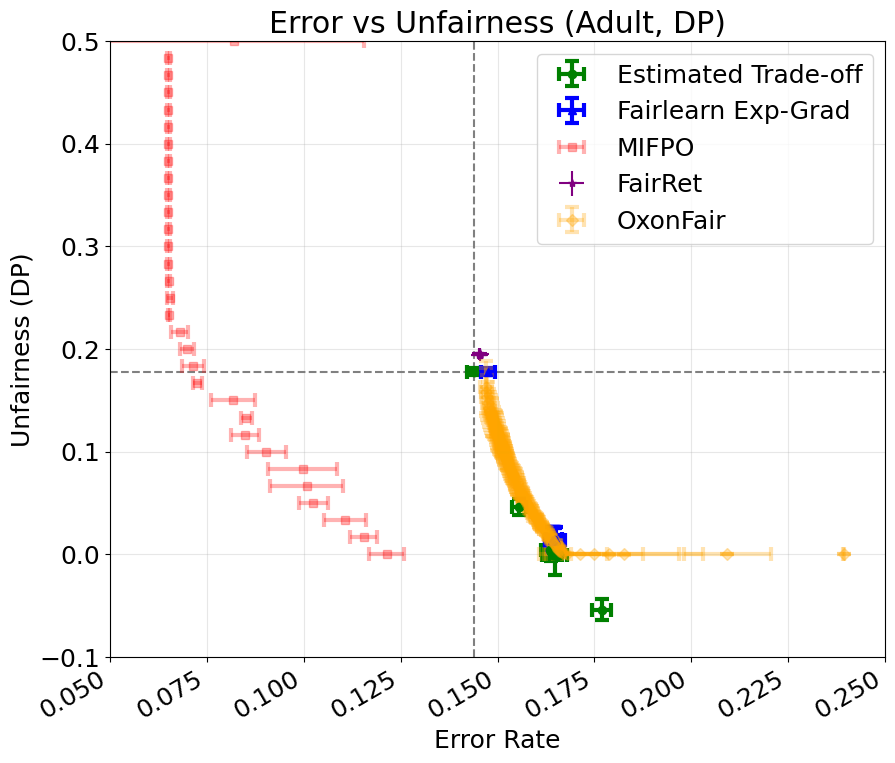}
    \end{subfigure}
    \begin{subfigure}{0.45\linewidth}
        \includegraphics[width=\linewidth]{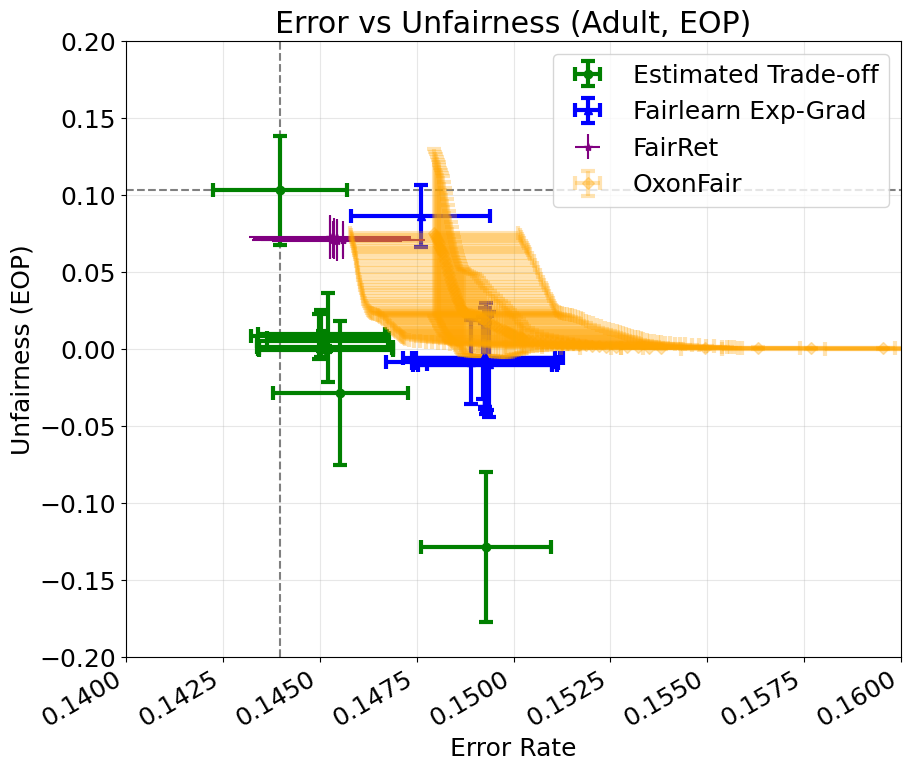}
    \end{subfigure}

    \caption{Unfairness-error tradeoffs on Cifar-10H and the Adult Dataset, for Demographic Parity (DP) and Equal Opportunity (EOP) constraints. The proposed estimate (green) establishes a lower bound against competing algorithms, except MIFPO (red), which computes a high-variance data-dependent lower bound different from the Bayes optimal. Dotted lines denote the unconstrained Bayes Optimal classifier at $\lambda = 0$.}
    \label{fig:siximages}
\end{figure*}

In Figures \ref{fig:introimage} and \ref{fig:siximages}, we show the results for the Gaussian, CIFAR-10H, and the Adult datasets. Averaged across 5 runs, we see that our estimate of the Pareto frontier consistently lower bounds those of all other methods and exhibits the least variance. A peculiar observation is from the demographic parity results, where MIFPO seems to be a lower bound for the method in almost all cases. This is where we would like to highlight how MIFPO operates. Given a dataset, MIFPO aims to optimize for a lower bound that would almost always lower bound any Pareto frontier obtained on the data (Section $4$ in \cite{kozdoba2024efficient}). However, for MIFPO, there are no guarantees that we are optimizing for the Bayes optimal fairness-accuracy tradeoff. The MIFPO algorithm does not optimize the true tradeoff curve and instead relies on convex programming to solve for a proxy objective. Hence, comparing algorithms against MIFPO may always yield a large gap, and such comparisons are not indicative of the true gap. 

Finally, note that our method at $\lambda=0$ corresponds to the Bayes error, as indicated by the intersection of the dotted black lines in Figures \ref{fig:introimage} and \ref{fig:siximages}. Note that our plots also indicate that many fairness tradeoff estimation techniques fail to remain close to the Bayes error when their objectives don't require fairness-constraining solutions. This might indicate that these algorithms are overfitting, similar to the phenomena observed in prior works on Bayes error estimation \citep{ishidaperformance, ushio2025practical}.


\section{Conclusion}

In this work, we set out to extend Bayes error estimation techniques towards estimating the fairness-accuracy Pareto frontier. We do so by generalizing the instance-free soft-label estimation techniques from prior work \citep{ishidaperformance, jeong2024data, ushio2025practical} for cost-sensitive risks. Since fairness constraints themselves can be treated as a cost-sensitive learning problem, and given the properties of the fairness-accuracy Pareto front, we can set up consistent estimators for the fair thresholds and for risk and unfairness estimates. Empirically, we can also lower-bound many recent algorithms to obtain the fairness-accuracy frontier.

Our guarantees show that, even with clean soft scores, the problem remains challenging due to strong dependence on the distribution's priors and the unfairness-dependent term $\lambda$. Recent work generalizes Bayes error estimation with soft labels to the case of monotonic noise in clean soft scores  \cite{ushio2025practical}, and extending our estimators to such noisy setups and beyond is an immediate future work.
One can also study other models of weak supervision in the Bayes optimal settings, such as positive-confidence (PC) data \citep{ishida2018binary}, and positive-unlabeled (PU) data \citep{du2014analysis}. Generalizing to these settings has attracted attention \cite{wu2022fairness, ishidaperformance}, and fine-tuning our theoretical framework for them is an interesting future direction. To move towards group-blind and sensitive attribute-unaware estimation, we can either use group-blind optimal characterizations (Proposition E.1 in \cite{chzhen2019leveraging}) or probabilistic predictors of sensitive attributes \cite{diana2022multiaccurate, prost2021measuring, bharti2023estimating, bharti2025multiaccuracy}, at the cost of increased sample complexity of the underlying estimators.
Finally, our guarantees and approach also serve as a template to approach multi-objective problems where the standard Bayes optimal classifier may not align with the optimal predictor for complex metrics \citep{koyejo2014consistent, cotter2019optimization, singh2022optimal} and broader applications \cite{sukenik2025generalization, wegel2026sample}. 

\section*{Acknowledgments}

The authors would like to thank the reviewers for a positive and constructive exchange during the review process. Takashi Ishida was supported by KAKENHI Grant Number 22K17946. Rajiv Ratn Shah would like to thank IIIT Delhi, where he has formerly been an associate professor. Masashi Sugiyama was supported by JST ASPIRE Grant Number JPMJAP25B1.

\newpage

\bibliography{main}
\bibliographystyle{unsrt}


\appendix

\section{Notations and Assumptions}


Our data domain is defined over $\mathcal{X} \times \mathcal{Y} \times \mathcal{A}$, where $\mathcal{X} \subseteq \mathbb{R}^d, \mathcal{Y} = \{0,1\} \text{ and } \mathcal{A} = \{ 0,1 \}$. We denote the posterior probability of the positive class, given a sample, as $\eta(x, a) = \prob{Y=1|X=x, A=a}$. We define the soft label for an instance $(x_i, a_i)$ as posterior probability: $c_i = \eta(x_i, a_i)$, and let $C=\eta(X,A)$ denote a random variable. Let $\pi_a = \prob{A=a}$, $\pi_{ya} = \prob{Y=y, A=a}$, and $\pi_{y|a} = \prob{Y=y|A=a}$. By the law of total probability, $\prob{Y=y} = \underset{a \in \mathcal{A}}{\sum} \pi_{ya}$. We assume the existence of all groups and classes, i.e., $\pi_a > 0, \pi_{y|a} >0$ for all $y \in \{0,1\}$ and for all $a \in \mathcal{A}$.

Whenever we work with finite sample approximations, we assume that we are given access to $N$ i.i.d. soft labels of the form $(c_i, a_i)$, and we will denote our dataset with $D = \{(c_i, a_i)\}_{i=1}^N$. Since our approach is model-free, we do not require access to instances or features $x_i$. We will also partition our dataset $D$ into two parts: $D = D_\alpha \cup D_\beta$, with $D_\alpha \cap D_\beta = \varnothing$, where $D_\alpha$ (with $N_\alpha$ samples) will be the data used to estimate the thresholds, and $D_\beta$ (with $N_\beta$ samples) will be used to construct the estimators of the required expectation, with $N = N_\alpha + N_\beta$. The splits are chosen independently of the sample values. Furthermore, we will denote with $N_{\alpha, a}$ (or $N_{\beta, a}$) the number of samples from group $A=a$. We now assume the following about our soft labels:

\begin{assumption}(No interior atoms / Continuity of the CDF of $\eta$) 
    The mapping $t \rightarrow F_a(t) = \prob{\eta(X, a) \leq t|A=a}, \forall a \in \mathcal{A}$ is continuous in $(0,1)$. Moreover, $\prob{\eta(X,a) \geq \nicefrac{1}{2}| A=a} > 0$ for all $a \in \{0,1\}$.
\end{assumption}

For later results, we would also need a bounded density assumption on our group-conditioned soft labels.

\begin{assumption}(Lipschitz continuous group CDF) 
    For each $a \in \{0,1\}$, the group-conditional distribution of $C=\eta(X,a) | A=a$ admits a density $f_a$ on $(0,1)$ such that $\underset{t \in (0,1)}{\sup} f_a(t) \leq L_a < \infty$. In particular, $F_a$ is $L_a$-Lipschitz: $|F_a(u) - F_a(v)| \leq L_a\cdot |u - v|$ for all $u,v \in (0,1)$.
\end{assumption}

Let $\hat{\pi}_a = \frac{1}{N_\alpha} \sum_{i \in [D_\alpha]} \id{a_i = a}$ denote our estimator for $\pi_a$. By the Weak Law of Large Numbers, $\hat{\pi}_a \overset{p}{\rightarrow} \pi_a$. Furthermore, the bias of the estimator $\hat{\pi}_a$ is zero, which is easy to see via the linearity of expectation. We can derive a similar estimator for $\pi = \prob{Y=1}$ as $\hat{\pi} = \frac{1}{N_\alpha} \sum_{i \in [D_\alpha]} c_i$.

Similarly, let $\hat{\pi}_{1a} = \frac{1}{N_\alpha} \sum_{i \in [D_\alpha]} \id{a_i = a}c_i$ denote our estimator for $\pi_{1a}$. By the Weak Law of Large Numbers, $\hat{\pi}_{1a} \overset{p}{\rightarrow} \pi_{1a}$. Furthermore, the bias of the estimator $\hat{\pi}_{1a}$ is zero, which is easy to see via the linearity of expectation. The same analysis works out for the unbiased estimator $\hat{\pi}_{1|a} = \frac{1}{N_{\alpha,a}} \sum_{i \in [D_\alpha]} \id{a_i = a}c_i$ and its expected value $\pi_{1|a}$. Since we will work with functions of $\hat{\pi}_a$ (and other estimators) that may have inverses of $\hat{\pi}_a$, we also define another estimator $\hat{\pi}^+_a = \max\{\hat{\pi}_a, \frac{1}{N_\alpha}\}$, which is still statistically consistent, but a biased estimator (same for $\hat{\pi}^+_{1a}$ and $\hat\pi_{1|a}^+$).

We would also require estimators for the True positive rates (TPR) and Positive rates (PR) later in the paper. We will first tackle TPR, and PR can be tackled similarly. A natural estimator for the TPR is to replace all quantities involved with their sample estimates, as done in \citet{jeong2024data}. Let $\rho_{\mathrm{TPR}, a}(g_\alpha)$ denote the true positive rate for a threshold classifier $g_\alpha(x,a) = \id{\eta(x,a) \geq \alpha_a}$ on group $A=a$. 

\begin{align*}
    \rho_{\mathrm{TPR}, a}(g_\alpha) = \prob{g_\alpha=1|Y=1, A=a} &= \frac{\expecto{X|A=a}{ \id{\eta(x,a) \geq \alpha_a}\cdot\prob{Y=1|X=x, A=a}}}{\pi_{1|a}} \\
    &= \frac{\expecto{X|A=a}{\id{C \geq \alpha_a}\cdot C}}{\pi_{1|a}}.
\end{align*}

We can now define a sample estimate as:


\begin{align*}
    \hat{\rho}_{\mathrm{TPR},a}(g_{\hat{\alpha}_a^+}) = \frac{\frac{1}{N_{\beta,a}} \sum_{i \in [D_{\beta, a}]} (c_i(\id{c_i \geq \hat{\alpha}_a^+})}{\hat{\pi}_{1|a}^+}, 
\end{align*}

where $N_{\beta, a}$ are the number of samples in $D_\beta$ from group $A=a$, and $\hat{\alpha}$ is the threshold constructed using finite samples or intermediate estimators using finite samples (for example, thresholds for fairness constraints derived in Lemma \ref{lem:fair-threshold}). 



Once we have our estimators for the TPR, we can define the equal opportunity difference of a classifier $g$ as $\Delta_{\mathrm{TPR}}(g) = |\rho_{\mathrm{TPR},1}(g) - \rho_{\mathrm{TPR},0}(g)|$ and its empirical counterpart $\hat{\Delta}_{\mathrm{TPR}}(g) = |\hat{\rho}_{\mathrm{TPR},1}(g) - \hat{\rho}_{\mathrm{TPR},0}(g)|$. This defines $\Delta_{\mathrm{TPR}}(g_\alpha)$ as the true equal opportunity difference with respect to the true threshold $\alpha_a$ and $\hat{\Delta}_{\mathrm{TPR}}(g_{\hat{\alpha}})$ as the empirical estimate of equal opportunity with respect to the empirical estimate of the threshold. 

If we only care about the positive rate (for ensuring demographic parity), we can come up with a natural estimator:

\begin{align*}
    \rho_{\mathrm{PR},a}(g_\alpha) = \prob{g_\alpha = 1|A=a} = \expecto{X|A=a}{\id{c \geq \alpha_a}},
\end{align*}

and its finite sample estimate:

\begin{align*}
    \hat{\rho}_{\mathrm{PR},a}(g_{\hat{\alpha}_a^+}) = \frac{1}{N_{\beta, a}}  \sum_{i \in [D_{\beta, a}]} \id{c_i \geq \hat{\alpha}_a^+}.
\end{align*}

Similar to the equal opportunity difference, we can write the demographic parity difference as $\Delta_{DP}(g) = |\rho_{\mathrm{PR}, 1}(g) - \rho_{\mathrm{PR},0}(g)|$. Then, the true DP difference on the true threshold is $\Delta_{DP}(g_\alpha)$ and the empirical DP difference on the estimated threshold is $\hat{\Delta}_{DP}(g_{\hat{\alpha}})$.

\section{Helper Results}

\begin{lemma} \label{lem:pi_hoeff}
    (Finite sample bound for estimation of $\pi$) Let $X_1, X_2, ..., X_N \in [0,1]$ be $N$ i.i.d. random variables with mean $\pi$. Let $\hat{\pi} = \frac{1}{N} \sum_{i \in [N]} X_i$.  For any $\epsilon > 0$, if  $N \geq \frac{1}{2\epsilon^2}\log(\frac{2}{\delta})$, then with $1 - \delta$ probability over the choice of the data sample,  $\hat{\pi} \in (\pi - \epsilon, \pi + \epsilon)$.
\end{lemma}

\begin{proof}
    The proof follows directly from the application of Hoeffding's inequality\footnote{\href{https://cs229.stanford.edu/extra-notes/hoeffding.pdf}{stanford.edu}}.
\end{proof}

\begin{lemma} \label{lem:var_upper_bound}
    (Upper bound on Mean Absolute Error) Consider an estimator $\hat{\pi}$ constructed using $N$ samples for some quantity $\pi$ (Lemma \ref{lem:pi_hoeff}). If $\hat{\pi}$ is an unbiased estimate of $\pi$, then $\expecto{D_N}{|\hat{\pi} - \pi|} \leq \sqrt{Var(\hat{\pi})}$. 
\end{lemma}

\begin{proof}
    Let $Z=\hat{\pi} - \pi$. Since $\hat{\pi}$ is an unbiased estimate of $\pi$, $\expecto{D_N}{Z} = 0$, $Var(Z) = \expecto{D_N}{Z^2} = \expecto{D_N}{|Z|^2} = Var(\hat{\pi})$. Observe that by Jensen's inequality, $\expecto{}{|Z|^2} \geq (\expecto{}{|Z|})^2 \implies \expecto{}{|Z|} \leq \sqrt{\expecto{}{|Z|^2}}$, Hence $\expecto{D_N}{|\hat{\pi} - \pi|} \leq \sqrt{Var(\hat{\pi})}$.
\end{proof}

\section{Proofs for Section \ref{sec:prelim}}

Let $g_{\lambda} = \underset{g}{\text{argmin}}~ R(g) + \lambda \cdot \Delta(g)$, for $\lambda \in \mathbb{R}$ denote the primal solution for a fixed $\lambda$. We can also dub this the $\lambda$-fair classifier, since this can potentially be $\delta(\lambda)$-fair classifier. Using this $\lambda$, we can also write down the Bayes optimal $\lambda$-fair classifier using the threshold characterization in Lemma \ref{lem:fair-threshold}, which will be useful later in this paper. We now introduce a key lemma \cite{zeng2024bayes} that enables us to study the fairness-accuracy Pareto frontier.


\begin{lemma} (Proposition 4.1 from \citep{zeng2024bayes}) \label{lem:monotone-lambda}
    As a function of $\lambda$, $\Delta(g_\lambda)$ is monotone non increasing, i.e. $\lambda_1 < \lambda_2 \implies \Delta(g_{\lambda_1}) \geq \Delta(g_{\lambda_2})$. The risk $R(g_\lambda)$ is monotone non-increasing on $(-\infty, 0)$ and monotone non-decreasing on $(0, \infty)$.
\end{lemma}

Lemma \ref{lem:monotone-lambda} is incredibly helpful to devise an algorithm to estimate the tradeoff, since it allows us to parametrize the curve with $\lambda$, establish the convexity of the tradeoff curve (Proposition $4.6$ in \cite{zeng2024bayes}), and hence enables the use of bisection search to get close to any level of unfairness required.
We will now attempt to bound the range of $\lambda$. Along with this information on endpoints and the monotonicity properties of $\lambda$ from Lemma \ref{lem:monotone-lambda}, this would allow us to set up a bisection search algorithm. 


\begin{lemma} (Lemma \ref{lem:lambda_max} in the main text)
    Assuming $w(., 0) < 0$ and $w(.,1) > 0$ (Definition \ref{def:linear-disparity}), when $Dis(g_0) > 0, \lambda_{\max} = \frac{1}{\underset{(x,a)}{\inf} |w(x,a)|}$, so that $\lambda \in [0, \lambda_{\max}]$ is sufficient to interpolate the entire Pareto frontier, including the $\lambda$ for exact fairness $\lambda^*(0)$. And when $Dis(g_0) < 0, \lambda_{\min} =  \frac{-1}{\underset{(x,a)}{\inf} |w(x,a)|}$, so that $\lambda  \in [\lambda_{\min}, 0]$, is sufficient to interpolate the entire Pareto frontier. 
\end{lemma}


\begin{proof}
    To construct our bounds on $\lambda$, we will try to study the case of exact fairness to obtain the worst possible bounds. Assume WLOG that $Dis(g_0) > 0$. The proof for $Dis(g_0) < 0$ is similar. 

    For demographic parity and equal opportunity fairness constraints (Definition \ref{def:linear-disparity}), $w(., 0) < 0$ and $w(.,1) > 0$. Consider the threshold classifier for groups $A=0$ and $A=1$ at any $\lambda$ (Lemma \ref{lem:fair-threshold}):

    \begin{align*}
        g_\lambda(x,0) = \id{\eta(x,0) \geq \frac{1}{2} - \frac{\lambda}{2} |w(x,0)|} \quad \text{and} \quad g_\lambda(x,1) = \id{\eta(x,1) \geq \frac{1}{2} + \frac{\lambda}{2} |w(x,1)|}.
    \end{align*}

    Also, by assumption, disparity is positive at $\lambda=0$, where disparity is defined as:

    \begin{align*}
        Dis(g_\lambda) =  \pi_1 \int_\mathcal{X}  |w(x,1)|\cdot g_\lambda(x,1) d\mathbb{P}_{X|A=1}(x) - \pi_0 \int_\mathcal{X}  |w(x,0)|\cdot g_\lambda(x,0) d\mathbb{P}_{X|A=0}(x) > 0.
    \end{align*}

     From Lemma \ref{lem:monotone-lambda}, we can see that as $\lambda$ increases, the threshold for group $A=1$, $t_1$ increases (since $w(.,1) > 0$) and $t_0$ decreases (since $w(.,0) < 0)$. If $t_1 > 1$, then the $g(x,1) = 0,  \forall x$, since $\eta(x,1) \in [0,1]$. Similarly, if $t_0 \leq 0$, $g(x,0) = 1,  \forall x$. 

     At that point, the disparity $Dis(g_\lambda)$ at $\lambda=0$, which was positive to begin with, becomes negative, and hence we cross over the point of zero unfairness.  At such a trivial classification point, $R(g_\lambda) > 0$ and $Dis(g_\lambda) < 0$, which is the minimum possible disparity since $Dis(g) \in [-1,1]$. 



    It remains to check the value of $\lambda$ at which the group-aware classifier $g$ becomes trivial for each group. Let such a $\lambda$ be denoted with $\lambda_{\max}$. Such a $\lambda_{\max}$ occurs at:

    \begin{align*}
        & \text{(A=1): } \frac{1}{2} + \lambda\frac{|w(x,1)|}{2} > 1 \implies \lambda > \frac{1}{\inf_x |w(x, 1)|}, \text{ and } \\
        & \text{(A=0): } \frac{1}{2} - \lambda\frac{|w(x,0)|}{2} \leq 0 \implies \lambda \geq \frac{1}{\inf_x |w(x, 0)|}\\
        & \text{Overall: } \implies \lambda > \frac{1}{\inf_x |w(x, a)|}.
    \end{align*}

    Now, note that due to Lemma \ref{lem:monotone-lambda}, the tradeoff function parametrized by $\lambda$ is convex, and by increasing $\lambda$ towards $\lambda_{\max}$ (the disparity point of $-1$), we would have also crossed the zero disparity point for some $\lambda \in [0, \lambda_{\max}]$. Hence $\lambda \in [0, \lambda_{\max}]$ whenever the Bayes optimal classifier has positive disparity ($Dis(g_0) > 0$). A similar argument follows for the other case when $Dis(g_0) < 0$.
\end{proof}

This condition becomes much simpler for the equal opportunity and the demographic parity fairness metrics: $\lambda_{\max} = \underset{a \in \mathcal{A}}{\max}~ \pi_{1a}$ and $\lambda_{\max} = \underset{a \in \mathcal{A}}{\max}~ \pi_a$, respectively. Restricting the possible range of $\lambda$ allows us to set up a bisection search to estimate the fairness-accuracy tradeoff (Algorithm \ref{algo:pareto}). We now argue that approximating this tradeoff curve pointwise is sufficient for our scenario to obtain a high-probability estimate of the true Pareto frontier. 

\begin{definition}
Given a risk measure $R(\cdot)$ and a fairness disparity measure $\Delta(\cdot)$, a classifier $g'$ Pareto-dominates another classifier $g$, if $R(g') < R(g)$ and $\abs{\Delta(g')} < \abs{\Delta(g)}$. The Pareto frontier is the set of those classifiers that are not Pareto-dominated by any other classifier.
\end{definition}

Using a well-known characterization of the Pareto frontier by \citep{varian1976two}, we can equivalently write the above set as $g_{\lambda} = \underset{g}{\text{argmin}}~ R(g) + \lambda \cdot \Delta(g)$. We now show that any point on the Pareto frontier will be covered by any curve that we generate by varying the values of $\lambda$ specified by the range obtained from Lemma \ref{lem:lambda_max}.

\begin{proposition}
(Proposition \ref{prop:pareto_proof} in the main text) If a classifier $g$ belongs to the Pareto-frontier, then $g = g_{\lambda}$, for some $\lambda \in \mathbb{R}$.   
\end{proposition}

\begin{proof}
Suppose $g$ belongs to the Pareto-frontier and WLOG assume $\Delta(g) \geq 0$. We know from Proposition $4.6$ in \cite{zeng2024bayes} that the Pareto front is convex.

Since $g$ lies on the Pareto-frontier, then $(R(g), \Delta(g))$ must lie on the boundary of the above convex set. Otherwise, if $(R(g), \Delta(g))$ lies in the interior, there exists a small neighborhood around $(R(g), \Delta(g))$ that is contained in this convex set, and would include $g'$ such that $R(g') < R(g)$ and $0 < \Delta(g') < \Delta(g)$. Now consider a supporting tangent to the convex set at $(R(g), \Delta(g))$ defined by a normal vector $(\alpha, \beta) \in \mathbb{R}^{2}$ and margin/shift $\gamma \in \mathbb{R}$ as follows: $\alpha R(h) + \beta \Delta(h) \geq \gamma$, for all classifiers $h$, and $\alpha R(g) + \beta \Delta(g) = \gamma$. Let $\lambda = \beta/\alpha$. Then $R(g) + \lambda \Delta(g) = R(g) + \beta/\alpha \cdot \Delta(g) = (\alpha R(g) + \beta \Delta(g))/\alpha = \gamma/\alpha$, whereas for any other $h$, we have $R(h) + \lambda \Delta(h) = R(h) + \beta/\alpha \cdot \Delta(h) = (\alpha R(h) + \beta \Delta(h))/\alpha \geq \gamma/\alpha$. Therefore, $g$ is a minimizer of $R(h) + \lambda g(h)$ objective over all $h$, and is itself $g_{\lambda}$ for $\lambda = \beta/\alpha$.

\end{proof}

We will show that any $g_\lambda$ can be constructed as a threshold classifier, where the threshold is a function of some distribution-dependent parameters and $\lambda$. And then using Lemma \ref{lem:lambda_max}, we will be able to estimate the risk and unfairness of several points on the Pareto frontier by varying the value of $\lambda$ in a fixed range.

\textbf{Cost sensitive classification}: In classification, we generally work with the $0$-$1$ loss. However, there can be scenarios where we need to weigh different errors unequally, such as the cost of missing a disease (it may be much riskier not to catch a disease than to falsely detect it in healthy individuals) and the cost of granting loans to undeserving candidates. This is the setting of cost-sensitive classification \citep{elkan2001foundations}.

\begin{lemma} \label{lem:scott_def}
    (Extension of \cite{scott2013classification} to the case of subgroups) Let $\alpha \in [0,1]$. Define the $\alpha$-CS risk as weighing the false negatives for a group $A=a$ with $(1 - \alpha_a)$ and the false positives with $\alpha_a$:
    \begin{align*}
        R_\alpha(g) &= \underset{a \in \mathcal{A}}{\sum} \Bigg( (1 - \alpha_a)\pi_{1a}\cdot\expecto{X|Y=1, A=a}{l_{01}(g(X,a))} + \alpha_a\pi_{0a}\cdot\expecto{X|Y=0, A=a}{l_{01}(g(X,a))} \Bigg)
    \end{align*}
\end{lemma}

where $\pi_{ya} = \prob{Y=y, A=a}, y \in \{0,1\}$ and $l_{01}$ is the $0$-$1$ loss that returns $1$ whenever the prediction and the true label mismatch. Then, the cost-sensitive group-dependent Bayes optimal classifier can be written down as: 
\begin{align*}
    g_\alpha(x,a) = \begin{cases}
    1, & \text{if } \eta(x,a) \geq \alpha_a \\
    0, & \text{otherwise }
  \end{cases}
\end{align*}

The proof for Lemma \ref{lem:scott_def} is a simple group-aware extension of Scott et al. \cite{scott2013classification}. We can immediately see that the thresholds for Bayes optimal fair classifiers from Lemma \ref{lem:fair-threshold} can be written down as group-dependent costs \citep{menon2018cost}. For example, for demographic parity, $\alpha_a = \frac{1}{2} + \frac{(2a - 1)\lambda^*(\delta)}{2\pi_a}$. We will exploit Lemmas \ref{lem:lambda_max} and \ref{lem:scott_def} in the coming sections to construct estimators for thresholds, risk, and disparity at arbitrary points up until $\lambda_{\max}$ to get an efficient estimate of the Pareto frontier, only using soft labels.

\section{Proofs for Section \ref{sec:estimation}}

\textbf{Proof Sketch: } We summarize the main ideas behind the proofs of the DP and EOP threshold (Theorem \ref{thm:alpha_cons}), and the risk and unfairness estimators (Theorems \ref{thm:bayes_cs_hat_cons}, \ref{thm:err_01_cons}, \ref{thm:beta_unf_cons}).

\begin{itemize}
    \item \textbf{Consistency}: 
    The consistency proofs begin with the Weak Law of Large Numbers for the empirical priors
    $\hat{\pi}$, $\hat{\pi}_a$, $\hat{\pi}_{ya}$, and $\hat{\pi}_{y\mid a}$.
    The threshold estimators $\hat{\alpha}_a$ are then shown to be consistent by applying the
    Continuous Mapping Theorem \footnote{\href{https://www.stat.cmu.edu/~arinaldo/Teaching/36752/S18/Scribed_Lectures/Apr5.pdf}{stat.cmu.edu}} to the corresponding plug-in formulas, together with the truncation
    operator $\Pi_{[0,1]}$.
    In the EOP case, we additionally require the relevant denominator to stay bounded away from zero.
    The later consistency results for risk and unfairness then follow by combining the consistency of
    $\hat{\alpha}_a$ with continuity of the corresponding functionals, along with standard concentration
    of empirical averages over the independent sample split.

    \item \textbf{Bias bounds}:
    For each estimator, we define a suitable good event $\mathcal{G}$ on which the empirical prior of interest is close to its population counterpart.
    On $\mathcal{G}$, the estimator error is bounded deterministically in terms of the prior estimation error, using elementary algebraic control of reciprocals whenever needed.
    Taking expectations then reduces to controlling quantities such as
    $\mathbb{E}[|\hat{\pi}-\pi|]$, which are bounded using explicit variance calculations and
    standard inequalities for bounded random variables.
    On the complement $\mathcal{G}^C$, the estimator error is controlled by the uniform boundedness of the estimator, multiplied by $\Pr(\mathcal{G}^C)$, and $\Pr(\mathcal{G}^C)$ is bounded using Hoeffding's inequality.
    For the later risk and unfairness results, the threshold estimation error is propagated through
    the corresponding risk or disparity functional using continuity or Lipschitz bounds under
    Assumptions \ref{assump: eta-continous} and \ref{assump: eta-density}.

    \item \textbf{Sample complexity}:
    The sample complexity results are obtained by combining concentration bounds for the threshold
    estimators built from $D_\alpha$ with concentration bounds for the empirical risk and unfairness
    estimators built from the independent split $D_\beta$.
    The total failure probability is then controlled by a union bound.
    For the later results, the final sample requirements inherit both the threshold-estimation complexity and the Lipschitz dependence of the downstream risk or disparity estimators on the estimated thresholds.
\end{itemize}


We now provide proofs for all of the results described in Section \ref{sec:estimation}. For convenience and readability, we will prove the DP and EOP results separately whenever applicable. 

\begin{theorem} (DP results from Theorem \ref{thm:alpha_cons} in the main text)
    (Consistency and Sample Complexity for DP threshold) Given a dataset $D_\alpha$ of clean soft labels ($c_i = \eta(x_i, a_i)$) and a fixed $\lambda \in \mathbb{R}$, for the case of Demographic Parity, consider the following estimator of the threshold: $\hat{\alpha}_a^+ = \Pi(\Tilde{\alpha}_a^+), \text{ where } \Tilde{\alpha}_a^+ = \frac{1}{2} + \frac{(2a - 1)\lambda}{2\hat{\pi}^+_a}, ~\forall a \in \{0,1\}$. Then the following holds:
    \begin{enumerate}
        \item Consistency: $\hat{\alpha}^+_a \overset{p}{\rightarrow} \alpha_a$.
        \item Bias: $|\alpha_a -\expecto{D_\alpha}{\hat{\alpha}_a^+}| \leq \frac{|
        \lambda|}{\pi_a }\sqrt{\frac{1 - \pi_a}{\pi_a N_\alpha}} + 2\exp(-\frac{\pi_a^2}{2}N_\alpha)$.
        \item Sample Complexity: With $1 - \delta$ probability and  $N_\alpha > \max\{ \frac{2}{\pi_a^2} \log{\frac{4}{\delta}}, \frac{\lambda^2}{2\epsilon^2\pi_a^4}\log\frac{4}{\delta}, \frac{2}{\pi_a} \}$,  $\prob{|\hat{\alpha}_a ^+- \alpha_a| > \epsilon}\leq \delta$.
    \end{enumerate}
\end{theorem}

\begin{proof}
    We will prove the results for an arbitrary $a \in \{0,1\}$. Due to Lemma \ref{lem:lambda_max} and our projection operation $\Pi$ on the empirical estimator, $\alpha_a$ and $\hat{\alpha}_a^+$ are both restricted to the interval $[0,1]$. Using this, note that in our domain of $\lambda \in [0,\lambda_{\max}]$ or $\lambda \in [-\lambda_{\min}, 0]$, $\alpha_a = \Pi(\alpha_a)$, which is a useful property we exploit throughout the proof.

    (1) \textbf{Consistency}: By the weak law of large numbers, $\hat{\pi}_a \overset{p}{\rightarrow} \pi_a$. Consider $\hat{\pi}_a^+$.  Since $\pi_a > 0$ by assumption, we can pick a sufficiently large $N_\alpha$ such that $1/N_\alpha \leq \pi_a/2$. Therefore the event $\{\hat{\pi}_a < 1/N_\alpha\}$ is contained in the event $\{\hat{\pi}_a < \pi_a/2\}$ which is further contained in $\{|\hat{\pi}_a - \pi_a| \geq \pi_a/2\}$. Since $\hat{\pi}_a \overset{p}{\rightarrow} \pi_a$, $\prob{|\hat{\pi}_a - \pi_a| \geq \pi_a/2} \rightarrow 0$ and hence $\prob{\hat{\pi}_a^+ \neq \hat{\pi}_a} \rightarrow 0$, and hence $\hat{\pi}^+_a\overset{p}{\rightarrow} \pi_a$.
    
    
    The estimator $\hat{\alpha}^+_a$ is a continuous function of $\hat{\pi}^+_a$ (including the truncation operation). We can use the Continuous Mapping Theorem \footnote{\href{https://www.stat.cmu.edu/~arinaldo/Teaching/36752/S18/Scribed_Lectures/Apr5.pdf}{stat.cmu.edu}} to claim that $\hat{\pi}^+_a \overset{p}{\rightarrow} \pi_a \implies \hat{\alpha}^+_a \overset{p}{\rightarrow} \alpha_a$.

    (2) \textbf{Bias}:  We will now proceed towards bounding the bias of the threshold for an arbitrary group $A=a$. We can break down our bias as follows: $|\alpha_a - \expecto{D_\alpha}{\hat{\alpha}^+_a}| = |\expecto{D_\alpha}{\alpha_a - \hat{\alpha}^+_a}| = |\expecto{D_\alpha}{(\alpha_a - \hat{\alpha}^+_a)\mathbbm{1}_\mathcal{G}} + \expecto{D_\alpha}{(\alpha_a - \hat{\alpha}^+_a)\mathbbm{1}_{\mathcal{G}^C}}| \leq |\expecto{D_\alpha}{(\alpha_a - \hat{\alpha}^+_a)\mathbbm{1}_\mathcal{G}}| + |\expecto{D_\alpha}{(\alpha_a - \hat{\alpha}^+_a)\mathbbm{1}_{\mathcal{G}^C}}|$, where $\mathcal{G}$ is defined next.
    
    Let $\mathcal{G} = \{ |\hat{\pi}_a - \pi_a| \leq \epsilon \}$ be the good event that our estimate of $\pi_a$ is close. On this event, for sufficiently large $N_\alpha$ ($N_\alpha > \frac{2}{\pi_a}$), $\hat{\pi}^+_a = \hat{\pi}_a$ and hence $\hat{\alpha}^+_a = \hat{\alpha}_a$. Using Hoeffding's inequality, we know that $\hat{\pi}_a \in (\pi_a - \epsilon, \pi_a + \epsilon)$ with $1 - \delta/2$ probability over the choice of data sample $D_\alpha$. We can set $\epsilon \leq \pi_a/2$, which means $N_\alpha > \frac{2}{\pi_a^2}\log(\frac{4}{\delta})$.

    Consider the first term:

    \begin{align*}
        |\expecto{D_\alpha}{(\alpha_a - \hat{\alpha}_a)\mathbbm{1}_\mathcal{G}}| &\leq \expecto{D_\alpha}{|\alpha_a - \hat{\alpha}_a|\mathbbm{1}_\mathcal{G}} = \expecto{D_\alpha}{|\Pi(\alpha_a) - \Pi(\Tilde{\alpha}_a)|\mathbbm{1}_\mathcal{G}} \\
        &\leq \expecto{D_\alpha}{|\alpha_a - \Tilde{\alpha}_a|\mathbbm{1}_\mathcal{G}} ~~(\text{Since $\Pi$ is Lipschitz})\\
        &= \expecto{D_\alpha}{\left|\frac{\lambda (2a - 1)}{2\pi_a} - \frac{\lambda(2a - 1)}{2\hat{\pi}_a}\right|\mathbbm{1}_\mathcal{G}} \\
        & \leq \frac{|\lambda|}{2}\cdot\expecto{D_\alpha}{\left|\frac{1}{\pi_a} - \frac{1}{\hat{{\pi}}_a}\right|\mathbbm{1}_\mathcal{G}} \\
        &= \frac{|\lambda|}{2}\cdot\expecto{D_\alpha}{\left|\frac{\hat{\pi}_a - \pi_a}{\pi_a \hat{\pi}_a}\right|\mathbbm{1}_\mathcal{G}} = \frac{|\lambda|}{2}\cdot\expecto{D_\alpha}{\left|\frac{\hat{\pi}_a - \pi_a}{\pi_a \hat{\pi}_a}\right|\mathbbm{1}_\mathcal{G}} \\
        &= \frac{|\lambda|}{2}\cdot\expecto{D_\alpha}{\frac{|\hat{\pi}_a - \pi_a|}{\pi_a \hat{\pi}_a}\mathbbm{1}_\mathcal{G}} \text{ (since $\pi_a, \hat{\pi}_a > 0$}) \\
        &\leq \frac{|
        \lambda|}{\pi_a^2} \expecto{D_\alpha}{|\hat{\pi}_a - \pi_a|\mathbbm{1}_\mathcal{G}} \text{ (On } \mathcal{G}, \frac{1}{\pi_a\hat{\pi}_a} \leq \frac{2}{\pi_a^2}).
    \end{align*}

    Since $\hat{\pi}_a$ is an unbiased estimate of $\pi_a$, using Lemma \ref{lem:var_upper_bound}, $\expecto{D_\alpha}{|\hat{\pi}_a - \pi_a|} \leq \sqrt{Var(\hat{\pi}_a})$. Using $\expecto{D_\alpha}{|\alpha_a - \hat{\alpha}_a|\mathbbm{1}_\mathcal{G}} \leq \expecto{D_\alpha}{|\alpha_a - \hat{\alpha}_a|} \leq \frac{|\lambda|\sqrt{Var(\hat{\pi}_a)}}{\pi_a^2}$, we get:

    \begin{align*}
        |\expecto{D_\alpha}{(\alpha_a - \hat{\alpha}_a)\mathbbm{1}_\mathcal{G}}| \leq \frac{|\lambda|\sqrt{Var(\hat{\pi}_a)}}{\pi_a^2} = \frac{|
        \lambda|}{\pi_a }\sqrt{\frac{1 - \pi_a}{\pi_a N_\alpha}}
    \end{align*}

    where $Var(\hat{\pi}_a) = Var(\frac{1}{N_{\alpha}}\sum_{i \in [N_{\alpha}]} \id{a_i = a}) = \frac{1}{N_{\alpha}^2} Var(\sum_{i \in [N_{\alpha}]} \id{a_i = a}) = \frac{\pi_a(1 - \pi_a)}{N_\alpha}$.

    On $\mathcal{G}^c$, we can use the fact that both $\alpha_a, \hat{\alpha}_a^+ \in [0,1]$:

    \begin{align*}
        &|\expecto{D_\alpha}{(\alpha_a - \hat{\alpha}^+_a)\mathbbm{1}_{\mathcal{G}^C}}| \leq |\expecto{D_\alpha}{1\cdot\mathbbm{1}_{\mathcal{G}^C}}| \\
        &= \prob{\mathcal{G}^C} \leq 2\exp\left(-\frac{\pi_a^2}{2}N_\alpha\right) \text{ (Using Hoeffding's inequality on } \prob{\mathcal{G^C}}).
    \end{align*}

    Overall, 

    \begin{align*}
        |\alpha_a -\expecto{D_\alpha}{\hat{\alpha}_a^+}| \leq \frac{|
        \lambda|}{\pi_a }\sqrt{\frac{1 - \pi_a}{\pi_a N_\alpha}} + 2\exp\left(-\frac{\pi_a^2}{2}N_\alpha\right).
    \end{align*}

    (3) \textbf{Sample Complexity}: We now want to study the number of samples required for a good estimator of $\alpha_a^+$. We want to ensure that $\prob{|\hat{\alpha}_a^+ - \alpha_a| > \epsilon} \leq \delta$. Note that $\{|\hat{\alpha}_a^+ - \alpha_a| > \epsilon\} \subseteq \{|\Tilde{\alpha}_a^+ - \alpha_a| > \epsilon\}$.
    


    Reusing the analysis from the previous part, we can first ensure that $\prob{\mathcal{G}^C} = \prob{|\pi_a - \hat{\pi}_a| > \pi_a/2} < \delta/2$ by using Hoeffding's inequality to obtain that $N_\alpha > \frac{2}{\pi_a^2} \log{\frac{4}{\delta}}$. 
    
    Next, as stated earlier, on the good event $\mathcal{G}$, as long as $N_\alpha > \frac{2}{\pi_a}$ (which means $1/N_\alpha < \pi_a/2$), $\hat{\pi}_a^+ = \hat{\pi}_a$ and hence $\hat{\alpha}_a^+ = \Tilde{\alpha}_a$. This implies that $|\Tilde{\alpha}_a^+ - \alpha_a| = |\frac{\lambda}{2}|\cdot|\frac{1}{\hat{\pi}_a} - \frac{1}{\pi_a}| \leq \frac{|\lambda|}{\pi_a^2} |\hat{\pi}_a - \pi_a|$.

    Therefore, on $\mathcal{G}$, $\{|\Tilde{\alpha}_a^+ - \alpha_a| > \epsilon\} \subseteq \{ |\hat{\pi}_a - \pi_a| >\frac{\epsilon\pi_a^2}{|\lambda|} \}$. Now, we can study the entire event and upper bound its failure probability:

    \begin{align*}
        &\prob{|\Tilde{\alpha}_a^+ - \alpha_a| > \epsilon} \leq \prob{|\Tilde{\alpha}_a^+ - \alpha_a| > \epsilon ~\cap \mathcal{G}} + \prob{|\Tilde{\alpha}_a^+ - \alpha_a| > \epsilon ~\cap \mathcal{G}^C} \\
        &\leq \prob{|\hat{\pi}_a - \pi_a| >\frac{\epsilon\pi_a^2}{|\lambda|}} + \prob{\mathcal{G}^C} \\
        &\leq 2\exp(-\frac{2N_\alpha\epsilon^2\pi_a^4}{\lambda^2}) + 2\exp(-\frac{N_\alpha\pi_a^2}{2}) \text{ (Using Hoeffding's Inequality and earlier bound on $\mathcal{G}^C$)}
    \end{align*}

    Bounding the second term by $\delta/2$ gave us $N_\alpha > \frac{2}{\pi_a^2} \log{\frac{4}{\delta}}$. Doing the same for the first term gives us $N_\alpha > \frac{\lambda^2}{2\epsilon^2\pi_a^4}\log\frac{4}{\delta}$. Overall, $N_\alpha > \max\{ \frac{2}{\pi_a^2} \log{\frac{4}{\delta}}, \frac{\lambda^2}{2\epsilon^2\pi_a^4}\log\frac{4}{\delta}, \frac{2}{\pi_a} \}$.

\end{proof}

\begin{theorem} (EOP results from Theorem \ref{thm:alpha_cons} in the main text)
    (Consistency and Sample Complexity for EOP threshold) Given a dataset $D_\alpha$ of clean soft labels ($c_i = \eta(x_i, a_i)$) and assuming $2 - \frac{(2a -1)\lambda}{\pi_{1a}} \geq \tau > 0$ (denominator bounded away from zero), for the case of equal opportunity, consider the following estimator of the threshold: $\hat{\alpha}_a^+ = \Pi(\Tilde{\alpha}_a^+), \text{ where } \Tilde{\alpha}_a^+ = \frac{1}{2 - \frac{(2a - 1)\lambda}{\hat{\pi}_{1a}^+}}$. Then the following holds:
    \begin{enumerate}
        \item Consistency: $\hat{\alpha}_a^+ \overset{p}{\rightarrow} \alpha_a$.
        \item Bias: $|\alpha_a -\expecto{D_\alpha}{\hat{\alpha}_a^+}| \leq \frac{4|\lambda|}{\tau^2\pi_{1a}^2}\sqrt{\frac{\pi_{1a}(1 - \pi_{1a})}{N_\alpha}} + 2\exp(-2\epsilon^2N_\alpha), \text{ where } \epsilon \leq \min\{\frac{\pi_{1a}}{2}, \frac{\tau\pi_{1a}^2}{4|\lambda|}\}$.
        \item Sample Complexity: With $1 - \delta$ probability over the choice of data sample $N_\alpha$, as long as $N_\alpha \geq \frac{1}{2t^2} \log(\frac{2}{\delta})$ for $t = \min \Big\{\frac{\pi_{1a}}{2}, \frac{\tau\pi_{1a}^2}{4|\lambda|},  \frac{\epsilon\tau^2\pi_{1a}^2}{4|\lambda|} \Big\}$, $\prob{|\hat{\alpha}_a - \alpha_a| > \epsilon}\leq \delta$.
    \end{enumerate}
\end{theorem}

\begin{proof}
    We will prove the results for an arbitrary $a \in \{0,1\}$. Due to Lemma \ref{lem:lambda_max} and our projection operation $\Pi$ on the empirical estimator, $\alpha_a$ and $\hat{\alpha}_a^+$ are both restricted to the interval $[0,1]$. Using this, note that in our domain of $\lambda \in [0,\lambda_{\max}]$ or $\lambda \in [-\lambda_{\min}, 0]$, $\alpha_a = \Pi(\alpha_a)$, which is a useful property we exploit throughout the proof.

    (1) \textbf{Consistency}: By the weak law of large numbers, $\hat{\pi}_{1a} \overset{p}{\rightarrow} \pi_{1a}$. Consider $\hat{\pi}_{1a}^+$.  Since $\pi_{1a} > 0$ by assumption, we can pick a sufficiently large $N_\alpha$ such that $1/N_\alpha \leq \pi_{1a}/2$. Therefore the event $\{\hat{\pi}_{1a} < 1/N_\alpha\}$ is contained in the event $\{\hat{\pi}_{1a} < \pi_{1a}/2\}$ which is further contained in $\{|\hat{\pi}_{1a} - \pi_{1a}| \geq \pi_{1a}/2\}$. Since $\hat{\pi}_{1a} \overset{p}{\rightarrow} \pi_{1a}$, $\prob{|\hat{\pi}_{1a} - \pi_{1a}| \geq \pi_{1a}/2} \rightarrow 0$ and hence $\prob{\hat{\pi}_{1a}^+ \neq \hat{\pi}_{1a}} \rightarrow 0$, and hence $\hat{\pi}^+_{1a}\overset{p}{\rightarrow} \pi_{1a}$.
    
    Now, we will show consistency in two steps. First, let the denominator be defined as $\Tilde{d}_a = 2 - \frac{(2a - 1)\lambda}{\hat{\pi}_{1a}^+} \text{and } d_a = 2 - \frac{(2a - 1)\lambda}{\pi_{1a}}$. Since $\pi_{1a} > 0$ by assumption, and $\Tilde{\pi}_{1a}^+ \overset{p}{\rightarrow} \pi_{1a}$, we can use the Continuous Mapping Theorem to claim that $\Tilde{d}_a \overset{p}{\rightarrow} d_a$. 

    Again, since $d_a \geq \tau > 0$, then by application of the Continuous Mapping Theorem, $\frac{1}{\Tilde{d}_a} \overset{p}{\rightarrow} \frac{1}{d_a}$ and hence $\hat{\alpha}_a \overset{p}{\rightarrow} \alpha_a$ (as the truncation operation $\Pi$ is also continuous).

    \textbf{(2) Bias}: We want to now bound the bias of the estimator: $|\alpha_a - \expecto{D_\alpha}{\hat{\alpha}_a}|$:

    \begin{align*}
         |\alpha_a - \expecto{D_\alpha}{\hat{\alpha}_a}| &= |\expecto{D_\alpha}{\alpha_a - \hat{\alpha}_a}| \leq \expecto{D_\alpha}{|\alpha_a - \hat{\alpha}_a|} = \expecto{D_\alpha}{|\Pi(\alpha_a) - \Pi(\Tilde{\alpha}_a)|} \\
         &\leq \expecto{D_\alpha}{|\alpha_a - \Tilde{\alpha}_a|} ~~(\text{Since $\Pi$ is Lipschitz}) \\
         &= \expecto{D_\alpha}{\left|\frac{1}{d_a} - \frac{1}{\Tilde{d}_a}\right|} = \expecto{D_\alpha}{\left|\frac{1}{d_a} - \frac{1}{\Tilde{d}_a}\right|\mathbbm{1}_\mathcal{G}} + \expecto{D_\alpha}{\left|\frac{1}{d_a} - \frac{1}{\Tilde{d}_a}\right|\mathbbm{1}_{\mathcal{G}^C}},
    \end{align*}

    where we let $\mathcal{G} = \{ |\hat{\pi}_{1a} - \pi_{1a}| \leq \epsilon \}$ denote the good event that $\hat{\pi}_{1a}$ is close to $\pi_{1a}$. On this event, for sufficiently large $N_\alpha$ ($N_\alpha > \frac{2}{\pi_{1a}}$), $\hat{\pi}^+_{1a} = \hat{\pi}_{1a}$ and hence $\hat{\alpha}^+_a = \hat{\alpha}_a$. 
    We can set $\epsilon \leq \pi_{1a}/2$ so that $\hat{\pi}_{1a} \geq \pi_{1a}/2$.

    On $\mathcal{G}$:

    \begin{align*}
        &|d_a - \Tilde{d}_a| \leq |\lambda|\Big|\frac{\hat{\pi}_{1a} - \pi_{1a}}{\pi_{1a}\hat{\pi}_{1a}}\Big| \leq \frac{2|\lambda|}{\pi_{1a}^2} |\hat{\pi}_{1a} - \pi_{1a}| 
    \end{align*}

    Similarly, we can also bound $\Tilde{d}_a$ on $\mathcal{G}$:

    \begin{align*}
        \Tilde{d}_a \geq d_a - |\Tilde{d}_a - d_a| \geq \tau - \frac{2|\lambda|\epsilon}{\pi_{1a}^2}.
    \end{align*}

    We can ensure that $\Tilde{d}_a \geq \tau/2$ on $\mathcal{G}$, which means $\epsilon \leq \min\{\frac{\pi_{1a}}{2}, \frac{\tau\pi_{1a}^2}{4|\lambda|}\}$. Finally, for the first term of the bias bound, on the good event $\mathcal{G}$:

    \begin{align*}
        \expecto{D_\alpha}{\Big|\frac{1}{d_a} - \frac{1}{\Tilde{d}_a}\Big|\mathbbm{1}_\mathcal{G}} &= \expecto{D_\alpha}{\frac{|\Tilde{d}_a - d_a|}{d_a \Tilde{d}_a}\mathbbm{1}_\mathcal{G}}| \leq \frac{2}{\tau^2}\cdot  \frac{2|\lambda|}{\pi_{1a}^2}\expecto{D_\alpha}{|\hat{\pi}_{1a} - \pi_{1a}|}.
    \end{align*}

    On $\mathcal{G}^c$, we can use the fact that both $\alpha_a, \hat{\alpha}_a^+ \in [0,1]$:

    \begin{align*}
        &\expecto{D_\alpha}{|\alpha_a - \hat{\alpha}^+_a|\mathbbm{1}_{\mathcal{G}^C}} 
        \leq \expecto{D_\alpha}{1\cdot\mathbbm{1}_{\mathcal{G}^C}} \\
        &= \prob{\mathcal{G}^C} \leq 2\exp(-2\epsilon^2N_\alpha) \text{ (Using Hoeffding's inequality on } \prob{\mathcal{G^C}}).
    \end{align*}

    Overall, 

    \begin{align*}
        |\alpha_a -\expecto{D_\alpha}{\hat{\alpha}_a^+}| \leq \frac{4|\lambda|}{\tau^2\pi_{1a}^2}\expecto{D_\alpha}{|\hat{\pi}_{1a} - \pi_{1a}|} + 2\exp(-2\epsilon^2N_\alpha), \text{ where } \epsilon \leq \min\left\{\frac{\pi_{1a}}{2}, \frac{\tau\pi_{1a}^2}{4|\lambda|}\right\}.
    \end{align*}

    To further simplify this, using Lemma \ref{lem:var_upper_bound}, we know that $\expecto{}{|\pi_{1a} - \hat{\pi}_{1a}|} \leq \sqrt{Var(\hat{\pi}_{1a})} = \sqrt{\frac{\pi_{1a}(1 - \pi_{1a})}{N_\alpha}}$, and hence:

    \begin{align*}
        |\alpha_a -\expecto{D_\alpha}{\hat{\alpha}_a^+}| \leq \frac{4|\lambda|}{\tau^2\pi_{1a}^2}\sqrt{\frac{\pi_{1a}(1 - \pi_{1a})}{N_\alpha}} + 2\exp(-2\epsilon^2N_\alpha), \text{ where } \epsilon \leq \min\left\{\frac{\pi_{1a}}{2}, \frac{\tau\pi_{1a}^2}{4|\lambda|}\right\}.
    \end{align*}

    (3) \textbf{Sample Complexity}: For sample complexity, we can proceed by returning a high probability bound on the good event derived in the previous part. Note that $\{|\hat{\alpha}_a^+ - \alpha_a| > \epsilon\} \subseteq \{|\Tilde{\alpha}_a^+ - \alpha_a| > \epsilon\}$.
    
    Recall that $\mathcal{G} = \{ |\hat{\pi}_{1a} - \pi_{1a}| \leq \epsilon \}$ denotes the good event that $\hat{\pi}_{1a}$ is close to $\pi_{1a}$. 
    We can set $\epsilon \leq \pi_{1a}/2$ so that $\hat{\pi}_{1a} \geq \pi_{1a}/2$, whenever $1/N_\alpha \leq \pi_{1a}/2 \implies N_\alpha \geq 2/\pi_{1a}$.

    From the previous part, we know that on $\mathcal{G}$:
    \begin{itemize}
        \item $|d_a - \Tilde{d}_a| \leq \frac{2|\lambda|\epsilon}{\pi_{1a}^2}$,
        \item $\Tilde{d}_a \geq \tau/2$ for $\epsilon \leq \min\{\frac{\pi_{1a}}{2}, \frac{\tau\pi_{1a}^2}{4|\lambda|}\}$, and
        \item Using both of the results above, $|\alpha_a - \tilde{\alpha}_a| \leq \frac{4|\lambda|\epsilon}{\tau^2\pi_{1a}^2}$ on $\mathcal{G}$.
    \end{itemize}
    
    Finally, if we want $|\tilde{\alpha}_a - \alpha_a| \leq \epsilon'$, then $\frac{4|\lambda|\epsilon}{\tau^2\pi_{1a}^2} \leq \epsilon' \implies \epsilon \leq \frac{\epsilon'\tau^2\pi_{1a}^2}{4|\lambda|}$.
    Combining this with the previous $\min$ bound on $\epsilon$, we get $\epsilon = \min \Big\{\frac{\pi_{1a}}{2}, \frac{\tau\pi_{1a}^2}{4|\lambda|},  \frac{\epsilon'\tau^2\pi_{1a}^2}{4|\lambda|} \Big\}$. 

    We have now shown that $\{ |\hat{\pi}_{1a} - \pi_{1a}| \leq \epsilon \} \subseteq \{ |\alpha_a - \hat{\alpha}_a| \leq \epsilon' \}$. Taking complements and probabilities on both sides, we get:

    \begin{align*}
        \prob{|\alpha_a - \hat{\alpha}_a| > \epsilon'} \leq \prob{|\hat{\pi}_{1a} - \pi_{1a}| > \epsilon} \leq 2\exp(-2N_\alpha\epsilon^2) \leq \delta,
    \end{align*}

    which gives us $N_\alpha \geq \frac{1}{2\epsilon^2} \log(\frac{2}{\delta})$ for $\epsilon = \min \Big\{\frac{\pi_{1a}}{2}, \frac{\tau\pi_{1a}^2}{4|\lambda|},  \frac{\epsilon'\tau^2\pi_{1a}^2}{4|\lambda|} \Big\}$. 

    Refactoring notations (using $t$ for $\epsilon$ and $\epsilon$ for $\epsilon'$), with $1 - \delta$ probability over the choice of data sample $N_\alpha$, as long as $N_\alpha \geq \frac{1}{2t^2} \log(\frac{2}{\delta})$ for $t = \min \Big\{\frac{\pi_{1a}}{2}, \frac{\tau\pi_{1a}^2}{4|\lambda|},  \frac{\epsilon\tau^2\pi_{1a}^2}{4|\lambda|} \Big\}$, $\prob{|\hat{\alpha}_a - \alpha_a| > \epsilon}\leq \delta$.

\end{proof}


\begin{theorem} (Theorem \ref{thm:beta_cs_cons} in the main text)
    Consider the Bayes optimal classifier for the cost-sensitive risk with group-dependent costs $\alpha_a$:
    \begin{align*}
        g^*_\alpha(x,a) = \begin{cases}
        1, & \text{if } \eta(x,a) \geq \alpha_a \\
        0, & \text{otherwise.}
      \end{cases}
    \end{align*}
    \begin{enumerate}
        \item Bayes Risk: The Bayes error with respect to the cost-sensitive risk can be expressed as $\beta_{CS} = \expecto{X,A}{\min((1 - \alpha_A)\eta(X,A), \alpha_A(1 - \eta(X,A)))}$. 
        \item Consistency: $\hat{\beta}_{CS} = \frac{1}{N_\beta}\underset{i \in [N_\beta]}{\sum}\min((1 - \alpha_i)c_i, \alpha_i(1 - c_i))$ is a consistent estimator of $\beta_{CS}$, where for the $i^{th}$ instance $(c_i, a_i)$, $c_i = \eta(x_i, a_i)$ and $\alpha_i = \alpha_{a_i}$, the group dependent cost.
        \item Sample Complexity: For any $\delta > 0$, with probability at least $1 - \delta$, $|\hat{\beta}_{CS} - \beta_{CS}| \leq \frac{1}{4}\sqrt{\frac{1}{N_\beta} \log(\frac{2}{\delta})}$.
    \end{enumerate}
\end{theorem}

\begin{proof}
    (1) \textbf{Bayes Risk}: We will adapt the standard proof technique\footnote{\href{https://ocw.mit.edu/courses/18-657-mathematics-of-machine-learning-fall-2015/86f311c7073869c5e0c199008787d5c9_MIT18_657F15_L2.pdf}{ocw.mit.edu}} of the Bayes error for cost-sensitive settings. The expected $\alpha$-cost sensitive risk for the classifier $g^*$ can be expressed as:
    \begin{align*}
        &R_\alpha(g^*) = \underset{a \in \mathcal{A}}{\sum} \Bigg( (1 - \alpha_a)\cdot\expecto{X,Y=1,A=a}{l_{01}(g^*(X,a), 1)} + \alpha_a\cdot\expecto{X,Y=0, A=a}{l_{01}(g^*(X,a), 0)} \Bigg) \\
        &= \underset{a \in \mathcal{A}}{\sum} \expecto{X,a}{\expecto{Y=1|X,a}{(1 - \alpha_a)\id{Y=1,g^*(X,a) = 0}} + \expecto{Y=0|X,a}{\alpha_a\id{Y=0, g^*(X,a)=1}}} \\
        &= \underset{a \in \mathcal{A}}{\sum} \expecto{X,a}{(1 - \alpha_a)\id{\eta(X,a) < \alpha_a}\eta(X,a) + \alpha_a\id{\eta \geq \alpha_a}(1 - \eta(X,a))}
    \end{align*}

    At this point, note that $\eta < \alpha_a \implies (1 - \alpha_a)\eta < \alpha_a(1 - \eta)$ and similarly $\eta \geq \alpha_a \implies \alpha_a(1 - \eta) \leq (1 - \alpha_a)\eta$. Therefore, 

    \begin{align*}
        &R_\alpha(g^*) = \underset{a \in \mathcal{A}}{\sum} \expecto{X,a}{(1 - \alpha_a)\id{\eta(X,a) < \alpha_a}\eta(X,a) + \alpha_a\id{\eta \geq \alpha_a}(1 - \eta(X,a))} \\
        &= \underset{a \in \mathcal{A}}{\sum} \mathrm{E}_{X,a}[\id{\eta(X,a) < \alpha_a}\min((1 - \alpha_a)\eta(X,a), \alpha_a(1 - \eta(X,a))) \\
        &\qquad \qquad+ \id{\eta(X,a) \geq \alpha_a}\min(\alpha_a(1 - \eta(X,a)), \eta(X,a)(1 - \alpha_a))] \\
        &= \expecto{X,A}{\min((1 - \alpha_A)\eta(X,A), \alpha_A(1 - \eta(X,A)))}.
    \end{align*}

    It remains to show that $g^*$ is the Bayes optimal classifier. 
    Let $g'$ be any arbitrary classifier. Then,

    \begin{align*}
        R_\alpha(g') - R_\alpha(g^*) &= \expecto{X,a}{(1 - \alpha_a)\eta(X,a)\id{g'(X,a) = 0} + \alpha_a(1 - \eta(X,a))\id{g'(X,a) = 1}} \\
        &- \expecto{X,a}{(1 - \alpha_a)\eta(X,a)\id{g^*(X,a) = 0} + \alpha_a(1 - \eta(X,a))\id{g^*(X,a) = 1}} \\
        &= \mathrm{E}_{X,a}[(\id{g' = 0} - \id{g^* = 0})(1 - \alpha_a)\eta(X,a) \\
        &\qquad \qquad \qquad \qquad+ (\id{g' = 1} - \id{g^* = 1})\alpha_a(1 - \eta(X,a))] \\
        &= \expecto{X,a}{(\id{g' = 0} - \id{g^* = 0})(\eta(X,a) - \alpha_a)} \\
        &= \expecto{X,a}{\id{g' \neq g^*}\cdot(\eta(X,a) - \alpha_a)\cdot sgn(\eta(X,a) - \alpha_a)}, \\
        &\qquad \qquad \qquad \qquad \qquad \qquad \qquad \qquad\text{ using the fact that $g^* = \id{\eta \geq \alpha_a}$} \\
        &= \expecto{X,a}{\id{g' \neq g^*}\cdot|\eta(X,a) - \alpha_a|} \geq 0.
    \end{align*}

    (2) \textbf{Consistency}:

    \begin{align*}
        \expecto{(c_i, a_i)_{i=1}^{N_\beta}}{\hat{\beta}_{CS}} &= \expecto{(c_i, a_i)_{i=1}^{N_\beta}}{\frac{1}{N_\beta}\underset{i \in [N_\beta]}{\sum}\min((1 - \alpha_i)c_i, \alpha_i(1 - c_i))} \\
        &= \frac{1}{N_\beta}\underset{i \in [N_\beta]}{\sum} \expecto{(c_i, a_i) \sim D}{\min((1 - \alpha_i)c_i, \alpha_i(1 - c_i))} \\
        &= \beta_{CS}.
    \end{align*}

    By the weak law of large numbers, $\hat{\beta}_{CS} \overset{p}{\rightarrow} \beta_{CS}$.

    (3) \textbf{Sample Complexity}: The proof for sample complexity is a straightforward application of Hoeffding's inequality, by noting that $\min((1 - \alpha_i)c_i, \alpha_i(1 - c_i)) \in [0, \nicefrac{1}{4}]$:

    \begin{align*}
        &\prob{|\hat{\beta}_{CS} - \beta_{CS}| \geq \epsilon} \leq 2\exp\Big(\frac{-2N_\beta \epsilon^2}{(1/4)^2}\Big) \leq \delta \\
        & \implies \epsilon \geq\frac{1}{4}\sqrt{\frac{1}{2N_\beta} \log(\frac{2}{\delta})}.
    \end{align*}
\end{proof}


\begin{theorem} (Theorem \ref{thm:bayes_cs_hat_cons} in the main text)
    Suppose the dataset $D_\alpha$ is used to construct a consistent estimator of the cost-sensitive threshold $\alpha_a$ (Theorem \ref{thm:alpha_cons}). Then, given a dataset $D_\beta$ the following statements are true:
    \begin{enumerate}
        \item Consistency: $\hat{\beta}_{CS, \hat{\alpha}} = \frac{1}{N_\beta}\underset{i \in [D_\beta]}{\sum}\min((1 - \hat{\alpha}_{a_i}^+)c_i, \hat{\alpha}_{a_i}^+(1 - c_i))$ is a consistent estimator of $\beta_{CS}$.
        \item Bias: For a fixed $\delta \in (0,1)$:
        \begin{itemize}
            \item Demographic Parity: $|\beta_{CS} - \expecto{D_{\beta}, D_\alpha}{\hat{\beta}_{CS, \hat{\alpha}}}| = \mathcal{O}\left( |\lambda|\cdot \sqrt{\frac{\log(1/\delta)}{N_\alpha}} \right)$,
            \item  Equal Opportunity: $|\beta_{CS} - \expecto{D_{\beta}, D_\alpha}{\hat{\beta}_{CS, \hat{\alpha}}}| = \mathcal{O}\left(\frac{|\lambda|}{\tau^2} \sqrt{\frac{\log(1/\delta)}{N_\alpha}}\right)$.
        \end{itemize}
        \item Sample Complexity: With $1 - \delta$ probability, $|\beta_{CS} - \hat{\beta}_{CS, \hat{\alpha}}| < \epsilon$ as long as $N_\beta \geq \frac{1}{8\epsilon^2} \log(\frac{4}{\delta})$ and:
        \begin{itemize}
            \item Demographic Parity: $N_\alpha = \mathcal{O}\left(\frac{\lambda^2}{\epsilon^2} \log\left(\frac{1}{\delta}\right)\right)$,
            \item Equal Opportunity: $N_\alpha = \mathcal{O}\left( \frac{\lambda^2}{\tau^4\epsilon^2}\log\left(\frac{1}{\delta}\right) \right)$.
        \end{itemize}
    \end{enumerate}
\end{theorem}

\begin{proof} 
    (1) \textbf{Consistency}: For showing consistency, note that conditioned on $D_\alpha$, our thresholds $\hat{\alpha}_a^+$ are fixed, and hence by Theorem \ref{thm:beta_cs_cons}, $\hat{\beta}_{CS, \hat{\alpha}} \overset{p}{\rightarrow} \beta_{CS, \hat{\alpha}}$, where $\beta_{CS, \hat{\alpha}}$ is the cost sensitive Bayes risk with respect to group-aware thresholds $\hat{\alpha}_a^+, \forall a \in \mathcal{A}$. It remains to show that $\beta_{CS, \hat{\alpha}} \overset{p}{\rightarrow} \beta_{CS}$. 

    From Theorem \ref{thm:alpha_cons}, $\hat{\alpha}_a^+ \overset{p}{\rightarrow} \alpha_a$. Note that $f(x) = \expecto{}{min\{(1 - x)\eta, x(1 - \eta)\}}$ is a continuous function of $x$. Hence, by the Continuous Mapping Theorem, $\beta_{CS, \hat{\alpha}} \overset{p}{\rightarrow} \beta_{CS}$.

    (2) \textbf{Bias}: 

    \begin{align*}
         |\beta_{CS} - \expecto{D_{\beta}, D_\alpha}{\hat{\beta}_{CS, \hat{\alpha}}}| &= |\expecto{D_{\beta}, D_\alpha}{\beta_{CS} - \hat{\beta}_{CS, \hat{\alpha}}}| \\
         &= |\expecto{D_\alpha}{\expecto{D_\beta}{(\beta_{CS} - \hat{\beta}_{CS, \hat{\alpha}})| D_\alpha}}| = |\expecto{D_\alpha}{\beta_{CS} - \beta_{CS, \hat{\alpha}}}| \\
         &\leq \expecto{D_\alpha}{|\beta_{CS} - \beta_{CS, \hat{\alpha}}|}. 
    \end{align*}



    At this point, consider the following:

    \begin{align*}
        |\beta_{CS, x} - \beta_{CS, y}| &= |\expecto{X,A}{\min(\eta(1 - x), x(1 - \eta)) - \min(\eta(1 - y), y(1 - \eta))}| \\
        &= |\expecto{}{\frac{\eta + x}{2} - x\eta - |\frac{x - \eta}{2}| - \frac{\eta + y}{2} + y\eta + |\frac{y - \eta}{2}|}| \\
        &(\text{Using } \min(a,b) = \frac{a + b}{2} - |\frac{a - b}{2}|) \\
        &= \frac{1}{2}|\expecto{}{(x - y)(1 - 2\eta) - |\eta - x| + |\eta - y|}| \\
        & = \frac{1}{2} |(x - y)(1 - 2\expecto{}{\eta}) + \expecto{}{|\eta - y| - |\eta - x|}| \\
        & \leq \frac{1}{2} |x - y|\cdot |1 - 2\expecto{}{\eta}| + \frac{1}{2}\expecto{}{||\eta - y| - |\eta - x||} \\
        & \leq \frac{1}{2} |x - y|\cdot |1 - 2\expecto{}{\eta}| + \frac{1}{2}\expecto{}{|x - y|} (\text{Using Reverse Triangle Inequality}) \\
        &= \frac{1}{2} |1 - 2\expecto{}{\eta}||x - y| + \frac{1}{2}|x - y| = \frac{|x - y|}{2} (1 + |1 - 2\pi|) \\
        &\leq |x - y| (\text{Since } \pi \in [0,1]).
    \end{align*}

    The above result shows that the Bayes risk is Lipschitz in the threshold argument $\alpha$ with $L=1$. This result will be useful later.
    
    
    We know from Theorem \ref{thm:alpha_cons} that the bias of $\hat{\alpha}_a^+$ is bounded: $|\expecto{}{\hat{\alpha}_a^+} - \alpha_a| \leq U_m$, where $U_m$ depends on the fairness metric. Therefore, setting $x = \expecto{D_\alpha}{\hat{\alpha}}$ and $y=\alpha_a$ yields us:

    \begin{align*}
        |\beta_{CS} - \expecto{D_{\beta}, D_\alpha}{\hat{\beta}_{CS, \hat{\alpha}}}| &\leq \expecto{D_\alpha}{|\beta_{CS} - \beta_{CS, \hat{\alpha}}|} \\
        &\leq \pi_0\expecto{D_\alpha}{|\beta_{CS} - \beta_{CS, \hat{\alpha}_0^+}|} + \pi_1\expecto{D_\alpha}{|\beta_{CS} - \beta_{CS, \hat{\alpha}_1^+}|} \\
        &\leq \pi_0\expecto{D_\alpha}{|\alpha_0 - \hat{\alpha}_0^+|} + \pi_1\expecto{D_\alpha}{|\alpha_1 - \hat{\alpha}_1^+|}
    \end{align*}

    From Theorem \ref{thm:alpha_cons}, we know that with $1 - \delta$ probability, $|\hat{\alpha}_a^+ - \alpha_a| < \epsilon$. Using this, we get:


    \begin{align*}
        &|\beta_{CS} - \expecto{D_{\beta}, D_\alpha}{\hat{\beta}_{CS, \hat{\alpha}}}| \leq \pi_0\expecto{D_\alpha}{|\alpha_0 - \hat{\alpha}_0^+|} + \pi_1\expecto{D_\alpha}{|\alpha_1 - \hat{\alpha}_1^+|} \\
        &= \underset{a \in \{0,1\}}{\sum} \pi_a\left(\expecto{D_\alpha}{|\alpha_a -  \hat{\alpha}_a^+|\id{|\alpha_a -  \hat{\alpha}_a^+| < \epsilon}} + \expecto{D_\alpha}{|\alpha_a -  \hat{\alpha}_a^+|\id{|\alpha_a -  \hat{\alpha}_a^+| \geq \epsilon}}\right) \\
        &\leq  \underset{a \in \{0,1\}}{\sum} \pi_a \left(\epsilon\cdot\prob{|\alpha_a -  \hat{\alpha}_a^+| < \epsilon} + 1\cdot\prob{|\alpha_a -  \hat{\alpha}_a^+| \geq \epsilon}\right) ~~\text{ (Since $|\alpha_a -  \hat{\alpha}_a^+| \leq 1$) }\\
        &\leq \underset{a \in \{0,1\}}{\sum} \pi_a(\epsilon_a(N_\alpha, \delta_a) + \delta_a).
    \end{align*}

    where, $\epsilon_a(N_\alpha, \delta_a)$ can de derived from the respective sample complexities in Theorem \ref{thm:alpha_cons}.

    For DP, using the dominant term, $\epsilon_a = \frac{|\lambda|}{\pi_a^2} \sqrt{\frac{\log(4/\delta_a)}{2N_\alpha}}$ and setting $\delta = \max\{\delta_0,\delta_1\}$, giving us:


    \begin{align*}
        |\beta_{CS} - \expecto{D_{\beta}, D_\alpha}{\hat{\beta}_{CS, \hat{\alpha}}}| = \mathcal{O}\left( |\lambda|\cdot \sqrt{\frac{\log(1/\delta)}{N_\alpha}} \right).
    \end{align*}

    For EOP, using the $\epsilon$ dependent term, $\epsilon_a = \frac{|\lambda|}{\tau^2\pi_{1a}^2} \sqrt{\frac{8\log(2/\delta_a)}{N_\alpha}}$ and setting $\delta = \max\{\delta_0,\delta_1\}$, giving us:


    \begin{align*}
        |\beta_{CS} - \expecto{D_{\beta}, D_\alpha}{\hat{\beta}_{CS, \hat{\alpha}}}| = \mathcal{O}\left(\frac{|\lambda|}{\tau^2} \sqrt{\frac{\log(1/\delta)}{N_\alpha}}\right).
    \end{align*}
    
    (3) \textbf{Sample Complexity}: We want to ensure the following:

    \begin{align*}
        \prob{|\beta_{CS} - \hat{\beta}_{CS, \hat{\alpha}}| > \epsilon} \leq \delta.
    \end{align*}

    To do so, consider the following breakdown: $|\beta_{CS} - \hat{\beta}_{CS, \hat{\alpha}}| \leq |\beta_{CS} - \beta_{CS, \hat{\alpha}}| + |\beta_{CS, \hat{\alpha}} - \hat{\beta}_{CS, \hat{\alpha}}|$.

    Consider the random error: $|\beta_{CS, \hat{\alpha}} - \hat{\beta}_{CS, \hat{\alpha}}|$. This will give us a sample complexity lower bound for $N_\beta$. Bounding this is a simple application of Hoeffding's inequality with $X_i = \min\{ (1 - \hat{\alpha}_{a_i}^+)c_i, \hat{\alpha}_{a_i}^+(1 - c_i) \} \in [0,1/4]$:

    \begin{align*}
        \prob{|\beta_{CS, \hat{\alpha}_a} - \hat{\beta}_{CS, \hat{\alpha}}| > \epsilon/2} \leq 2\exp(-8N_\beta\epsilon^2) \leq \delta/2 \implies N_\beta \geq \frac{1}{8\epsilon^2} \log(\frac{4}{\delta}) 
    \end{align*}

    Now consider the error due to threshold estimation. We know from our analysis of bias that $|\beta_{CS, x} - \beta_{CS, y}| \leq |x - y| \implies |\beta_{CS} - \beta_{CS, \hat{\alpha}}| \leq \underset{a \in \{0,1\}}{\sum} \pi_a\cdot|\hat{\alpha}_a^+ - \alpha_a|$. 
    
    From the sample complexity results in Theorem \ref{thm:alpha_cons}, we can derive the following: with $\prob{|\hat{\alpha}_a^+ - \alpha_a| > \epsilon/2} \leq U_m(\epsilon/2) \leq \delta/2$, where $U_m$ depends on the fairness metric we choose. This will provide us with a bound on $N_\alpha \geq N_\alpha(m)$.

    For demographic parity, again using the dominant term: 



    \begin{align*}
        \prob{|\hat{\alpha}_a^+ - \alpha_a| > \epsilon/4\pi_a} \leq \delta/4 \implies N_\alpha \geq \frac{8\lambda^2}{\epsilon^2\pi_a^2} \log\left(\frac{16}{\delta}\right) = \mathcal{O}\left(\frac{\lambda^2}{\epsilon^2} \log\left(\frac{1}{\delta}\right)\right),
    \end{align*}

    For equal opportunity, recall that with $1 - \delta$ probability over the choice of data sample $N_\alpha$, as long as $N_\alpha \geq \frac{1}{2t^2} \log(\frac{2}{\delta})$ for $t = \min \Big\{\frac{\pi_{1a}}{2}, \frac{\tau\pi_{1a}^2}{4|\lambda|},  \frac{\epsilon\tau^2\pi_{1a}^2}{4|\lambda|} \Big\}$, $\prob{|\hat{\alpha}_a - \alpha_a| > \epsilon}\leq \delta$. Using these bounds with our new $\epsilon$ and $\delta$ we get:

    \begin{align*}
        \prob{|\hat{\alpha}_a^+ - \alpha_a| > \epsilon/4\pi_a} \leq \delta/4 &\implies N_\alpha \geq \frac{1}{2}\log\left(\frac{8}{\delta}\right)\max\left\{ \frac{2}{\pi_{1a}}, \frac{4|\lambda|}{\tau\pi_{1a}^2}, \frac{4|\lambda|}{\epsilon\tau^2\pi_{1a}^2} \right\}^2 \\
        &= \mathcal{O}\left( \frac{\lambda^2}{\tau^4\epsilon^2}\log\left(\frac{1}{\delta}\right) \right).
    \end{align*}
\end{proof}

\textbf{Proof for Theorem \ref{thm:err_01_cons}}: Recall from Definition \ref{def:risk_disparity} the definition of expected risk:

\begin{align*}
    R(g) &= \underset{a \in \{0,1\}}{\sum} \pi_a\cdot\Big(\expecto{X|A=a}{(1 - 2\eta(X,a))g(X,a)}\Big) + \pi \\
    &= \expecto{X, A}{(1 - 2C)\id{C \geq \alpha_A}} + \pi
\end{align*}

A finite sample estimator of the above risk can be written down as follows:


\begin{align*}
    \hat{R}(\hat{g}) = \frac{1}{N_{\beta}} \underset{i \in [D_{\beta}]}{\sum} (1 - 2c_i)\id{c_i \geq \hat{\alpha}_{a_i}^+} + \hat{\pi}
\end{align*}

\begin{theorem} (Proof for Theorem \ref{thm:err_01_cons} from the main text)
    Consider $\hat{R}(\hat{g})$, the empirical estimator for the expected risk from Definition \ref{def:risk_disparity}. Then, the following properties hold:
    \begin{enumerate}
        \item Consistency: $\hat{R}(\hat{g}) \overset{p}{\rightarrow} R(g)$.
        \item Bias: For a fixed $\delta \in (0,1)$:
        \begin{itemize}
            \item Demographic Parity: $|R(g) - \expecto{D_\alpha, D_\beta}{\hat{R}(\hat{g})}| =  \mathcal{O}\left(|\lambda|\cdot\max\{L_0, L_1\}\cdot \sqrt{\frac{\log(1/\delta)}{N_\alpha}} \right)$,
            \item Equal Opportunity: $|R(g) - \expecto{D_\alpha, D_\beta}{\hat{R}(\hat{g})}| =  \mathcal{O}\left(\frac{|\lambda| \max\{L_0, L_1\}}{\tau^2} \sqrt{\frac{\log(1/\delta)}{N_\alpha}}\right)$.
        \end{itemize}
        \item Sample Complexity: With $1 - \delta$ probability, $|R(g) - \hat{R}(\hat{g})| < \epsilon$ as long as $N_\beta \geq \frac{18}{\epsilon^2} \log\Big(\frac{6}{\delta}\Big)$ and:
        \begin{itemize}
            \item Demographic Parity: $N_\alpha = \mathcal{O}\left(\frac{\lambda^2\max\{L_0^2, L_1^2\}}{\epsilon^2} \log\left(\frac{1}{\delta}\right)\right)$,
            \item Equal Opportunity: $N_\alpha = \mathcal{O}\left( \frac{\lambda^2\max\{L_0^2, L_1^2\}}{\epsilon^2\tau^4} \log\left(\frac{1}{\delta}\right) \right).$
        \end{itemize}
    \end{enumerate}
\end{theorem}

\begin{proof}

    To prove the above results, we will use an intermediate estimator $R(\hat{g}) = \expecto{X, A}{(1 - 2C)\id{C \geq \hat{\alpha}_A^+}} + \pi$.
    
    1. \textbf{Consistency}: We want to show the following result:

    \begin{align*}
        \prob{|R(g) - \hat{R}(\hat{g})| > \epsilon} \overset{p}{\rightarrow} 0 \text{ as } N_\alpha, N_\beta \rightarrow \infty.
    \end{align*}

    We can decompose the consistency gap as follows:

    \begin{align*}
        |R(g) - \hat{R}(\hat{g})| \leq |R(g) - R(\hat{g})| + |R(\hat{g}) - \hat{R}(\hat{g})|
    \end{align*}


    Since $D_\alpha$ and $D_\beta$ are independently drawn samples, conditioned on $D_\alpha$, $\hat{\alpha}_a^+$ is a constant for all $A=a$ and hence $\hat{R}(\hat{g}) \overset{p}{\rightarrow} R(\hat{g})$ by the weak law of large numbers. It remains to show how $R(\hat{g}) \overset{p}{\rightarrow} R(g)$.

    Let $\phi_a(t) = \expecto{X|A=a}{(1 - 2C)\id{C \geq t}|A=a}, t \in (0,1)$, so that $R(g) = \underset{a \in \{0,1\}}{\sum} \pi_a\cdot \phi_a(\alpha_a)$. Let $0 \leq u < v \leq 1$ and consider the following analysis for a fixed $A=a$:

    \begin{align*}
        |\phi_a(u) - \phi_a(v)| &= |\expecto{}{(1 - 2C)(\id{C \geq u} - \id{C \geq v})|A=a}| \\
        &\leq \prob{u \leq C < v| A=a} \text{ (Since $(\id{C \geq u} - \id{C \geq v}) \neq 0$ when $C \in [u, v)$)} \\
        &= |F_a(v) - F_a(u)|
    \end{align*}

    Now, from Theorem \ref{thm:alpha_cons}, we know that $\hat{\alpha}_a^+ \overset{p}{\rightarrow} \alpha_a$ and since $F_a$ is continuous due to Assumption \ref{assump: eta-continous} and $\alpha_a > 0$, by the continuous mapping theorem, $\phi(\hat{\alpha}_a^+) \overset{p}{\rightarrow} \phi(\alpha_a)$. Since $R(g) = \underset{a \in \{0,1\}}{\sum} \pi_a\cdot \phi_a(\alpha_a)$, $R(\hat{g}) \overset{p}{\rightarrow} R(g)$.

    2. \textbf{Bias}: We want to bound the following quantity: $|R(g) - \expecto{D_\alpha, D_\beta}{\hat{R}(\hat{g})}|$. Let $R_0(g) = R(g) - \pi$, and similarly define $R_0(\hat{g}) = R(\hat{g}) - \pi$.

    Note that since $D_\alpha$ and $D_\beta$ are independently sampled, $\expecto{D_\alpha, D_\beta}{\hat{R}(\hat{g})} = \expecto{D_\alpha}{R_0(\hat{g})} + \expecto{D_\alpha}{\hat{\pi}}$, which means:

    \begin{align*}
        |R(g) - \expecto{D_\alpha, D_\beta}{\hat{R}(\hat{g})}| &\leq |R_0(g) - \expecto{D_\alpha}{R_0(\hat{g})}| + |\pi - \expecto{D_\alpha}{\hat{\pi}}| \\
        &= |R_0(g) - \expecto{D_\alpha}{R_0(\hat{g})}| \text{ (Since $\hat{\pi}$ is an unbiased estimate of $\pi$)}
    \end{align*}

    Let $\phi_a(t) = \expecto{X|A=a}{(1 - 2C)\id{C \geq t}|A=a}$. From our consistency proof, we know that $|\phi_a(u) - \phi_a(v)| \leq |F_a(v) - F_a(u)|$. Using this:


    \begin{align*}
         &|R(g) - \expecto{D_\alpha, D_\beta}{\hat{R}(\hat{g})}| \leq |R_0(g) - \expecto{D_\alpha}{R_0(\hat{g})}| = |\underset{a \in \{0,1\}}{\sum} \pi_a\expecto{D_\alpha}{\phi_a(\alpha_a) - \phi_a(\hat{\alpha}_a^+)}| \\
         &\leq \underset{a \in \{0,1\}}{\sum} \pi_a\expecto{D_\alpha}{|\phi_a(\alpha_a) - \phi_a(\hat{\alpha}_a^+)|} \leq \underset{a \in \{0,1\}}{\sum} \pi_aL_a\expecto{D_\alpha}{|\alpha_a - \hat{\alpha}_a^+|} \text{ (Using Assumption \ref{assump: eta-density})}
    \end{align*}

    From Theorem \ref{thm:alpha_cons}, we know that with $1 - \delta$ probability, $|\hat{\alpha}_a^+ - \alpha_a| < \epsilon$ We also know that $|\alpha_a - \hat{\alpha}_a^+| \leq 1$. Using this, we get:

    \begin{align*}
        &\expecto{D_\alpha}{|\alpha_a -  \hat{\alpha}_a^+|} = \expecto{D_\alpha}{|\alpha_a -  \hat{\alpha}_a^+|\id{|\alpha_a -  \hat{\alpha}_a^+| < \epsilon}} + \expecto{D_\alpha}{|\alpha_a -  \hat{\alpha}_a^+|\id{|\alpha_a -  \hat{\alpha}_a^+| \geq \epsilon}} \\
        &\leq \epsilon\cdot\prob{|\alpha_a -  \hat{\alpha}_a^+| < \epsilon} + 1\cdot\prob{|\alpha_a -  \hat{\alpha}_a^+| \geq \epsilon} \text{ (Since $|\alpha_a -  \hat{\alpha}_a^+| \leq 1$) } \leq \epsilon + \delta.
    \end{align*}

    and therefore,
    
    \begin{align*}
         &|R(g) - \expecto{D_\alpha, D_\beta}{\hat{R}(\hat{g})}| \leq \underset{a \in \{0,1\}}{\sum} \pi_aL_a\expecto{D_\alpha}{|\alpha_a - \hat{\alpha}_a^+|} \leq \underset{a \in \{0,1\}}{\sum} \pi_aL_a(\epsilon_a + \delta_a).
    \end{align*}

    For DP, using the sample complexity bound from Theorem \ref{thm:alpha_cons} and the dominant term, $\epsilon_a = \frac{|\lambda|}{\pi_a^2} \sqrt{\frac{\log(4/\delta_a)}{2N_\alpha}}$, and setting $\delta = \min\{\delta_0, \delta_1\}$:


    \begin{align*}
        |R(g) - \expecto{D_\alpha, D_\beta}{\hat{R}(\hat{g})}| = \mathcal{O}\left(|\lambda|\cdot\max\{L_0, L_1\}\cdot \sqrt{\frac{\log(1/\delta)}{N_\alpha}} \right).
    \end{align*}

    For EOP, using the sample complexity bound from Theorem \ref{thm:alpha_cons} and the $\epsilon$ dependent term, $\epsilon_a = \frac{|\lambda|}{\tau^2\pi_{1a}^2} \sqrt{\frac{8\log(4/\delta_a)}{N_\alpha}}$, and setting $\delta = \min\{\delta_0, \delta_1\}$:


    \begin{align*}
       |R(g) - \expecto{D_\alpha, D_\beta}{\hat{R}(\hat{g})}| = \mathcal{O}\left(\frac{|\lambda| \max\{L_0, L_1\}}{\tau^2} \sqrt{\frac{\log(1/\delta)}{N_\alpha}}\right).
    \end{align*}

    (3) \textbf{Sample Complexity}: We want to ensure the following:

    \begin{align*}
        \prob{|R(g) - \hat{R}(\hat{g})| > \epsilon} \leq \delta.
    \end{align*}

    Note that we can decompose the absolute error as follows via the triangle inequality:

    \begin{align*}
        |R(g) - \hat{R}(\hat{g})| \leq |R_0(g) - R_0(\hat{g})| + |R_0(\hat{g}) - \hat{R}_0(\hat{g})| + |\hat{\pi} - \pi|.
    \end{align*}

    Applying the union bound, we get:

    \begin{align*}
         \prob{|R(g) - \hat{R}(\hat{g})| > \epsilon} &\leq \prob{|R_0(g) - R_0(\hat{g})| > \epsilon/3} + \prob{|R_0(\hat{g}) - \hat{R}_0(\hat{g})| > \epsilon/3} \\
         &+ \prob{|\hat{\pi} - \pi| > \epsilon/3}.
    \end{align*}

    Conditioned on $D_\alpha$, $\hat{\alpha}_a^+$ is a constant and hence we can control $\prob{|R_0(\hat{g}) - \hat{R}_0(\hat{g})| > \epsilon/3}$ via the Hoeffding's inequality. Note that the random variable $Z_i = (1 - 2c_i)\id{c_i \geq \hat{\alpha}_{a_i}^+} \in [-1,1]$.

    \begin{align*}
        \prob{|R_0(\hat{g}) - \hat{R}_0(\hat{g})| > \epsilon/3 | D_\alpha} \leq \delta/3 \implies N_\beta \geq \frac{18}{\epsilon^2} \log\Big(\frac{6}{\delta}\Big). 
    \end{align*}

    Similarly, we can also control $\prob{|\hat{\pi} - \pi| > \epsilon/3}$:

    \begin{align*}
        \prob{|\hat{\pi} - \pi| > \epsilon/3} \leq \delta/3 \implies N_\alpha \geq \frac{9}{2\epsilon^2} \log\Big(\frac{6}{\delta}\Big)
    \end{align*}
    
    Finally, to control $\prob{|R_0(g) - R_0(\hat{g})| > \epsilon/3}$, we can borrow our analysis from the previous proof of bias:

    \begin{align*}
        |R_0(g) - R_0(\hat{g})| \leq \underset{a \in \{0,1\}}{\sum} \pi_aL_a|\alpha_a - \hat{\alpha}_a^+|.
    \end{align*}

    By union bound and splitting the error budget between each group-dependent term (since there are two groups, if the sum exceeds $\epsilon/3$, at least one term exceeds $\epsilon/6)$: 

    \begin{align*}
        \prob{|R_0(g) - R_0(\hat{g})| > \epsilon/3} \leq \underset{a \in \{0,1\}}{\sum} \prob{|\alpha_a - \hat{\alpha}_a^+| > \frac{\epsilon}{6\pi_aL_a}}.
    \end{align*}

    For demographic parity, again using the dominant term: 


    \begin{align*}
        \prob{|\alpha_a - \hat{\alpha}_a^+| > \frac{\epsilon}{6\pi_aL_a}} \leq \delta/6 \implies N_\alpha \geq \frac{18\lambda^2L_a^2}{\epsilon^2\pi_a^2} \log\left(\frac{24}{\delta}\right) = \mathcal{O}\left(\frac{\lambda^2L_a^2}{\epsilon^2} \log\left(\frac{1}{\delta}\right)\right).
    \end{align*}

    For equal opportunity, recall that with $1 - \delta$ probability over the choice of data sample $N_\alpha$, as long as $N_\alpha \geq \frac{1}{2t^2} \log(\frac{2}{\delta})$ for $t = \min \Big\{\frac{\pi_{1a}}{2}, \frac{\tau\pi_{1a}^2}{4|\lambda|},  \frac{\epsilon\tau^2\pi_{1a}^2}{4|\lambda|} \Big\}$, $\prob{|\hat{\alpha}_a - \alpha_a| > \epsilon}\leq \delta$. Using these bounds with our new $\epsilon$ and $\delta$ we get:


    \begin{align*}
        \prob{|\alpha_a - \hat{\alpha}_a^+| > \frac{\epsilon}{6\pi_aL_a}} \leq \delta/6 &\implies N_\alpha \geq \frac{1}{2}\log\left(\frac{12}{\delta}\right)\max\left\{ \frac{2}{\pi_{1a}}, \frac{4|\lambda|}{\tau\pi_{1a}^2}, \frac{24|\lambda|\pi_aL_a}{\epsilon\tau^2\pi_{1a}^2} \right\}^2 \\
        &= \mathcal{O}\left( \frac{\lambda^2L_a^2}{\epsilon^2\tau^4} \log\left(\frac{1}{\delta}\right) \right).
    \end{align*}

    Taking a max over $L_a$ completes our proof. 
\end{proof}


Again, for convenience and readability, we will prove the DP and EOP results for Theorem \ref{thm:beta_unf_cons}.

\begin{theorem} (Proof of the DP part of Theorem \ref{thm:beta_unf_cons} in the main text)
Suppose a dataset $D_\beta$ is used to construct $\hat{\Delta}_{DP}(g_{\hat{\alpha}_a^+})$, where $\hat{\alpha}_a^+$ is the estimator for the DP threshold (Theorem \ref{thm:alpha_cons}). Then the following holds:
\begin{enumerate}
    \item Consistency: $\hat{\Delta}_{DP}(g_{\hat{\alpha}_a^+}) \overset{p}{\rightarrow} \Delta_{DP}(g_\alpha))$.
    \item Bias: For $\delta \in (0,1)$, $ |\Delta_{DP}(g) - \expecto{D_{\beta}, D_\alpha}{\hat{\Delta}_{DP}(\hat{g})}| = \mathcal{O} \left( \frac{1}{\sqrt{N_{\beta}}} +  |\lambda|\cdot\max\{L_0, L_1\} \sqrt{\frac{\log(1/\delta_a)}{N_\alpha}}\right)$.
    \item Sample Complexity: With $1 - \delta$ probability, $|\hat{\Delta}(\hat{g}) - \Delta(g)| \leq \epsilon$, as long as $N_\beta = \mathcal{O}\left( \frac{1}{\epsilon^2}\log(\frac{1}{\delta}) \right)$ and $N_\alpha = \mathcal{O}\left( \frac{\lambda^2\cdot\max\{L_0^2, L_1^2\}}{\epsilon^2} \log\left( \frac{1}{\delta} \right)\right)$.
\end{enumerate}
\end{theorem}

\begin{proof}
For the sake of this proof, we will denote $\Delta_{DP}(g_\alpha)$ as $\Delta(g)$ and  $\hat{\Delta}_{DP}(g_{\hat{\alpha}_a^+})$ as $\hat{\Delta}(\hat{g})$. Then an intermediate estimator $\hat{\Delta}_{DP}(g_{\alpha_a})$ can be denoted as $\hat{\Delta}(g)$. We similarly overload the notations in this fashion for the positive rate estimator $\rho_{\mathrm{PR}, a}$ with $\rho_a$.

(1) \textbf{Consistency}: Most of the inequality arguments in the proof are a direct consequence of the triangle inequality, by introducing an intermediate estimate of the PR ($\hat{\rho}_{a}(g)$). 

We want to show that $\prob{|\hat{\Delta}(\hat{g}) - \Delta(g)| > \epsilon} \rightarrow 0 \text{ as } N_\beta \rightarrow \infty$. By Slutsky's theorem, it will be enough to upper-bound our target event as a sum of simpler events and show that if each of them converges in probability to zero, then we can claim our final result.

Consider the absolute difference term and apply the reverse triangle inequality $(||A| - |B|| \leq |A - B|)$:

\begin{align*}
    &|\Delta(g) - \hat{\Delta}(\hat{g})| \leq |\rho_1(g) - \rho_0(g) - \hat{\rho}_1(\hat{g}) + \hat{\rho}_0(\hat{g})| \\
    &\leq |\rho_1(g) - \hat{\rho}_1(g)| + |\rho_0(g) - \hat{\rho}_0(g)| + |\hat{\rho}_1(g) - \hat{\rho}_1(\hat{g})| + |\hat{\rho}_0(g) - \hat{\rho}_0(\hat{g})|.
\end{align*}

Since similar arguments will work for any group $A=a$, we will now study each term separately for an arbitrary group $A=a$. 

Consider $|\rho_a(g) - \hat{\rho}_a(g)|$. By the Weak law of large numbers, $\frac{1}{N_{\beta,a}} \sum_{i \in [D_{\beta, a}]} \id{c_i \geq \alpha_a} \overset{p}{\rightarrow} \expecto{X|A=a}{\id{c_a \geq \alpha_a}}$, and hence $|\rho_a(g) - \hat{\rho}_a(g)| \overset{p}{\rightarrow} 0$.

Next, consider the term $|\hat{\rho}_a(g) - \hat{\rho}_a(\hat{g})|$. We will now attempt to study the convergence in probability properties of this term:

\begin{align*}
    &|\hat{\rho}_a(g) - \hat{\rho}_a(\hat{g})| = \Bigg| \frac{1}{N_{\beta, a}}\sum_{i \in [D_{\beta, a}]} \id{c_i \geq \alpha_a} - \frac{1}{N_{\beta, a}} \sum_{i \in [D_{\beta, a}]} \id{c_i \geq \hat{\alpha}_a^+} \Bigg| \leq \frac{1}{N_{\beta, a}}  \sum_{i \in [D_{\beta, a}]} Z_i,
\end{align*}

where $Z_i = |\id{c_i \geq \alpha_a} - \id{c_i \geq \hat{\alpha}_a^+}| = \id{c_i \in (\min \{\hat{\alpha}_a^+, \alpha_a \}, \max \{\hat{\alpha}_a^+, \alpha_a \}}$.

We will now show that the numerator converges to zero. Note that conditioned on $D_\alpha$, $\hat{\alpha}_a^+$ is fixed. Since $D_\alpha$ and $D_\beta$ are sampled independently, $\{Z_i, i \in [D_{\beta,a}]\}$ are iid and bounded.
Hence, by the weak law of large numbers, $\frac{1}{N_{\beta, a}}  \sum_{i \in [D_{\beta, a}]} Z_i \overset{p}{\rightarrow} \expecto{D_{\beta,a}}{Z_i | D_\alpha, A=a} = \expecto{D_{\beta, a}}{\id{c_i \in (\min \{\hat{\alpha}_a^+, \alpha_a \}, \max \{\hat{\alpha}_a^+, \alpha_a \})}|D_\alpha, A=a} = \prob{c \in (\min \{\hat{\alpha}_a^+, \alpha_a \}, \max \{\hat{\alpha}_a^+, \alpha_a \})|D_\alpha, A=a} = |F_a(\alpha_a) - F_a(\hat{\alpha}_a^+)|$, where $F_a$ is the group-conditional CDF of $c = \eta(x, a)$, and the absolute difference accounts for both scenarios ($\hat{\alpha}_a^+ > \alpha_a$ and $\hat{\alpha}_a^+ \leq  \alpha_a$).

We are now ready to apply the Continuous Mapping Theorem, since $F_a$ is continuous by Assumption \ref{assump: eta-continous}. Since we ensure $\alpha_a \in (0,1)$, by CMT, $F_a(\hat{\alpha}_a^+) \overset{p}{\rightarrow} F_a(\alpha_a)$, since $\hat{\alpha}_a^+ \overset{p}{\rightarrow} \alpha_a$ (Theorem \ref{thm:alpha_cons}), and therefore $\frac{1}{N_{\beta, a}}  \sum_{i \in [N_{\beta, a}]} Z_i \overset{p}{\rightarrow} 0$ and:

\begin{align*}
    |\hat{\rho}_a(g) - \hat{\rho}_a(\hat{g})| \leq \frac{1}{N_{\beta, a}}  \sum_{i \in [N_{\beta, a}]} Z_i  \overset{p}{\rightarrow} |F_a(\alpha_a) - F_a(\hat{\alpha}_a^+)| \overset{p}{\rightarrow} 0.
\end{align*}

With this, we now have that all our components of $|\Delta(g) - \hat{\Delta}(\hat{g})|$ converge to zero in probability and hence we can claim that $\prob{|\hat{\Delta}_{DP}(\hat{g}) - \Delta_{DP}(g)| > \epsilon} \rightarrow 0 \text{ as } N_\beta \rightarrow \infty$.

(2) \textbf{Bias}: We now want to bound the bias of our estimator, $|\Delta_{DP}(g) - \expecto{D_{\beta}, D_\alpha}{\hat{\Delta}_{DP}(\hat{g})}|$.

\begin{align*}
    & |\Delta_{DP}(g) - \expecto{D_{\beta}, D_\alpha}{\hat{\Delta}_{DP}(\hat{g})}| \\
    &= |\expecto{D_{\beta}, D_\alpha}{\Delta_{DP}(g) - \hat{\Delta}_{DP}(\hat{g})}| \leq \expecto{D_{\beta}, D_\alpha}{|\hat{\Delta}_{DP}(\hat{g}) - \Delta_{DP}(g)|} \\
    &\leq \expecto{D_{\beta}, D_\alpha}{|\hat{\rho}_1(\hat{g}) - \hat{\rho}_0(\hat{g}) - \rho_1(g) + \rho_0(g)|} \\
    &\text{(By using the Reverse Triangle Inequality) } \\
    &\leq \expecto{D_{\beta,1}, D_\alpha}{|\rho_1(g) - \hat{\rho}_1(g)|} + \expecto{D_{\beta, 0}, D_\alpha}{|\rho_0(g) - \hat{\rho}_0(g)|} \\
    &+ \expecto{D_{\beta, 1}, D_\alpha}{|\hat{\rho}_1(g) - \hat{\rho}_1(\hat{g})|} + \expecto{D_{\beta, 0}, D_\alpha}{|\hat{\rho}_0(g) - \hat{\rho}_0(\hat{g})|} \\
    &\text{(By introducing an intermediate estimator $\hat{\rho}_a(g)$,} \\
    & \text{ and noting that the expectation is over group-specific samples $D_{\beta. a}$)}.
\end{align*}

Once again, we will only focus on deriving the statistical bias for an arbitrary group $A=a$. Consider the first term. At this point, note that $\hat{\rho}_a(g)$ is an unbiased estimate of $\rho_a(g)$, and hence by Lemma \ref{lem:var_upper_bound}, 

\begin{align*}
    \expecto{D_{\beta,a}, D_\alpha}{|\hat{\rho}_a(g) - \rho_a(g)|} \leq \sqrt{Var(\hat{\rho}_a(g))} \leq \frac{1}{2\sqrt{N_{\beta,a}}},
\end{align*}

where the last step follows due to Popoviciu's Inequality on $\id{c_i \geq \alpha_a}$.

We can now focus on the second term: $\expecto{D_{\beta, a}, D_\alpha}{|\hat{\rho}_a(g) - \hat{\rho}_a(\hat{g})|}$.

\begin{align*}
    \expecto{D_{\beta, a}, D_\alpha}{|\hat{\rho}_a(\hat{g}) - \hat{\rho}_a(g)|} &= \expecto{D_{\beta, a}, D_\alpha}{\Bigg|\frac{1}{N_{\beta,a}} \sum_{i \in [D_{\beta,a}]} (\id{c_i \geq \hat{\alpha}_a^+} - \id{c_i \geq \alpha_a})\Bigg|}\\
    & \leq \expecto{D_{\beta, a}, D_\alpha}{\frac{1}{N_{\beta,a}} \sum_{i \in [D_{\beta,a}]} |\id{c_i \geq \hat{\alpha}_a^+} - \id{c_i \geq \alpha_a}|} 
\end{align*}

From our consistency analysis, we know that $Z_i = |\id{c_i \geq \alpha_a} - \id{c_i \geq \hat{\alpha}_a^+}| = \id{c_i \in (\min\{\hat{\alpha}_a^+, \alpha_A\}, \max\{\hat{\alpha}_a^+, \alpha_a\})}$, and hence:

\begin{align*}
    &\expecto{D_{\beta, a}, D_\alpha}{\frac{1}{N_{\beta,a}} \sum_{i \in [D_{\beta,a}]} Z_i} \\
    &= \frac{1}{N_{\beta,a}} \sum_{i \in [D_{\beta,a}]} \expecto{D_{\beta, a}, D_\alpha}{Z_i} = \expecto{D_{\beta, a}, D_\alpha}{\id{c_i \in (\min\{\hat{\alpha}_a^+, \alpha_a\}, \max\{\hat{\alpha}_a^+, \alpha_a\})}} \\
    &= \expecto{D_\alpha}{\expecto{D_{\beta,a}}{\id{c_i \in (\min\{\hat{\alpha}_a^+, \alpha_a\}, \max\{\hat{\alpha}_a^+, \alpha_a\})}}|D_\alpha} = \expecto{D_\alpha}{|F_a(\alpha_a) - F_a(\hat{\alpha}_a^+)|}. \\
\end{align*}


For a fixed $\gamma \in (0,1)$, let $\mathcal{G} = \{ |\hat{\alpha}_a^+ - \alpha_a| \leq \gamma_a \}$ be the good event that the threshold error is at most $\gamma_a$. Then,

\begin{align*}
    &\expecto{D_{\beta, a}, D_\alpha}{|\hat{\rho}_a(\hat{g}) - \hat{\rho}_a(g)|} \leq \expecto{D_\alpha}{|F_a(\alpha_a) - F_a(\hat{\alpha}_a^+)|} \\
    &= \expecto{D_\alpha}{|F_a(\alpha_a) - F_a(\hat{\alpha}_a^+)|\mathbbm{1}_{\mathcal{G}}} + \expecto{D_\alpha}{|F_a(\alpha_a) - F_a(\hat{\alpha}_a^+)|\mathbbm{1}_{\mathcal{G}^C}} \\
    &\leq L_a\gamma_a + \prob{\mathcal{G}^c} \text{ (Using Assumption \ref{assump: eta-density} and using $|F_a(\alpha_a) - F_a(\hat{\alpha}_a^+)| \leq 1$)} \\
    &= L_a\gamma_a + \prob{|\alpha_a - \hat{\alpha}_a^+| > \gamma_a}.
\end{align*}



Overall, 

\begin{align*}
    |\Delta_{DP}(g) - \expecto{D_{\beta}, D_\alpha}{\hat{\Delta}_{DP}(\hat{g})}| \leq \underset{a \in \{0,1\}}{\sum} \left( \frac{1}{2\sqrt{N_{\beta,a}}} +  L_a\gamma_a + \prob{|\alpha_a - \hat{\alpha}_a^+| > \gamma_a} \right).
\end{align*}

Using the sample complexity bound from Theorem \ref{thm:alpha_cons} and the dominant term, $\gamma_a = \frac{|\lambda|}{\pi_a^2} \sqrt{\frac{\log(4/\delta_a)}{2N_\alpha}}$, giving us:
\begin{align*}
    |\Delta_{DP}(g) - \expecto{D_{\beta}, D_\alpha}{\hat{\Delta}_{DP}(\hat{g})}| \leq \underset{a \in \{0,1\}}{\sum} \left( \frac{1}{2\sqrt{N_{\beta,a}}} +  \frac{L_a|\lambda|}{\pi_a^2} \sqrt{\frac{\log(4/\delta_a)}{2N_\alpha}} + \delta_a \right).
\end{align*}

Setting $N_\beta = \min\{N_{\beta,0}, N_{\beta,1}\}$, we can obtain:

\begin{align*}
    |\Delta_{DP}(g) - \expecto{D_{\beta}, D_\alpha}{\hat{\Delta}_{DP}(\hat{g})}| = \mathcal{O} \left( \frac{1}{\sqrt{N_{\beta}}} +  |\lambda|\cdot\max\{L_0, L_1\} \sqrt{\frac{\log(1/\delta_a)}{N_\alpha}}\right).
\end{align*}

(3) \textbf{Sample Complexity}: To bound the sample complexity, we need to bound the following event:

\begin{align*}
    \prob{|\hat{\Delta}(\hat{g}) - \Delta(g)| > \epsilon} \leq \delta.
\end{align*}

Let $\Delta(\hat{g})$ denote the true DP difference with respect to the estimated threshold $\hat{\alpha}_a^+$. Consider the following quantity and its breakdown via the triangle inequality:

\begin{align*}
    &|\hat{\Delta}(\hat{g}) - \Delta(g)| \leq |\hat{\Delta}(\hat{g}) - \Delta(\hat{g})| + |\Delta(\hat{g}) - \Delta(g)|\text{ (By Triangle Inequality) }
\end{align*}


By the union bound, we can upper bound our total error

\begin{align*}
    &\prob{|\hat{\Delta}(\hat{g}) - \Delta(g)| > \epsilon} \leq  \prob{|\hat{\Delta}(\hat{g}) - \Delta(\hat{g})| > \epsilon/2} + \prob{|\Delta(\hat{g}) - \Delta(g)| > \epsilon/2}.
\end{align*}


We can first use the Hoeffding bound for the first term by noting that $|\hat{\Delta}(\hat{g}) - \Delta(\hat{g})| \leq \sum_{a \in \mathcal{A}} |\rho_a(\hat{g}) - \hat{\rho}_a(\hat{g})|$:


\begin{align*}
    &\prob{|\rho_a(\hat{g}) - \hat{\rho}_a(\hat{g})| > \epsilon/4} \leq 2\exp\Big(\frac{-2N_{\beta,a}\epsilon^2}{16}\Big) \leq \frac{\delta}{4} \implies N_{\beta,a} \geq \frac{8\log(\frac{8}{\delta})}{\epsilon^2}.
\end{align*}

For both groups, we would need $N_\beta = \mathcal{O}\left( \frac{1}{\epsilon^2}\log(\frac{1}{\delta}) \right)$.



Next, consider the second term:

\begin{align*}
    |\Delta(\hat{g}) - \Delta(g)| &\leq \sum_{a \in \mathcal{A}}|\rho_a(g) - \rho_a(\hat{g})| = \sum_{a \in \mathcal{A}}|\prob{C \geq \alpha_a | A=a} - \prob{C \geq \hat{\alpha}_a^+ | A=a})| \\
    &= \sum_{a \in \mathcal{A}}|1 - F_a(\alpha_a) - 1 + F_a(\hat{\alpha}_a^+)| \leq \sum_{a \in \mathcal{A}} L_a\cdot |\alpha_a - \hat{\alpha}_a^+| \text{ (Using Assumption \ref{assump: eta-density})}.
\end{align*}

Now, we can figure out the sample complexity of each term:

\begin{align*}
    \prob{|\rho_a(g) - \rho_a(\hat{g})| > \epsilon/4}  &= \prob{|\alpha_a - \hat{\alpha}_a^+|  > \frac{\epsilon}{4L_a}} < \frac{\delta}{4}.
\end{align*}


Now, we can use Theorem \ref{thm:alpha_cons} to get the number of samples required to upper bound the above error: $N_\alpha = \mathcal{O}\left( \frac{\lambda^2\cdot\max\{L_0^2, L_1^2\}}{\epsilon^2} \log\left( \frac{1}{\delta} \right)\right)$.
\end{proof}


\begin{theorem} (Proof of the EOP part of Theorem \ref{thm:beta_unf_cons} in the main text)
Suppose a dataset $D_\beta$ is used to construct $\hat{\Delta}_{EOP}(g_{\hat{\alpha}_a^+})$, where $\hat{\alpha}_a^+$ is the estimator for the EOP threshold (Theorem \ref{thm:alpha_cons}). Then the following holds:
\begin{enumerate}
    \item Consistency: $\hat{\Delta}_{EOP}(g_{\hat{\alpha}_a^+}) \overset{p}{\rightarrow} \Delta_{EOP}(g_\alpha)$.
    \item Bias: $|\Delta_{EOP}(g) - \expecto{D_{\beta}, D_\alpha}{\hat{\Delta}_{EOP}(\hat{g})}| =  \mathcal{O}\Bigg( \frac{|\lambda|\cdot\max\{L_0, L_1\}}{\tau^2}   \sqrt{\frac{\log(1/\delta)}{N_\alpha}} + \frac{1}{\sqrt{N_{\beta}}} \Bigg).$
    \item Sample Complexity:  With $1 - \delta$ probability, $|\hat{\Delta}(\hat{g}) - \Delta(g)| \leq \epsilon$, as long as $N_{\alpha} = \mathcal{O}\left( \frac{\lambda^2\max\{L_0^2, L_1^2\}}{\epsilon^2\tau^4}\log\left(\frac{1}{\delta}\right) \right) \text{ and } N_{\beta} = \mathcal{O}\left\{ \frac{1}{\epsilon^2} \log\left( \frac{1}{\delta} \right) \right\}$.
\end{enumerate}
\end{theorem}

\begin{proof}
For the sake of this proof, we will denote $\Delta_{EOP}(g_\alpha)$ as $\Delta(g)$ and  $\hat{\Delta}_{EOP}(g_{\hat{\alpha}_a^+})$ as $\hat{\Delta}(\hat{g})$. Then an intermediate estimator $\hat{\Delta}_{EOP}(g_{\alpha_a})$ can be denoted as $\hat{\Delta}(g)$. We similarly overload the notations in this fashion for the positive rate estimator $\rho_{\mathrm{TPR}, a}$ with $\rho_a$.

Recalling the definitions of our estimations, 

\begin{align*}
    \rho_{\mathrm{TPR}, a}(g_\alpha) &= \prob{g_\alpha=1|Y=1, A=a} = \frac{\expecto{X|A=a}{ \id{\eta(x,a) \geq \alpha_a}\cdot\prob{Y=1|X=x, A=a}}}{\pi_{1|a}} \\
    &= \frac{\expecto{X|A=a}{\id{C \geq \alpha_a}\cdot C}}{\pi_{1|a}},
\end{align*}

and,


\begin{align*}
    \hat{\rho}_{\mathrm{TPR},a}(g_{\hat{\alpha}_a^+}) = \frac{\frac{1}{N_{\beta,a}} \sum_{i \in [D_{\beta, a}]} (c_i(\id{c_i \geq \hat{\alpha}_a^+})}{\hat{\pi}_{1|a}^+} = \frac{\hat{\mathrm{E}}_{X|A=a}[\id{c \geq \hat{\alpha}_a^+}c]}{\hat{\pi}_{1|a}^+}
\end{align*}

(1) \textbf{Consistency}: Most of the inequality arguments in the proof are a direct consequence of the triangle inequality, by introducing an intermediate estimate of the TPR ($\hat{\rho}_{a}(g)$). 

We want to show that $\prob{|\hat{\Delta}(\hat{g}) - \Delta(g)| > \epsilon} \rightarrow 0 \text{ as } N_\beta \rightarrow \infty$. By Slutsky's theorem, it will be enough to upper-bound our target event as a sum of simpler events and show that if each of them converges in probability to zero, then we can claim our final result.

Consider the absolute difference term and apply the reverse triangle inequality $(||A| - |B|| \leq |A - B|)$:

\begin{align*}
    &|\Delta(g) - \hat{\Delta}(\hat{g})| \leq |\rho_1(g) - \rho_0(g) - \hat{\rho}_1(\hat{g}) + \hat{\rho}_0(\hat{g})| \\
    &\leq |\rho_1(g) - \rho_1(\hat{g})| + |\rho_0(g) - \rho_0(\hat{g})| + |\rho_1(\hat{g}) - \hat{\rho}_1(\hat{g})| + |\rho_0(\hat{g}) - \hat{\rho}_0(\hat{g})|.
\end{align*}

Since similar arguments will work for any group $A=a$, we will now study each term separately for an arbitrary group $A=a$. 

Consider $|\rho_a(\hat{g}) - \hat{\rho}_a(\hat{g})|$. $\hat{\mathrm{E}}_{X|A=a}[\id{c \geq \hat{\alpha}_a^+}c] \rightarrow \mathrm{E}_{X|A=a}[\id{c \geq \hat{\alpha}_a^+}c]$ and $\hat{\pi}_{1|a}^+ \overset{p}{\rightarrow} \pi_{1|a}$ due to the weak law of large numbers, since conditioned on $D_\alpha$, $\hat{\alpha}_a^+$ and $\hat{\pi}_{1|a}^+$ are fixed numbers.

Since $\pi_{1|a} > 0$ by assumption, by Slutsky's theorem, $\hat{\rho}_a(\hat{g}) = \frac{\hat{\mathrm{E}}_{X|A=a}[\id{c \geq \hat{\alpha}_a^+}c]}{\hat{\pi}_{1|a}^+} \overset{p}{\rightarrow} \rho_a(\hat{g})$.

Next, consider the term $|\rho_a(\hat{g}) - \rho_a(g)|$. Since the denominator for both is a constant $\pi_{1|a}$, we will focus on showing that the difference in the numerator converges to zero.

\begin{align*}
    &|\expecto{X|A=a}{\id{C \geq \alpha_a}C} - \expecto{X|A=a}{\id{C \geq \hat{\alpha}_a^+}C}| \leq \expecto{X|A=a}{Z}
\end{align*}

where $Z = |\id{C \geq \alpha_a} - \id{C \geq \hat{\alpha}_a^+}| = \id{C \in (\min \{\hat{\alpha}_a^+, \alpha_a \}, \max \{\hat{\alpha}_a^+, \alpha_a \})}$ and $C \leq 1$. Using this we can write $\expecto{X|A=a}{Z} = \prob{\min \{\hat{\alpha}_a^+, \alpha_a \} \leq C < \max \{\hat{\alpha}_a^+, \alpha_a \}|A=a} = |F_a(\hat{\alpha}_a^+) - F_a(\alpha_a)|$. Since $F_a$ is continuous, and from Theorem \ref{thm:alpha_cons}, $\hat{\alpha}_a^+ \overset{p}{\rightarrow} \alpha_a$, by the continuous mapping theorem, $F_a(\hat{\alpha}_a^+) \overset{p}{\rightarrow} F_a(\alpha_a)$. Combining all the pieces, $\hat{\Delta}_{EOP}(g_{\hat{\alpha}_a^+}) \overset{p}{\rightarrow} \Delta_{EOP}(g_\alpha)$.

(2) \textbf{Bias}: We now want to bound the bias of our estimator, $|\Delta_{EOP}(g) - \expecto{D_{\beta}, D_\alpha}{\hat{\Delta}_{EOP}(\hat{g})}|$.

\begin{align*}
    & |\Delta_{EOP}(g) - \expecto{D_{\beta}, D_\alpha}{\hat{\Delta}_{EOP}(\hat{g})}| = |\expecto{D_{\beta}, D_\alpha}{\Delta_{EOP}(g) - \hat{\Delta}_{EOP}(\hat{g})}| \\
    &\leq \expecto{D_{\beta}, D_\alpha}{|\hat{\Delta}_{EOP}(\hat{g}) - \Delta_{EOP}(g)|} \leq \expecto{D_{\beta}, D_\alpha}{|\hat{\rho}_1(\hat{g}) - \hat{\rho}_0(\hat{g}) - \rho_1(g) + \rho_0(g)|} \\
    &\text{(By using the Reverse Triangle Inequality) } \\
    &\leq \expecto{D_{\beta,1}, D_\alpha}{|\rho_1(g) - \rho_1(\hat{g})}| + \expecto{D_{\beta, 0}, D_\alpha}{|\rho_0(g) - \rho_0(\hat{g})|} \\
    &+ \expecto{D_{\beta, 1}, D_\alpha}{|\rho_1(\hat{g}) - \hat{\rho}_1(\hat{g})|} + \expecto{D_{\beta, 0}, D_\alpha}{|\rho_0(\hat{g}) - \hat{\rho}_0(\hat{g})|} \\
    &\text{(By introducing an intermediate estimator $\rho_a(\hat{g})$,} \\
    & \text{ and noting that the expectation is over group-specific samples $D_{\beta. a}$)}.
\end{align*}

Once again, we will only focus on deriving the statistical bias for an arbitrary group $A=a$. Consider the term $\expecto{D_{\beta, a}, D_\alpha}{|\rho_a(\hat{g}) - \rho_a(g)|}$. Since the denominator for both is a constant $\pi_{1|a}$, we will focus on showing that the difference in the numerator converges to zero. Reusing our analysis from the previous proof on consistency and using Assumption \ref{assump: eta-density}, we can write:

\begin{align*}
   \expecto{D_{\beta, a}, D_\alpha}{|\rho_a(\hat{g}) - \rho_a(g)|} \leq \frac{1}{\pi_{1|a}} \expecto{D_\alpha}{|F_a(\hat{\alpha}_a^+) - F_a(\alpha_a)|} \leq \frac{L_a}{\pi_{1|a}} \expecto{D_\alpha}{|\hat{\alpha}_a^+ - \alpha_a|}
\end{align*}

From Theorem \ref{thm:alpha_cons}, we know that with $1 - \delta$ probability, $|\hat{\alpha}_a^+ - \alpha_a| < \epsilon$. We also know that $|\alpha_a - \hat{\alpha}_a^+| \leq 1$. Using this, we get:

\begin{align*}
    &\expecto{D_\alpha}{|\alpha_a -  \hat{\alpha}_a^+|} = \expecto{D_\alpha}{|\alpha_a -  \hat{\alpha}_a^+|\id{|\alpha_a -  \hat{\alpha}_a^+| < \epsilon_a}} + \expecto{D_\alpha}{|\alpha_a -  \hat{\alpha}_a^+|\id{|\alpha_a -  \hat{\alpha}_a^+| \geq \epsilon_a}} \\
    &\leq \epsilon_a\cdot\prob{|\alpha_a -  \hat{\alpha}_a^+| < \epsilon_a} + 1\cdot\prob{|\alpha_a -  \hat{\alpha}_a^+| \geq \epsilon_a} \text{ (Since $|\alpha_a -  \hat{\alpha}_a^+| \leq 1$) } \leq \epsilon_a + \delta_a.
\end{align*}

Using $\epsilon_a = \frac{|\lambda|}{\tau^2\pi_{1a}^2} \sqrt{\frac{8\log(2/\delta_a)}{N_\alpha}}$, our final bound becomes:

\begin{align*}
   \expecto{D_{\beta, a}, D_\alpha}{|\rho_a(\hat{g}) - \rho_a(g)|} \leq \frac{L_a}{\pi_{1|a}} \left( \frac{|\lambda|}{\tau^2\pi_{1a}^2} \sqrt{\frac{8\log(2/\delta_a)}{N_\alpha}} +  \delta_a \right)
\end{align*}

Next, consider the term $\expecto{D_{\beta, a}, D_\alpha}{|\rho_a(\hat{g}) - \hat{\rho}_a(\hat{g})|}$. 

\begin{align*}
    &\expecto{D_{\beta, a}, D_\alpha}{|\rho_a(\hat{g}) - \hat{\rho}_a(\hat{g})|} = \expecto{D_{\beta,a}, D_\alpha}{|\frac{\expecto{X|A=a}{\id{C \geq \hat{\alpha}_a^+}C}}{\pi_{1|a}} - \frac{\hat{\mathrm{E}}_{X|A=a}[\id{C \geq \hat{\alpha}_a^+}C]}{\hat{\pi}_{1|a}^+}|} \\
    &= \mathrm{E}_{D_{\beta,a}, D_\alpha}\Bigg[ |\frac{\expecto{X|A=a}{\id{C \geq \hat{\alpha}_a^+}C}}{\pi_{1|a}} - \frac{\hat{\mathrm{E}}_{X|A=a}[\id{C \geq \hat{\alpha}_a^+}C]}{\pi_{1|a}} \\
    &+ \frac{\hat{\mathrm{E}}_{X|A=a}[\id{C \geq \hat{\alpha}_a^+}C]}{\pi_{1|a}} - \frac{\hat{\mathrm{E}}_{X|A=a}[\id{C \geq \hat{\alpha}_a^+}C]}{\hat{\pi}_{1|a}^+}|\Bigg] \\
    &\text{ (By introducing an intermediate term $\frac{\hat{\mathrm{E}}_{X|A=a}[\id{C \geq \hat{\alpha}_a^+}C]}{\pi_{1|a}}$)} \\
    &\leq \expecto{D_{\beta,a}, D_\alpha}{\frac{1}{\pi_{1|a}} \cdot |\expecto{X|A=a}{\id{C \geq \hat{\alpha}_a^+}C}  - \hat{\mathrm{E}}_{X|A=a}[\id{C \geq \hat{\alpha}_a^+}C]|} \\
    &+ \expecto{D_{\beta,a}, D_\alpha}{\hat{\mathrm{E}}_{X|A=a}[\id{C \geq \hat{\alpha}_a^+}C]\cdot\Big| \frac{1}{\pi_{1|a}} - \frac{1}{\hat{\pi}_{1|a}^+}\Big|} \\
    &\leq \frac{1}{\pi_{1|a}}\expecto{D_{\beta,a}, D_\alpha}{|\expecto{X|A=a}{\id{C \geq \hat{\alpha}_a^+}C}  - \hat{\mathrm{E}}_{X|A=a}[\id{C \geq \hat{\alpha}_a^+}C]|} \\
    &+ \expecto{D_{\beta,a}, D_\alpha}{\Big| \frac{1}{\pi_{1|a}} - \frac{1}{\hat{\pi}_{1|a}^+}\Big|} \text{ (Since $\hat{\mathrm{E}}_{X|A=a}[\id{C \geq \hat{\alpha}_a^+}C] \leq 1$)}
\end{align*}

Conditioned on $D_\alpha$, $\hat{\alpha}_a^+$ is fixed. Therefore, we can use Lemma \ref{lem:var_upper_bound} to upper bound the term $\expecto{D_{\beta,a}, D_\alpha}{|\expecto{X|A=a}{\id{C \geq \hat{\alpha}_a^+}C}  - \hat{\mathrm{E}}_{X|A=a}[\id{C \geq \hat{\alpha}_a^+}C]|}$:

\begin{align*}
    &\expecto{D_{\beta,a}, D_\alpha}{|\expecto{X|A=a}{\id{C \geq \hat{\alpha}_a^+}C}  - \hat{\mathrm{E}}_{X|A=a}[\id{C \geq \hat{\alpha}_a^+}C]|} \\
    &= \expecto{D_{\beta,a}}{|\expecto{X|A=a}{\id{C \geq \hat{\alpha}_a^+}C}  - \hat{\mathrm{E}}_{X|A=a}[\id{C \geq \hat{\alpha}_a^+}C]|~|D_\alpha} \\
    &= \sqrt{Var( \hat{\mathrm{E}}_{X|A=a}[\id{C \geq \hat{\alpha}_a^+}C]|D_\alpha}) = \sqrt{\frac{1}{N_{\beta,a}}Var(\id{C \geq \alpha_a^+}C)} \\
    &\leq \frac{1}{2\sqrt{N_{\beta,a}}} \text{ (Using Popoviciu's Inequality).}
\end{align*}

To bound $\expecto{D_{\beta,a}, D_\alpha}{\Big| \frac{1}{\pi_{1|a}} - \frac{1}{\hat{\pi}_{1|a}^+}\Big|} = \expecto{D_\alpha}{\Big| \frac{1}{\pi_{1|a}} - \frac{1}{\hat{\pi}_{1|a}^+}\Big|}$ (Since $\hat{\pi}_{1|a}^+$ is estimated using $D_\alpha$), let $\mathcal{G} = \{\hat{\pi}_{1|a} \geq \frac{\pi_{1|a}}{2}\}$ denote the event that $\hat{\pi}_{1|a}$ and $\pi_{1|a}$ are close. Note that on $\mathcal{G}$, $\hat{\pi}_{1|a}^+ = \hat{\pi}_{1|a}$ for sufficiently many samples ($1/N_{\alpha,a} \leq \pi_{1|a}/2$):

\begin{align*}
    &\expecto{D_\alpha}{\Big| \frac{1}{\pi_{1|a}^+} - \frac{1}{\hat{\pi}_{1|a}}\Big|} = \expecto{D_\alpha}{\Big| \frac{1}{\pi_{1|a}} - \frac{1}{\hat{\pi}_{1|a}}\Big|\mathbbm{1}_{\mathcal{G}}} + \expecto{D_\alpha}{\Big| \frac{1}{\pi_{1|a}} - \frac{1}{\hat{\pi}_{1|a}^+}\Big|\mathbbm{1}_{\mathcal{G}_c}} \\
    &= \expecto{D_\alpha}{\frac{|\hat{\pi}_{1|a} - \pi_{1|a}|}{\hat{\pi}_{1|a}\pi_{1|a}}\mathbbm{1}_{\mathcal{G}}} + \expecto{D_\alpha}{\Big| \frac{1}{\pi_{1|a}} - \frac{1}{\hat{\pi}_{1|a}^+}\Big|\mathbbm{1}_{\mathcal{G}_c}} \\
    &\leq \frac{2}{\pi_{1|a}^2} \expecto{D_\alpha}{|\hat{\pi}_{1|a} - \pi_{1|a}|} + \expecto{D_\alpha}{\Big| \frac{1}{\pi_{1|a}} - \frac{1}{\hat{\pi}_{1|a}^+}\Big|\mathbbm{1}_{\mathcal{G}_c}} \text{ (On $\mathcal{G}$, $\hat{\pi}_{1|a} \geq \pi_{1|a}/2$) }\\
    &\leq \frac{1}{\pi_{1|a}^2\sqrt{N_{\alpha,a}}} + \expecto{D_\alpha}{\Big| \frac{1}{\pi_{1|a}} - \frac{1}{\hat{\pi}_{1|a}^+}\Big|\mathbbm{1}_{\mathcal{G}_c}} \text{ (Using Lemma \ref{lem:var_upper_bound} and Popoviciu's Inequality) }\\
    &\leq \frac{1}{\pi_{1|a}^2\sqrt{N_{\alpha,a}}} + \expecto{D_\alpha}{\Big( \frac{1}{\pi_{1|a}} + N_{\alpha,a} \Big)\mathbbm{1}_{\mathcal{G}_c}} \text{ (Using the form of our truncated estimator $\hat{\pi}_{1|a}^+$) }\\
    &\leq \frac{1}{\pi_{1|a}^2\sqrt{N_{\alpha,a}}} + \Big( \frac{1}{\pi_{1|a}} + N_{\alpha,a} \Big)\prob{\mathcal{G}_c} \text{ (Using the Cauchy-Schwartz Inequality) }\\
    &\leq \frac{1}{\pi_{1|a}^2\sqrt{N_{\alpha,a}}} + \Big( \frac{1}{\pi_{1|a}} + N_{\alpha,a} \Big)\exp\Big(-\frac{N_{\alpha,a}\pi_{1|a}^2}{2}\Big) \text{ (Using the Hoeffding's Inequality) }\\
\end{align*}

Our upper bound becomes:

\begin{align*}
    \expecto{D_{\beta, a}, D_\alpha}{|\rho_a(\hat{g}) - \hat{\rho}_a(\hat{g})|} \leq \frac{1}{2\pi_{1|a}\sqrt{N_{\beta,a}}} + \frac{1}{\pi_{1|a}^2\sqrt{N_{\alpha,a}}} + \Big( \frac{1}{\pi_{1|a}} + N_{\alpha,a} \Big)\exp\Big(-\frac{N_{\alpha,a}\pi_{1|a}^2}{2}\Big)
\end{align*}

and our overall upper bound on the bias of the estimator takes the form:

\begin{align*}
    &|\Delta_{EOP}(g) - \expecto{D_{\beta}, D_\alpha}{\hat{\Delta}_{EOP}(\hat{g})}| \\
    &\leq \underset{a \in \{0,1\}}{\sum} \Bigg( \frac{L_a}{\pi_{1|a}} \left( \frac{|\lambda|}{\tau^2\pi_{1a}^2} \sqrt{\frac{8\log(2/\delta_a)}{N_\alpha}} +  \delta_a \right) + \frac{1}{2\pi_{1|a}\sqrt{N_{\beta,a}}} + \frac{1}{\pi_{1|a}^2\sqrt{N_{\alpha,a}}} \\
    &+ \Big( \frac{1}{\pi_{1|a}} + N_{\alpha,a} \Big)\exp\Big(-\frac{N_{\alpha,a}\pi_{1|a}^2}{2}\Big) \Bigg) \\
    &= \underset{a \in \{0,1\}}{\sum} \mathcal{O}\Bigg( \frac{L_a}{\pi_{1|a}} \left( \frac{|\lambda|}{\tau^2\pi_{1a}^2} \sqrt{\frac{\log(2/\delta_a)}{N_\alpha}} +  \delta_a \right) + \frac{1}{2\pi_{1|a}\sqrt{N_{\beta,a}}} + \frac{1}{\pi_{1|a}^2\sqrt{N_{\alpha,a}}} \Bigg)
\end{align*}

Setting $\delta = \min\{\delta_0, \delta_1\}$, $N_\alpha = \min\{N_{\alpha,0}, N_{\alpha,1}\}$ and $N_\beta = \min\{N_{\beta,0}, N_{\beta,1}\}$:

\begin{align*}
    |\Delta_{EOP}(g) - \expecto{D_{\beta}, D_\alpha}{\hat{\Delta}_{EOP}(\hat{g})}| =  \mathcal{O}\Bigg( \frac{|\lambda|\cdot\max\{L_0, L_1\}}{\tau^2}   \sqrt{\frac{\log(1/\delta)}{N_\alpha}} + \frac{1}{\sqrt{N_{\beta}}} \Bigg)
\end{align*}

(3) \textbf{Sample Complexity}: To bound the sample complexity, we need to bound the following event:

\begin{align*}
    \prob{|\hat{\Delta}(\hat{g}) - \Delta(g)| > \epsilon} \leq \delta.
\end{align*}

Let $\Delta(\hat{g})$ denote the true EOP difference with respect to the estimated threshold $\hat{\alpha}_a^+$. Consider the absolute difference term and apply the reverse triangle inequality $(||A| - |B|| \leq |A - B|)$:

\begin{align*}
    &|\Delta(g) - \hat{\Delta}(\hat{g})| \leq |\rho_1(g) - \rho_0(g) - \hat{\rho}_1(\hat{g}) + \hat{\rho}_0(\hat{g})| \\
    &\leq |\rho_1(g) - \rho_1(\hat{g})| + |\rho_0(g) - \rho_0(\hat{g})| + |\rho_1(\hat{g}) - \hat{\rho}_1(\hat{g})| + |\rho_0(\hat{g}) - \hat{\rho}_0(\hat{g})|.
\end{align*}

By the union bound, we want to ensure the following:

\begin{align*}
    &\prob{|\Delta(g) - \hat{\Delta}(\hat{g})| > \epsilon} \leq \underset{a \in \{ 0,1 \}}{\sum}\prob{|\rho_a(g) - \rho_a(\hat{g})| > \epsilon/4} + \prob{|\rho_a(\hat{g}) - \hat{\rho}_a(\hat{g})| > \epsilon/4}.
\end{align*}

Since similar arguments will work for any group $A=a$, we will now study each term separately for an arbitrary group $A=a$. From our analysis in previous part, we know that $|\rho_a(g) - \rho_a(\hat{g})| \leq \frac{L_a}{\pi_{1|a}} |\hat{\alpha}_a^+ - \alpha_a|$ and therefore $\prob{|\rho_a(g) - \rho_a(\hat{g})| > \epsilon/4} \leq \prob{|\hat{\alpha}_a^+ - \alpha_a| > \frac{\epsilon\pi_{1|a}}{4L_a}}$.

Using Theorem \ref{thm:alpha_cons}, recall that with $1 - \delta$ probability over the choice of data sample $N_\alpha$, as long as $N_\alpha \geq \frac{1}{2t^2} \log(\frac{2}{\delta})$ for $t = \min \Big\{\frac{\pi_{1a}}{2}, \frac{\tau\pi_{1a}^2}{4|\lambda|},  \frac{\epsilon\tau^2\pi_{1a}^2}{4|\lambda|} \Big\}$, $\prob{|\hat{\alpha}_a - \alpha_a| > \epsilon}\leq \delta$. Using these bounds with our new $\epsilon$ and $\delta$ we get:


\begin{align*}
    \prob{|\alpha_a - \hat{\alpha}_a^+| > \frac{\epsilon\pi_{1|a}}{4L_a}} \leq \delta/4 &\implies N_\alpha \geq \frac{1}{2}\log\left(\frac{8}{\delta}\right)\max\left\{ \frac{2}{\pi_{1a}}, \frac{4|\lambda|}{\tau\pi_{1a}^2}, \frac{16|\lambda|L_a}{\epsilon\tau^2\pi_{1|a}\pi_{1a}^2} \right\}^2 \\
    &= \mathcal{O}\left( \frac{\lambda^2L_a^2}{\epsilon^2\tau^4}\log\left(\frac{1}{\delta}\right) \right).
\end{align*}

Now we focus on the other event. Just like the bias proof, we will proceed by introducing an intermediate estimator: 

\begin{align*}
    |\rho_a(\hat{g}) - \hat{\rho}_a(\hat{g})| &= \Bigg|\frac{\expecto{X|A=a}{\id{C \geq \hat{\alpha}_a^+}C}}{\pi_{1|a}} - \frac{\hat{\mathrm{E}}_{X|A=a}[\id{C \geq \hat{\alpha}_a^+}C]}{\hat{\pi}_{1|a}^+}\Bigg| \\
    &= \Bigg|\frac{\expecto{X|A=a}{\id{C \geq \hat{\alpha}_a^+}C}}{\pi_{1|a}} - \frac{\hat{\mathrm{E}}_{X|A=a}[\id{C \geq \hat{\alpha}_a^+}C]}{\pi_{1|a}} \\
    &+ \frac{\hat{\mathrm{E}}_{X|A=a}[\id{C \geq \hat{\alpha}_a^+}C]}{\pi_{1|a}} - \frac{\hat{\mathrm{E}}_{X|A=a}[\id{C \geq \hat{\alpha}_a^+}C]}{\hat{\pi}_{1|a}^+}\Bigg| \\
    &\text{ (By introducing an intermediate term $\frac{\hat{\mathrm{E}}_{X|A=a}[\id{C \geq \hat{\alpha}_a^+}C]}{\pi_{1|a}}$)} \\
    &\leq \frac{1}{\pi_{1|a}} \cdot |\expecto{X|A=a}{\id{C \geq \hat{\alpha}_a^+}C}  - \hat{\mathrm{E}}_{X|A=a}[\id{C \geq \hat{\alpha}_a^+}C]| \\
    &+ \hat{\mathrm{E}}_{X|A=a}[\id{C \geq \hat{\alpha}_a^+}C]\cdot\Big| \frac{1}{\pi_{1|a}} - \frac{1}{\hat{\pi}_{1|a}^+}\Big| \\
    &\leq \frac{1}{\pi_{1|a}}|\expecto{X|A=a}{\id{C \geq \hat{\alpha}_a^+}C}  - \hat{\mathrm{E}}_{X|A=a}[\id{C \geq \hat{\alpha}_a^+}C]| + \Big| \frac{1}{\pi_{1|a}} - \frac{1}{\hat{\pi}_{1|a}^+}\Big| \\
    &\text{ (Since $\hat{\mathrm{E}}_{X|A=a}[\id{C \geq \hat{\alpha}_a^+}C] \leq 1$)}
\end{align*}

Now we can use the union bound:

\begin{align*}
    &\prob{|\rho_a(\hat{g}) - \hat{\rho}_a(\hat{g})| > \epsilon/4} \\
    &\leq \prob{\frac{1}{\pi_{1|a}}|\expecto{X|A=a}{\id{C \geq \hat{\alpha}_a^+}C}  - \hat{\mathrm{E}}_{X|A=a}[\id{C \geq \hat{\alpha}_a^+}C]| > \epsilon/8} \\
    &+ \prob{\Big| \frac{1}{\pi_{1|a}} - \frac{1}{\hat{\pi}_{1|a}^+}\Big| > \epsilon/8}
\end{align*}

To further simplify $\Big| \frac{1}{\pi_{1|a}} - \frac{1}{\hat{\pi}_{1|a}^+}\Big|$, we can borrow our analysis from the bias proof. Recall $\mathcal{G} = \{\hat{\pi}_{1|a} \geq \frac{\pi_{1|a}}{2}\}$ to be the event that $\hat{\pi}_{1|a}$ and $\pi_{1|a}$ are close. Note that on $\mathcal{G}$, $\hat{\pi}_{1|a}^+ = \hat{\pi}_{1|a}$ for sufficiently many samples ($1/N_{\alpha,a} \leq \pi_{1|a}/2$). On this good event $\mathcal{G}$, we derived that $\Big| \frac{1}{\pi_{1|a}} - \frac{1}{\hat{\pi}_{1|a}^+}\Big| \leq \frac{2}{\pi_{1|a}^2} |\hat{\pi}_{1|a} - \pi_{1|a}|$. 

\begin{align*}
    &\prob{|\rho_a(\hat{g}) - \hat{\rho}_a(\hat{g})| > \epsilon/4} \\
    &\leq \prob{\frac{1}{\pi_{1|a}}|\expecto{X|A=a}{\id{C \geq \hat{\alpha}_a^+}C}  - \hat{\mathrm{E}}_{X|A=a}[\id{C \geq \hat{\alpha}_a^+}C]| > \epsilon/8} \\
    &+ \prob{\Big| \frac{1}{\pi_{1|a}} - \frac{1}{\hat{\pi}_{1|a}^+}\Big| > \epsilon/8} \\
    &= \prob{\frac{1}{\pi_{1|a}}|\expecto{X|A=a}{\id{C \geq \hat{\alpha}_a^+}C}  - \hat{\mathrm{E}}_{X|A=a}[\id{C \geq \hat{\alpha}_a^+}C]| > \epsilon/8} \\
    &+ \prob{\Big| \frac{1}{\pi_{1|a}} - \frac{1}{\hat{\pi}_{1|a}^+}\Big| > \epsilon/8, \mathcal{G}} + \prob{\Big| \frac{1}{\pi_{1|a}} - \frac{1}{\hat{\pi}_{1|a}^+}\Big| > \epsilon/8, \mathcal{G}_C} \\
    &\leq \prob{\frac{1}{\pi_{1|a}}|\expecto{X|A=a}{\id{C \geq \hat{\alpha}_a^+}C}  - \hat{\mathrm{E}}_{X|A=a}[\id{C \geq \hat{\alpha}_a^+}C]| > \epsilon/8} \\
    &+ \prob{\frac{2}{\pi_{1|a}^2} |\hat{\pi}_{1|a} - \pi_{1|a}| > \epsilon/8} + \prob{\mathcal{G}_C} \\
    &= \prob{|\expecto{X|A=a}{\id{C \geq \hat{\alpha}_a^+}C}  - \hat{\mathrm{E}}_{X|A=a}[\id{C \geq \hat{\alpha}_a^+}C]| > \frac{\pi_{1|a}\epsilon}{8}} \\
    &+ \prob{|\hat{\pi}_{1|a} - \pi_{1|a}| > \frac{\pi_{1|a}^2\epsilon}{16}} + \prob{\hat{\pi}_{1|a} < \frac{\pi_{1|a}}{2}} \\
    &\leq 2\Bigg( \exp\Big( -\frac{N_{\beta,a}\pi_{1|a}^2\epsilon^2}{32} \Big) + \exp\Big( -\frac{N_{\alpha,a}\pi_{1|a}^4\epsilon^2}{128} \Big) + \exp\Big( -\frac{N_{\alpha,a}\pi_{1|a}^2}{2} \Big) \Bigg) \\
    &\text{ (By using Hoeffding's Inequality on each event).} \\
\end{align*}

We can now derive our sample complexities by splitting out $\delta/4$ budget between all three terms to obtain $N_{\beta,a} \geq \frac{32}{\pi_{1|a}^2 \epsilon^2} \log\left( \frac{24}{\delta} \right)$ and $N_{\alpha,a} \geq \max\left\{ \frac{2}{\pi_{1|a}^2}\log\left( \frac{24}{\delta} \right), \frac{128}{\pi_{1|a}^4\epsilon^2}\log\left( \frac{24}{\delta} \right) \right\}$.

Overall, our sample complexities take the following form:




\begin{align*}
    N_{\alpha} = \mathcal{O}\left( \frac{\lambda^2\max\{L_0^2, L_1^2\}}{\epsilon^2\tau^4}\log\left(\frac{1}{\delta}\right) \right) \text{ and } N_{\beta} = \mathcal{O}\left\{ \frac{1}{\epsilon^2} \log\left( \frac{1}{\delta} \right) \right\}.
\end{align*}



\end{proof}

\section{Details for the Tradeoff Estimation algorithm and Experiments}

\begin{algorithm}[t]
    \small 
    \caption{Tradeoff Estimation with Soft Labels}
    \label{algo:pareto}
    \let\AND\relax 
    \begin{algorithmic}[1]
        \STATE \textbf{Input:} Disjoint datasets $D_\alpha, D_\beta$; Unfairness Tolerance $\gamma$, Precision $\xi$; Priors $\hat{\Pi}$
        \STATE \textbf{Functions:} \\
        - $\mathrm{AlphaEst}(\lambda, \hat{\Pi}) \to \hat{\alpha}_a^+$ (Function to estimate the cost-sensitive thresholds $\alpha_a$ given a $\lambda$ and distribution priors);\\
        - $\mathrm{ErrUnfEst}(\lambda, \mathrm{AlphaEst}(\lambda, \hat{\Pi})) \to R(g_{\hat{\alpha}_a^+}), \Delta(g_{\hat{\alpha}_a^+})$ (Function to estimate the error and unfairness, given a threshold $\alpha_a$ computed using the previous function)
        
        \STATE $\hat{R}_0, \hat{\Delta}_0 \leftarrow \mathrm{ErrUnfEst}(0, \hat{\alpha}_0)$
        \STATE \textbf{if} {$|\hat{\Delta}_0| \leq \gamma$} \textbf{then} return No-Tradeoff.
        \STATE Set range: \textbf{if} $\hat{\Delta}_0 > 0$ \textbf{then} $\lambda \in [0, \frac{1}{\inf w}]$ \textbf{else} $\lambda \in [-\frac{1}{\inf w}, 0]$
        
        \STATE $\mathrm{points} \leftarrow \{\}$
        \WHILE{$\lambda_{\max} - \lambda_{\min} > \xi$}
            \STATE $\lambda \leftarrow (\lambda_{\max} + \lambda_{\min})/2$
            \STATE $\hat{R}, \hat{\Delta} \leftarrow \mathrm{ErrUnfEst}(\lambda, \mathrm{AlphaEst}(\lambda, \hat{\Pi}))$
            \STATE Append $(\lambda, \hat{R}, \hat{\Delta})$ to $\mathrm{points}$
            
            \STATE \textbf{if} $|\hat{\Delta}| \leq \gamma$ \textbf{then} break loop
            \STATE \textbf{if} $\hat{\Delta} > \gamma$ \textbf{then} $\lambda_{\max} \leftarrow \lambda$ \textbf{else} $\lambda_{\min} \leftarrow \lambda$
        \ENDWHILE
        
        \STATE return $\mathrm{points}$
    \end{algorithmic}
\end{algorithm}

With all the tools in place, we now present the algorithm for estimating the Pareto frontier using the estimators derived using soft labels in the previous section. We will now describe the algorithm formally in Algorithm \ref{algo:pareto} and discuss each design choice in more detail. 

\begin{itemize}
    \item [1-2] Assuming access to sufficiently large, disjoint datasets $D_\alpha. D_\beta$ sampled i.i.d. from our target distribution, we obtain estimates for all the distribution-dependent quantities $\hat{\Pi} = \{\hat\pi, \hat\pi_a, \hat\pi_{ya}, \hat\pi_{y|a}\}$. Using these priors $\hat{\Pi}$ and any $\lambda$, we can estimate our thresholds $\hat{\alpha}_a^+$ ($\mathrm{AlphaEst}$) and then our risk $R$ and unfairness $\Delta$ ($\mathrm{ErrUnfEst}$).
    \item [3-4] If the Bayes optimal classifier (at $\lambda=0$) already satisfies the required fairness ($|\hat\Delta_{0}| \leq \gamma$), then we are done.
    \item [5] Depending on sign of $\hat{\Delta}_0$, we can set $\lambda_{\max}$ from Lemma \ref{lem:lambda_max} to fix our range for the bisection search.
    \item [8-12] We compute $\lambda$ and calculate the updated $\hat{R}, \hat{\Delta}$. If we immediately achieve our required unfairness $\gamma$, we are done. Otherwise, depending on the sign, we update our bound $\lambda_{\max}/\lambda_{\min}$ and continue until the gap between $\lambda_{\min}$ and $\lambda_{\max}$ is under a fixed precision $\xi$.
\end{itemize}

\textbf{Details of Data and Experiments}: The synthetic dataset setup in our experiments is derived from Zeng et al. \cite{zeng2022bayes},  which consists of a binary label $y \in \{0,1\}$ and a binary sensitive attribute $a \in \{0,1\}$. We sample data from subgroup-specific $100$-dimensional Gaussian distributions $X \sim \mathcal{N}(\mu_{ya},\sigma^{2}I_{100})$ where $I_{100}$ denotes a $100 \times 100$ identity matrix and all entries of $\mu_{ya}$ are sampled from $U(0,1)$. We sample $20,000$ training examples and $5000$ test examples. 

The overall setup of soft labels is borrowed from previous state-of-the-art studies in Bayes error estimation: Ishida et al. \cite{ishidaperformance} and Ushio et al. \cite{ushio2025practical}. CIFAR-10H \cite{petersen2021post} is a dataset constructed using 50 additional annotations of the CIFAR-10 test dataset. The ground-truth soft label is constructed as the average of the underlying hard labels. Since the original dataset is multi-class (10 classes), the binary labels are constructed by grouping classes into two: land vs. other animals, animal vs. artifacts, odd vs. even, and so on. Because we are working with unfairness constraints that also include a notion of a sensitive group, we assume one of these partitions is the class and the other is the sensitive group. Note that no dataset of soft labels exists for standard fairness datasets, and hence, we had to retrofit a popular benchmark from the soft-label-based Bayes error estimation literature. The code for setting up the CIFAR-10H experiments is from Ishida et al. \cite{ishidaperformance} \footnote{https://github.com/ishida-lab/irreducible}. The code and data to reproduce the experiments is provided \href{https://github.com/mohitsharma29/Soft-Labels-Fairness}{here}.



\newpage
\section*{NeurIPS Paper Checklist}

\begin{enumerate}

\item {\bf Claims}
    \item[] Question: Do the main claims made in the abstract and introduction accurately reflect the paper's contributions and scope?
    \item[] Answer: \answerYes{} 
    \item[] Justification: The main contributions have been well motivated and highlighted in the abstract and the introduction.
    \item[] Guidelines:
    \begin{itemize}
        \item The answer \answerNA{} means that the abstract and introduction do not include the claims made in the paper.
        \item The abstract and/or introduction should clearly state the claims made, including the contributions made in the paper and important assumptions and limitations. A \answerNo{} or \answerNA{} answer to this question will not be perceived well by the reviewers. 
        \item The claims made should match theoretical and experimental results, and reflect how much the results can be expected to generalize to other settings. 
        \item It is fine to include aspirational goals as motivation as long as it is clear that these goals are not attained by the paper. 
    \end{itemize}

\item {\bf Limitations}
    \item[] Question: Does the paper discuss the limitations of the work performed by the authors?
    \item[] Answer: \answerYes{} 
    \item[] Justification: The conclusion section discusses limitations and future extensions of our work.
    \item[] Guidelines:
    \begin{itemize}
        \item The answer \answerNA{} means that the paper has no limitation, while the answer \answerNo{} means that the paper has limitations, but those are not discussed in the paper. 
        \item The authors are encouraged to create a separate ``Limitations'' section in their paper.
        \item The paper should point out any strong assumptions and how robust the results are to violations of these assumptions (e.g., independence assumptions, noiseless settings, model well-specification, asymptotic approximations only holding locally). The authors should reflect on how these assumptions might be violated in practice and what the implications would be.
        \item The authors should reflect on the scope of the claims made, e.g., if the approach was only tested on a few datasets or with a few runs. In general, empirical results often depend on implicit assumptions, which should be articulated.
        \item The authors should reflect on the factors that influence the performance of the approach. For example, a facial recognition algorithm may perform poorly when image resolution is low or images are taken in low lighting. Or a speech-to-text system might not be used reliably to provide closed captions for online lectures because it fails to handle technical jargon.
        \item The authors should discuss the computational efficiency of the proposed algorithms and how they scale with dataset size.
        \item If applicable, the authors should discuss possible limitations of their approach to address problems of privacy and fairness.
        \item While the authors might fear that complete honesty about limitations might be used by reviewers as grounds for rejection, a worse outcome might be that reviewers discover limitations that aren't acknowledged in the paper. The authors should use their best judgment and recognize that individual actions in favor of transparency play an important role in developing norms that preserve the integrity of the community. Reviewers will be specifically instructed to not penalize honesty concerning limitations.
    \end{itemize}

\item {\bf Theory assumptions and proofs}
    \item[] Question: For each theoretical result, does the paper provide the full set of assumptions and a complete (and correct) proof?
    \item[] Answer: \answerYes{} 
    \item[] Justification: All assumptions required are included in the main body and all the proofs are provided in the Appendix.
    \item[] Guidelines:
    \begin{itemize}
        \item The answer \answerNA{} means that the paper does not include theoretical results. 
        \item All the theorems, formulas, and proofs in the paper should be numbered and cross-referenced.
        \item All assumptions should be clearly stated or referenced in the statement of any theorems.
        \item The proofs can either appear in the main paper or the supplemental material, but if they appear in the supplemental material, the authors are encouraged to provide a short proof sketch to provide intuition. 
        \item Inversely, any informal proof provided in the core of the paper should be complemented by formal proofs provided in appendix or supplemental material.
        \item Theorems and Lemmas that the proof relies upon should be properly referenced. 
    \end{itemize}

    \item {\bf Experimental result reproducibility}
    \item[] Question: Does the paper fully disclose all the information needed to reproduce the main experimental results of the paper to the extent that it affects the main claims and/or conclusions of the paper (regardless of whether the code and data are provided or not)?
    \item[] Answer: \answerYes{} 
    \item[] Justification: A link to anonymized code to reproduce the empirical results has been included in the Appendix.
    \item[] Guidelines:
    \begin{itemize}
        \item The answer \answerNA{} means that the paper does not include experiments.
        \item If the paper includes experiments, a \answerNo{} answer to this question will not be perceived well by the reviewers: Making the paper reproducible is important, regardless of whether the code and data are provided or not.
        \item If the contribution is a dataset and\slash or model, the authors should describe the steps taken to make their results reproducible or verifiable. 
        \item Depending on the contribution, reproducibility can be accomplished in various ways. For example, if the contribution is a novel architecture, describing the architecture fully might suffice, or if the contribution is a specific model and empirical evaluation, it may be necessary to either make it possible for others to replicate the model with the same dataset, or provide access to the model. In general. releasing code and data is often one good way to accomplish this, but reproducibility can also be provided via detailed instructions for how to replicate the results, access to a hosted model (e.g., in the case of a large language model), releasing of a model checkpoint, or other means that are appropriate to the research performed.
        \item While NeurIPS does not require releasing code, the conference does require all submissions to provide some reasonable avenue for reproducibility, which may depend on the nature of the contribution. For example
        \begin{enumerate}
            \item If the contribution is primarily a new algorithm, the paper should make it clear how to reproduce that algorithm.
            \item If the contribution is primarily a new model architecture, the paper should describe the architecture clearly and fully.
            \item If the contribution is a new model (e.g., a large language model), then there should either be a way to access this model for reproducing the results or a way to reproduce the model (e.g., with an open-source dataset or instructions for how to construct the dataset).
            \item We recognize that reproducibility may be tricky in some cases, in which case authors are welcome to describe the particular way they provide for reproducibility. In the case of closed-source models, it may be that access to the model is limited in some way (e.g., to registered users), but it should be possible for other researchers to have some path to reproducing or verifying the results.
        \end{enumerate}
    \end{itemize}

\item {\bf Open access to data and code}
    \item[] Question: Does the paper provide open access to the data and code, with sufficient instructions to faithfully reproduce the main experimental results, as described in supplemental material?
    \item[] Answer: \answerYes{} 
    \item[] Justification: A link to anonymized code to reproduce the empirical results has been included in the Appendix.
    \item[] Guidelines:
    \begin{itemize}
        \item The answer \answerNA{} means that paper does not include experiments requiring code.
        \item Please see the NeurIPS code and data submission guidelines (\url{https://neurips.cc/public/guides/CodeSubmissionPolicy}) for more details.
        \item While we encourage the release of code and data, we understand that this might not be possible, so \answerNo{} is an acceptable answer. Papers cannot be rejected simply for not including code, unless this is central to the contribution (e.g., for a new open-source benchmark).
        \item The instructions should contain the exact command and environment needed to run to reproduce the results. See the NeurIPS code and data submission guidelines (\url{https://neurips.cc/public/guides/CodeSubmissionPolicy}) for more details.
        \item The authors should provide instructions on data access and preparation, including how to access the raw data, preprocessed data, intermediate data, and generated data, etc.
        \item The authors should provide scripts to reproduce all experimental results for the new proposed method and baselines. If only a subset of experiments are reproducible, they should state which ones are omitted from the script and why.
        \item At submission time, to preserve anonymity, the authors should release anonymized versions (if applicable).
        \item Providing as much information as possible in supplemental material (appended to the paper) is recommended, but including URLs to data and code is permitted.
    \end{itemize}

\item {\bf Experimental setting/details}
    \item[] Question: Does the paper specify all the training and test details (e.g., data splits, hyperparameters, how they were chosen, type of optimizer) necessary to understand the results?
    \item[] Answer: \answerYes{} 
    \item[] Justification: Data preparation scripts and the random seeds used are provided in the anonymized code. 
    \item[] Guidelines:
    \begin{itemize}
        \item The answer \answerNA{} means that the paper does not include experiments.
        \item The experimental setting should be presented in the core of the paper to a level of detail that is necessary to appreciate the results and make sense of them.
        \item The full details can be provided either with the code, in appendix, or as supplemental material.
    \end{itemize}

\item {\bf Experiment statistical significance}
    \item[] Question: Does the paper report error bars suitably and correctly defined or other appropriate information about the statistical significance of the experiments?
    \item[] Answer: \answerYes{} 
    \item[] Justification: The experiments are performed with bootstrapping, and the error bars are also reported with the results.
    \item[] Guidelines:
    \begin{itemize}
        \item The answer \answerNA{} means that the paper does not include experiments.
        \item The authors should answer \answerYes{} if the results are accompanied by error bars, confidence intervals, or statistical significance tests, at least for the experiments that support the main claims of the paper.
        \item The factors of variability that the error bars are capturing should be clearly stated (for example, train/test split, initialization, random drawing of some parameter, or overall run with given experimental conditions).
        \item The method for calculating the error bars should be explained (closed form formula, call to a library function, bootstrap, etc.)
        \item The assumptions made should be given (e.g., Normally distributed errors).
        \item It should be clear whether the error bar is the standard deviation or the standard error of the mean.
        \item It is OK to report 1-sigma error bars, but one should state it. The authors should preferably report a 2-sigma error bar than state that they have a 96\% CI, if the hypothesis of Normality of errors is not verified.
        \item For asymmetric distributions, the authors should be careful not to show in tables or figures symmetric error bars that would yield results that are out of range (e.g., negative error rates).
        \item If error bars are reported in tables or plots, the authors should explain in the text how they were calculated and reference the corresponding figures or tables in the text.
    \end{itemize}

\item {\bf Experiments compute resources}
    \item[] Question: For each experiment, does the paper provide sufficient information on the computer resources (type of compute workers, memory, time of execution) needed to reproduce the experiments?
    \item[] Answer: \answerNA{} 
    \item[] Justification: Our experiments are lightweight and can be run on most systems with moderate compute power. 
    \item[] Guidelines:
    \begin{itemize}
        \item The answer \answerNA{} means that the paper does not include experiments.
        \item The paper should indicate the type of compute workers CPU or GPU, internal cluster, or cloud provider, including relevant memory and storage.
        \item The paper should provide the amount of compute required for each of the individual experimental runs as well as estimate the total compute. 
        \item The paper should disclose whether the full research project required more compute than the experiments reported in the paper (e.g., preliminary or failed experiments that didn't make it into the paper). 
    \end{itemize}
    
\item {\bf Code of ethics}
    \item[] Question: Does the research conducted in the paper conform, in every respect, with the NeurIPS Code of Ethics \url{https://neurips.cc/public/EthicsGuidelines}?
    \item[] Answer: \answerYes{}  
    \item[] Justification: All points in the code of ethics have been conformed to. We do not believe our research process has caused harm --- it does not involve human participation, and we use standard public datasets from the literature. Our research will help society curb discrimination by directly tackling the themes of bias and unfairness. We do not see any possible adverse impact on society due to our research work.
    \item[] Guidelines:
    \begin{itemize}
        \item The answer \answerNA{} means that the authors have not reviewed the NeurIPS Code of Ethics.
        \item If the authors answer \answerNo, they should explain the special circumstances that require a deviation from the Code of Ethics.
        \item The authors should make sure to preserve anonymity (e.g., if there is a special consideration due to laws or regulations in their jurisdiction).
    \end{itemize}

\item {\bf Broader impacts}
    \item[] Question: Does the paper discuss both potential positive societal impacts and negative societal impacts of the work performed?
    \item[] Answer: \answerYes{} 
    \item[] Justification:  We discuss the benefits of our work for practitioners and the society at large, and do not foresee any negative impact.
    \item[] Guidelines:
    \begin{itemize}
        \item The answer \answerNA{} means that there is no societal impact of the work performed.
        \item If the authors answer \answerNA{} or \answerNo, they should explain why their work has no societal impact or why the paper does not address societal impact.
        \item Examples of negative societal impacts include potential malicious or unintended uses (e.g., disinformation, generating fake profiles, surveillance), fairness considerations (e.g., deployment of technologies that could make decisions that unfairly impact specific groups), privacy considerations, and security considerations.
        \item The conference expects that many papers will be foundational research and not tied to particular applications, let alone deployments. However, if there is a direct path to any negative applications, the authors should point it out. For example, it is legitimate to point out that an improvement in the quality of generative models could be used to generate Deepfakes for disinformation. On the other hand, it is not needed to point out that a generic algorithm for optimizing neural networks could enable people to train models that generate Deepfakes faster.
        \item The authors should consider possible harms that could arise when the technology is being used as intended and functioning correctly, harms that could arise when the technology is being used as intended but gives incorrect results, and harms following from (intentional or unintentional) misuse of the technology.
        \item If there are negative societal impacts, the authors could also discuss possible mitigation strategies (e.g., gated release of models, providing defenses in addition to attacks, mechanisms for monitoring misuse, mechanisms to monitor how a system learns from feedback over time, improving the efficiency and accessibility of ML).
    \end{itemize}
    
\item {\bf Safeguards}
    \item[] Question: Does the paper describe safeguards that have been put in place for responsible release of data or models that have a high risk for misuse (e.g., pre-trained language models, image generators, or scraped datasets)?
    \item[] Answer: \answerNA{} 
    \item[] Justification: Our paper does not involve any data or models that can cause harm or have a risk of misuse.
    \item[] Guidelines:
    \begin{itemize}
        \item The answer \answerNA{} means that the paper poses no such risks.
        \item Released models that have a high risk for misuse or dual-use should be released with necessary safeguards to allow for controlled use of the model, for example by requiring that users adhere to usage guidelines or restrictions to access the model or implementing safety filters. 
        \item Datasets that have been scraped from the Internet could pose safety risks. The authors should describe how they avoided releasing unsafe images.
        \item We recognize that providing effective safeguards is challenging, and many papers do not require this, but we encourage authors to take this into account and make a best faith effort.
    \end{itemize}

\item {\bf Licenses for existing assets}
    \item[] Question: Are the creators or original owners of assets (e.g., code, data, models), used in the paper, properly credited and are the license and terms of use explicitly mentioned and properly respected?
    \item[] Answer: \answerYes{} 
    \item[] Justification: We have credited the authors of the datasets and the codebases we used as baselines.
    \item[] Guidelines:
    \begin{itemize}
        \item The answer \answerNA{} means that the paper does not use existing assets.
        \item The authors should cite the original paper that produced the code package or dataset.
        \item The authors should state which version of the asset is used and, if possible, include a URL.
        \item The name of the license (e.g., CC-BY 4.0) should be included for each asset.
        \item For scraped data from a particular source (e.g., website), the copyright and terms of service of that source should be provided.
        \item If assets are released, the license, copyright information, and terms of use in the package should be provided. For popular datasets, \url{paperswithcode.com/datasets} has curated licenses for some datasets. Their licensing guide can help determine the license of a dataset.
        \item For existing datasets that are re-packaged, both the original license and the license of the derived asset (if it has changed) should be provided.
        \item If this information is not available online, the authors are encouraged to reach out to the asset's creators.
    \end{itemize}

\item {\bf New assets}
    \item[] Question: Are new assets introduced in the paper well documented, and is the documentation provided alongside the assets?
    \item[] Answer: \answerNA{} 
    \item[] Justification: The link to the anonymized code implementation is included in the Appendix.
    \item[] Guidelines:
    \begin{itemize}
        \item The answer \answerNA{} means that the paper does not release new assets.
        \item Researchers should communicate the details of the dataset\slash code\slash model as part of their submissions via structured templates. This includes details about training, license, limitations, etc. 
        \item The paper should discuss whether and how consent was obtained from people whose asset is used.
        \item At submission time, remember to anonymize your assets (if applicable). You can either create an anonymized URL or include an anonymized zip file.
    \end{itemize}

\item {\bf Crowdsourcing and research with human subjects}
    \item[] Question: For crowdsourcing experiments and research with human subjects, does the paper include the full text of instructions given to participants and screenshots, if applicable, as well as details about compensation (if any)? 
    \item[] Answer: \answerNA{} 
    \item[] Justification: No human subjects involved.
    \item[] Guidelines:
    \begin{itemize}
        \item The answer \answerNA{} means that the paper does not involve crowdsourcing nor research with human subjects.
        \item Including this information in the supplemental material is fine, but if the main contribution of the paper involves human subjects, then as much detail as possible should be included in the main paper. 
        \item According to the NeurIPS Code of Ethics, workers involved in data collection, curation, or other labor should be paid at least the minimum wage in the country of the data collector. 
    \end{itemize}

\item {\bf Institutional review board (IRB) approvals or equivalent for research with human subjects}
    \item[] Question: Does the paper describe potential risks incurred by study participants, whether such risks were disclosed to the subjects, and whether Institutional Review Board (IRB) approvals (or an equivalent approval/review based on the requirements of your country or institution) were obtained?
    \item[] Answer: \answerNA{} 
    \item[] Justification: Human subjects not involved.
    \item[] Guidelines:
    \begin{itemize}
        \item The answer \answerNA{} means that the paper does not involve crowdsourcing nor research with human subjects.
        \item Depending on the country in which research is conducted, IRB approval (or equivalent) may be required for any human subjects research. If you obtained IRB approval, you should clearly state this in the paper. 
        \item We recognize that the procedures for this may vary significantly between institutions and locations, and we expect authors to adhere to the NeurIPS Code of Ethics and the guidelines for their institution. 
        \item For initial submissions, do not include any information that would break anonymity (if applicable), such as the institution conducting the review.
    \end{itemize}

\item {\bf Declaration of LLM usage}
    \item[] Question: Does the paper describe the usage of LLMs if it is an important, original, or non-standard component of the core methods in this research? Note that if the LLM is used only for writing, editing, or formatting purposes and does \emph{not} impact the core methodology, scientific rigor, or originality of the research, declaration is not required.
    \item[] Answer: \answerNA{} 
    \item[] Justification: LLMs were not used for any important, original or non-standard components.
    \item[] Guidelines:
    \begin{itemize}
        \item The answer \answerNA{} means that the core method development in this research does not involve LLMs as any important, original, or non-standard components.
        \item Please refer to our LLM policy in the NeurIPS handbook for what should or should not be described.
    \end{itemize}

\end{enumerate}

\end{document}